\documentclass[final]{cvpr}

\usepackage{times}
\usepackage{epsfig}
\usepackage{graphicx}
\usepackage{amsmath}
\usepackage{amssymb}
\usepackage{caption}
\usepackage{subcaption}
\usepackage[ruled]{algorithm2e}

\usepackage{xr}
\makeatletter
\newcommand*{\addFileDependency}[1]{
  \typeout{(#1)}
  \@addtofilelist{#1}
  \IfFileExists{#1}{}{\typeout{No file #1.}}
}
\makeatother
\newcommand*{\myexternaldocument}[1]{
    \externaldocument{#1}
    \addFileDependency{#1.tex}
    \addFileDependency{#1.aux}
}
\myexternaldocument{appendix}

\usepackage[pagebackref=true,breaklinks=true,colorlinks,bookmarks=false]{hyperref}

\newtheorem{definition}{\textbf{Definition}}

\newtheorem{proposition}{\textbf{Proposition}}

\usepackage{dsfont}
\usepackage{enumitem}
\usepackage[dvipsnames]{xcolor}
\usepackage{amsmath}
\usepackage{soul}
\usepackage{subcaption}
\usepackage{multirow}
\usepackage{booktabs}
\usepackage{csquotes}
\usepackage{graphicx}
\usepackage{mathtools}

\newcommand{\model}{h}
\newcommand{\image}{x}
\newcommand{\gt}{y}

\newcommand{\benchmark}{S}
\newcommand{\cluster}{C}
\newcommand{\leaf}{l}
\newcommand{\tree}{T}

\newcommand{\errorrate}{\mathrm{ER}}
\newcommand{\errorcoverage}{\mathrm{EC}}
\newcommand{\baseerrorrate}{\mathrm{BER}}
\newcommand{\avgtree}{\mathrm{ALER}}
\newcommand{\importance}{\mathrm{IV}}

\newcommand{\errorratetext}{ER}
\newcommand{\errorcoveragetext}{EC}
\newcommand{\baseerrorratetext}{BER}
\newcommand{\avgtreetext}{ALER}
\DeclareMathOperator*{\argmax}{arg\,max}

\def\methodname{Barlow}

\newcommand{\myquote}[1]{
     {{``}\emph{#1}{''}}
}

\usepackage{enumitem}

\setlist[description]{leftmargin=*,labelindent=*}
\setlist[itemize]{leftmargin=*}
\setlist[enumerate]{leftmargin=*}

\def\cvprPaperID{10484} 
\def\confYear{CVPR 2021}
\usepackage{nopageno}

\begin{document}

\title{Understanding Failures of Deep Networks via Robust Feature Extraction}



\author{
  Sahil Singla\thanks{Work carried out during a research internship at Microsoft Research.}\\
  University of Maryland\\
  \small{\texttt{ssingla@umd.edu}}
  \and
  Besmira Nushi, Shital Shah, Ece Kamar, Eric Horvitz\\
  Microsoft Research \\
   \small{\texttt{\{benushi,shitals,eckamar,horvitz\}}
  \texttt{@microsoft.com}}
}

\maketitle

\begin{abstract}
    Traditional evaluation metrics for learned models that report aggregate scores over a test set are insufficient for surfacing important and informative patterns of failure over features and instances. We introduce and study a method aimed at characterizing and explaining failures by identifying visual attributes whose presence or absence results in poor performance. In distinction to previous work that relies upon crowdsourced labels for visual attributes, we leverage the representation of a separate robust model to extract interpretable features and then harness these features to identify failure modes. We further propose a visualization method aimed at enabling humans to understand the meaning encoded in such features and we test the comprehensibility of the features. An evaluation of the methods on the ImageNet dataset demonstrates that: (i) the proposed workflow is effective for discovering important failure modes, (ii) the visualization techniques help humans to understand the extracted features, and (iii) the extracted insights can assist engineers with error analysis and debugging.
\end{abstract}

\section{Introduction}
It is critically important to understand the failure modes of machine learning (ML) systems, especially when they are employed in high-stakes applications. Aggregrate metrics in common use capture summary statistics on failure. While reporting overall performance is important, gaining an understanding of the specifics of failure is a core responsibility in the fielding of ML systems and components. For example, we need to understand situations where a self-driving car will fail to detect a pedestrian even when the system has high overall accuracy. Similarly, it is important to understand for which features misdiagnosis is most probable in chest x-rays even if the model has higher overall accuracy than humans. Such  situation-specific insights can guide the iterative process of model development and debugging.

\begin{figure*}[t!]
\centering
\begin{subfigure}{.19\textwidth}
  \includegraphics[trim=26cm 0 0 0.8cm, clip, width=0.9\linewidth]{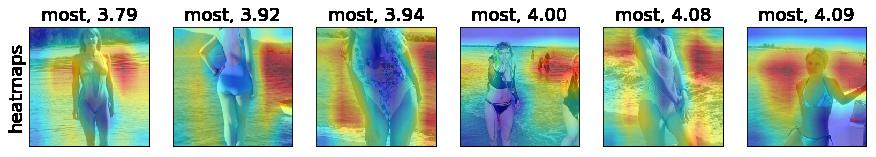}  
\end{subfigure}\ \ \begin{subfigure}{.19\textwidth}
  \includegraphics[trim=16cm 0 10cm 0.8cm, clip, width=0.9\linewidth]{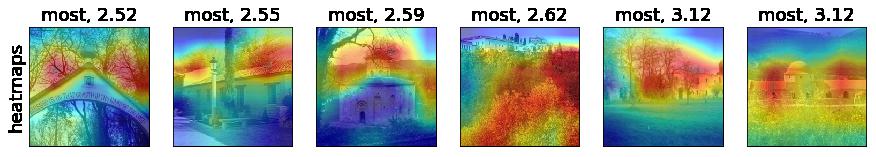}  
\end{subfigure}\ \ \begin{subfigure}{.19\textwidth}
  \includegraphics[trim=0.9cm 0 25.1cm 0.8cm, clip, width=0.9\linewidth]{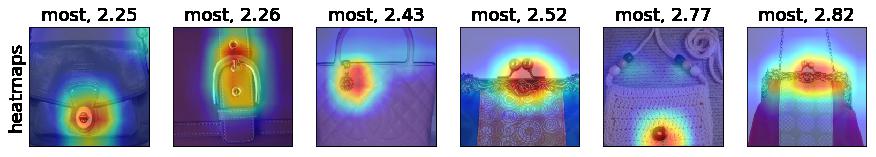}
\end{subfigure}\begin{subfigure}{.19\textwidth}
  \includegraphics[trim=25.9cm 0 0.1cm 0.8cm, clip, width=0.9\linewidth]{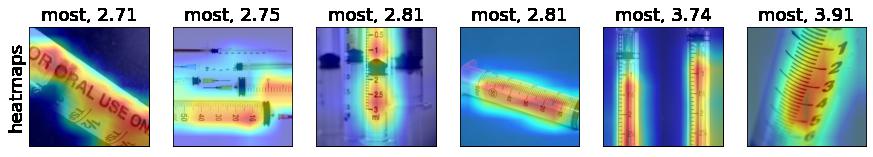}  
  \end{subfigure}\ \ \ \begin{subfigure}{.19\textwidth}
  \includegraphics[trim=6cm 0 20cm 0.8cm, clip, width=0.9\linewidth]{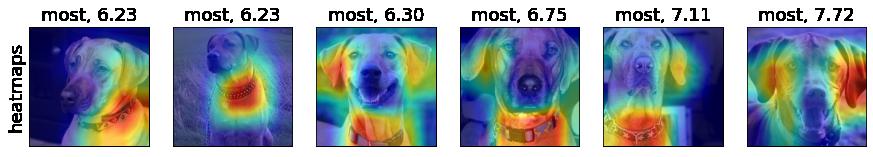}  
\end{subfigure}\\
\begin{subfigure}{.19\textwidth}
  \includegraphics[trim=26cm 0 0 0.8cm, clip, width=0.9\linewidth]{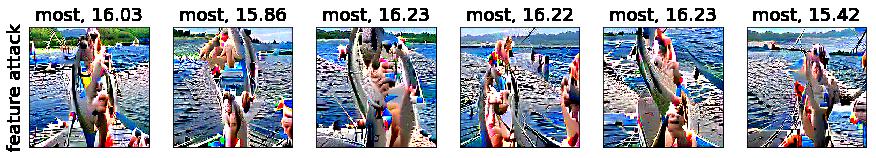}  
  \caption{class: maillot \\ feature: water \\ error increase: \textcolor{red}{\textbf{+9.72\%}} }
  \label{fig:attack_maillot}
\end{subfigure}\ \ \begin{subfigure}{.19\textwidth}
  \includegraphics[trim=16cm 0 10cm 0.8cm, clip, width=0.9\linewidth]{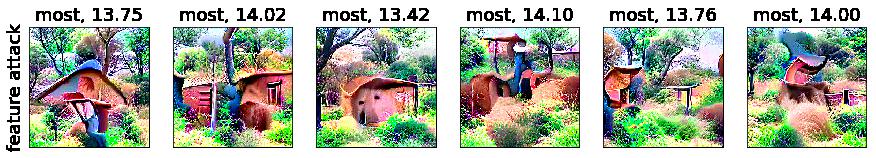}  
  \caption{class: monastery \\ feature: greenery \\ error increase: \textcolor{red}{\textbf{+24.84\%}} }
  \label{fig:attack_monastery}
\end{subfigure}\ \ \begin{subfigure}{.19\textwidth}
  \includegraphics[trim=0.9cm 0 25.1cm 0.8cm, clip, width=0.9\linewidth]{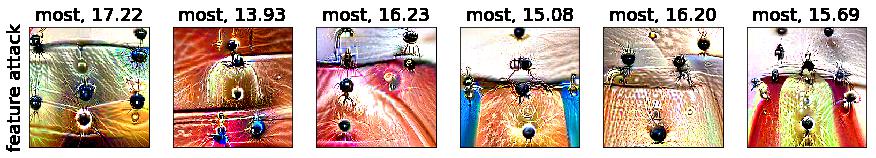}  
  \caption{class: purse \\ feature: buckle \\ error increase: \textcolor{red}{\textbf{+10.94\%}} }
  \label{fig:attack_purse}
\end{subfigure}\begin{subfigure}{.19\textwidth}
  \includegraphics[trim=25.9cm 0 0.1cm 0.8cm, clip, width=0.9\linewidth]{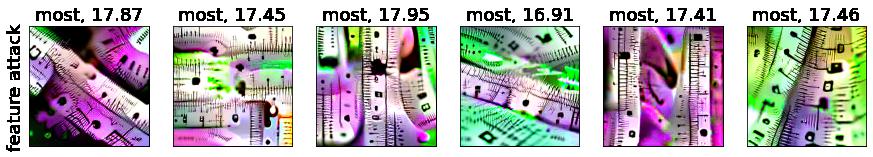} 
  \caption{class: syringe \\ feature: markings \\ error increase: \textcolor{red}{\textbf{+14.99\%}} }
  \label{fig:attack_syringe}
\end{subfigure}\ \ \ \begin{subfigure}{.19\textwidth}
  \includegraphics[trim=6cm 0 20cm 0.8cm, clip, width=0.9\linewidth]{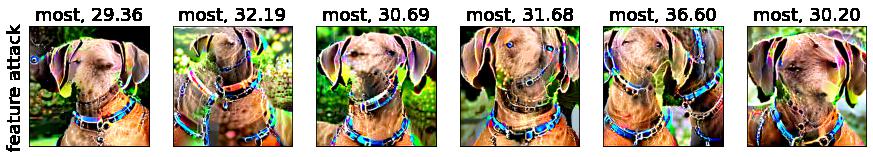}  
  \caption{class: rhodesian \\ feature: dog collar \\ error increase: \textcolor{red}{\textbf{+10.91\%}} }
  \label{fig:attack_rhodesian} 
\end{subfigure}
\caption{Failure modes discovered using the proposed methodology for a standard Resnet-50 neural network trained on ImageNet. In the top row, red denotes the region that a specific feature is paying attention to. In the bottom row, we show the image generated by visually amplifying the same feature. We observe that, due to the presence of spurious correlations, the failure rate of the model increases significantly on the relevant class. Additional examples in Appendix Section \ref{sec:failure_modes_large}.}
\label{fig:label_nonrobust}
\end{figure*}

Model performance can be wildly non-uniform for different clusterings of instances and such heterogeneity is not reflected by standard metrics such as AUC or accuracy. For example, it was shown in \cite{howard2017emotion} that a commercial model for emotion detection from facial expressions systematically failed for young children. Buolamwini \etal \cite{Buolamwini2018GenderSI} found that gender detection in multiple commercial models had significantly higher error rates for women with darker skin tone. These examples highlight the importance of identifying natural clusters in the data with high failure rates. However, practical problems with these approaches still remain: (a) they require an expensive and time-consuming collection of metadata by humans, and (b) visual attributes that machine learning procedures pay attention to can be very different from the ones humans focus on (see Appendix Section \ref{sec:failure_modes_large}).

To resolve these issues, we propose to leverage the internal representation of a robust model \cite{Madry2017TowardsDL} to generate the metadata. The key property that makes robust representations useful is that the features can be visualized more easily than for a standard model~\cite{ Engstrom2019LearningPR, tsipras2019robustness}. Our method, named  \methodname\footnote{In honor of perceptual psychologist, Horace Barlow \cite{Barlow72}.}, is inspired from Fault Tree Analysis~\cite{faulttree} in safety engineering and uses robust representations as a building block.\footnote{Code repository: https://github.com/singlasahil14/barlow} We demonstrate the results on the ImageNet dataset \cite{5206848} and find that it reveals two types of failures:

\noindent \textbf{Spurious correlations}: A spurious correlation is a feature that is causally unrelated to the desired class but is likely to co-occur with the same class in the training/test data. For example, food is likely to co-occur with plates. However, the absence of the food from a plate image should not result in misclassification (see examples in Figures \ref{fig:attack_maillot}, \ref{fig:attack_monastery} and \ref{fig:attack_rhodesian}).

\noindent \textbf{Overemphasized features}: An overemphasized feature is a feature that is causally related to the desired class but where the model gives \textit{excessive} importance for classification, disregarding the other relevant features, and is unable to make a correct prediction when that feature is absent from the image. For example, a model may be likely to fail on a purse image if the buckle is absent (see Figures \ref{fig:attack_purse} and \ref{fig:attack_syringe}). 

While determining the type of failure (spurious vs. causal but overemphasized feature) and formulating mitigation steps remains a task that depends on human expertise, \methodname\xspace assists practitioners in this process by efficiently identifying and providing visualizations of failure modes, showing how input characteristics \emph{correlate} with failures. For example, failures identified in Figure \ref{fig:label_nonrobust} suggest the following interventions: (a) add images of maillot/monastery with diverse backgrounds, Rhodesian Ridgebacks without collar (b) mask overemphasized features (buckle from purse, markings from syringe) in the training set. 

In summary, we provide the following contributions:
   \begin{enumerate}[leftmargin=7pt]
   \setlength{\itemsep}{2.0pt}
   \setlength{\parskip}{-2pt}
   \setlength{\parsep}{-2pt}
     \item An error analysis framework for discovering critical failure modes for a given model.
     \item A feature extraction and visualization method based on robust model representations to enable humans to understand the semantics of a learned feature.
     \item A large-scale crowdsourcing study to evaluate the effectiveness of the visualization technique and the interpretability of robust feature representations.
      \item A user study with engineers with experience using machine learning for vision tasks to evaluate the effectiveness of the methodology for model debugging.
   \end{enumerate}

\section{Related work}

\noindent \textbf{Feature Visualization, Interpretability, and Robustness}: For a trained neural network, feature visualization creates images to either (i) visualize the region that neurons are paying attention to using heatmaps, or (ii) maximize the activation of neurons that we are interested in and visualize the resulting image. Previous works \cite{ZeilerF13, YosinskiCNFL15, DosovitskiyB15, MahendranV15, NguyenYC16, ZhouKLOT15, Simonyan2013DeepIC, SelvarajuDVCPB16, olah2018the, sturmfels2020visualizing, carter2019activation, Bau30071} provide evidence that the internal representations of a neural network can capture important semantic concepts. For example, the network dissection method \cite{zhou2018interpreting,Bau30071,zhou2018interpreting} visualizes hidden network units by showing regions that are most activated for that unit and correlates the activation maps with dense human-labeled concepts. In our studies, we observe that activation maps alone are not sufficient for narrowing down the feature concept and that methods for maximizing neuron activation produce noisy visualizations with changes not visible to the human eye (see Appendix Section \ref{sec:comparison_robust_nonrobust}). Similar limitations are also discussed by 
Olah \etal\cite{olah2017feature} in the context of understanding the interpretability of feature visualization methods for standard models. Recent work~\cite{tsipras2019robustness,allen2020feature,Madry2017TowardsDL} shows that saliency maps are qualitatively more interpretable for adversarially trained models (when compared to standard training) and align well with human perception. 
Moreover, Engstrom \etal  \cite{Engstrom2019LearningPR} showed that, for robust models (in contrast to standard models), optimizing an image directly to maximize a certain neuron helps with visualizing the respective relevant learned features. We note that disentangled representations \cite{kulkarni2015deep, Higgins2017betaVAELB, burgess2018understanding, kim2019disentangling, chen2019isolating, locatello2019challenging} based on VAEs~\cite{KingmaW13} can also be used for feature extraction. However, it is difficult to scale these methods to large-scale, rich datasets such as ImageNet. \\
\noindent \textbf{Failure explanation}: In recent years, there has been increasing interest in the understandability of predictions made by machine-learned models \cite{Simonyan2013DeepIC, Smilkov2017SmoothGradRN, Sundararajan2017AxiomaticAF, sanitychecks2018, SelvarajuDVCPB16}. Most of these efforts have focused on \emph{local} explanations, where decisions about single images are inspected~\cite{Engstrom2019LearningPR, deviparikhcounterfactual,SelvarajuDVCPB16, Goyal2019ExplainingCW, chang2018explaining, oshaughnessy2020generative, verma2020counterfactual, Yeh2019OnT}. Examining numerous local explanations for debugging can however be time-consuming. Thus, we focus on identifying major failure modes across the entire dataset and presenting them in a useful way to guide model improvement by actions as fixing the data distribution and performing data augmentation. For this, we build upon prior work that discovers generalizable failure modes based on metadata or features~\cite{pandora1809, zhang2018manifold, chung2019slice, wu2019errudite}. These methods operate typically on tabular data where features are already available~\cite{chung2019slice, zhang2018manifold}, on language data with query-definable text operators~\cite{wu2019errudite}, or on image data where features are collected from intermediate multi-component output and crowdsourcing~\cite{pandora1809}. In this work, we focus on identifying failure modes for images but with features extracted in an automated manner from learned robust representations. 

\section{Background and method overview}\label{sec:background}
Let $\model: \image \rightarrow \gt$ be a trained neural network that classifies an input image $x$ to one of the classes $y \in Y$. For a cluster of images $\cluster$ in an overall benchmark $\benchmark$ (i.e., $\cluster \subset \benchmark$), we use the following  definitions to quantify failures:
\begin{definition}
\label{def:er}
The error rate of a cluster is the fraction of images in the cluster that are misclassified:
\begin{align*}
\errorrate(\cluster) =\frac{\sum_{\image \in \cluster}{\Large\mathds{1}}_{\model(\image) \neq \gt}(\image)}{|\cluster|}
\end{align*}
\end{definition}

\begin{definition}
\label{def:ec}
The error coverage of a cluster is the fraction of all errors in the benchmark that fall in the cluster:
\begin{align*}
\errorcoverage(\cluster) =\frac{\sum_{\image \in \cluster}{\Large\mathds{1}}_{\model(\image) \neq \gt}(\image)}{\sum_{\image \in \benchmark}{\Large\mathds{1}}_{\model(\image) \neq \gt}(\image)}
\end{align*}
\end{definition}

\begin{definition}
The base error rate is the error rate for the whole benchmark, treating it as one cluster:
\begin{align*}
\baseerrorrate = \errorrate(S)
\end{align*}
\end{definition}

\begin{figure}[t]
\begin{center}
\includegraphics[ width=\linewidth]{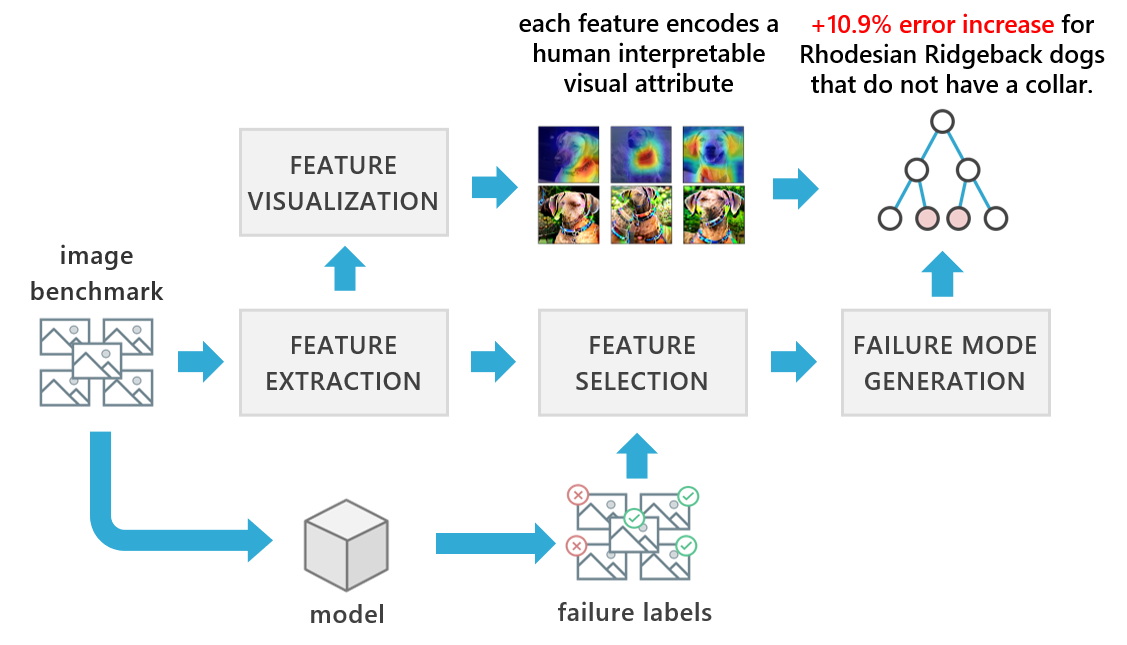}
\end{center}
\caption{\methodname\ error analysis workflow. \textit{Feature extraction} and \textit{visualization} are described in Section \ref{sec:feature_extraction_visualization}. \textit{Feature selection} and \textit{failure mode generation} in Section \ref{sec:feature_identification}.}
\label{fig:workflow}
\end{figure}

We seek to describe failures in a benchmark $S$ with low-dimensional rules using a reduced set of human-interpretable features. For example, for an image recognition model that detects traffic lights, we want to generate failure explanations, such as \emph{``The error rate for detecting traffic lights is 20\% higher when an image is captured in rainy weather and low light."}. The failure explanation in this case would be $\texttt{weather=rainy} \wedge \texttt{light\_intensity=low}$. Such rules slice the data into clusters for which we can report metrics as in Definitions~\ref{def:er} and \ref{def:ec}. Ideally, we would like to find clusters that jointly have large error rates and that cover a significant fraction of all errors in the benchmark.  These criteria ensure that the explanatory rules will be of sufficient importance and generality. Note that the purpose of these explanations is not to predict failure but rather to provide actionable guidance to engineers via a small set of rules about failures and their indicators. Figure \ref{fig:workflow} depicts the end-to-end \methodname\ workflow.

\section{Feature extraction and visualization}\label{sec:feature_extraction_visualization}
For each image in the set $S$, we first construct feature representation $F$ as a vector such that each element of the vector encodes some human-interpretable visual attribute. For this purpose, we use the penultimate layer (i.e., the layer adjacent to the logits layer) of a pretrained robust neural network to extract this feature vector. The robust model is trained via objective \eqref{eq:adv} for $l_{2}$ robustness.
\begin{align}
    \min_{\theta}\ \ \mathbb{E}_{(x,y) \sim \mathbb{D}} \left[ \max_{\|x' - x\|_{2} \leq \rho} \ell(h_{\theta}(x'), y) \right] \label{eq:adv}
\end{align}
Tsipras \etal \cite{tsipras2019robustness} show that for robust models, saliency maps are more interpretable than standard models and align well with human perception. Engstrom \etal \cite{Engstrom2019LearningPR} show that direct maximization of neurons of robust models is sufficient to generate recognizable visual attributes. Such output stands in contrast to the opacity of the visualizations of features generated from standard models for which, even with regularization, visualized features are rarely human interpretable~\cite{olah2017feature} (examples in Appendix Section \ref{sec:comparison_robust_nonrobust}). In Sections~\ref{sec:mturk_study} and~\ref{sec:MLEng_study}, we validate the earlier findings about human interpretability via performing studies with human subjects. 

Given the findings on the interpretability of robust models, for feature extraction, we use an adversarially trained Resnet-50 model \cite{He2015DeepRL} pretrained on ImageNet (using $\rho=3$). We note that we can perform feature extraction using a robust model even when the model under inspection is not robust. In this case, we can consider the extracted features as attributes of the data and not necessarily as part of the representation employed within the model. 

\begin{figure}[t]
\centering
\begin{subfigure}{0.9\linewidth}
\centering
\includegraphics[trim=0cm 0cm 15cm 0.9cm, clip, width=\linewidth]{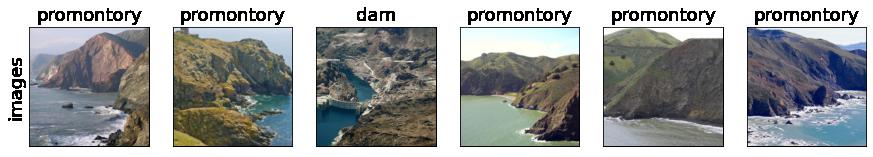}
\end{subfigure}\\
\begin{subfigure}{0.9\linewidth}
\centering
\includegraphics[trim=0cm 0cm 15cm 0.9cm, clip, width=\linewidth]{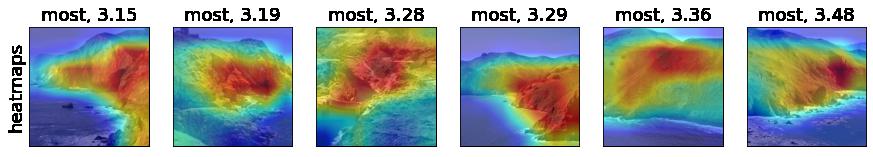}
\end{subfigure}
\caption{Illustration of heatmaps. From the most activating images (top row) for a neuron, it is not clear if the focus is on sky, water, or ground. Heatmaps (bottom row) resolve the ambiguity. The neuron appears to focus on ground.}
\label{fig:heatmaps_viz}
\end{figure}

We next describe our methodology for visualizing the neuron in the representation layer of a robust model:

\noindent\textbf{Most activating images:} A common approach to visualizing a neuron's sensitivity is to search through the given set of images to find the top-$k$ instances that maximally activate the desired neuron. A challenge with this approach is that it does not identify the specific attributes of images that are responsible for the activation~\cite{olah2017feature}. For example, consider the images in the top row of Figure \ref{fig:heatmaps_viz} where it is not clear whether the neuron is focused on sky, water or ground. 

\begin{figure}[t]
\centering
\begin{subfigure}{0.9\linewidth}
\centering
\includegraphics[trim=0cm 26.5cm 33cm 1.3cm, clip, width=\linewidth]{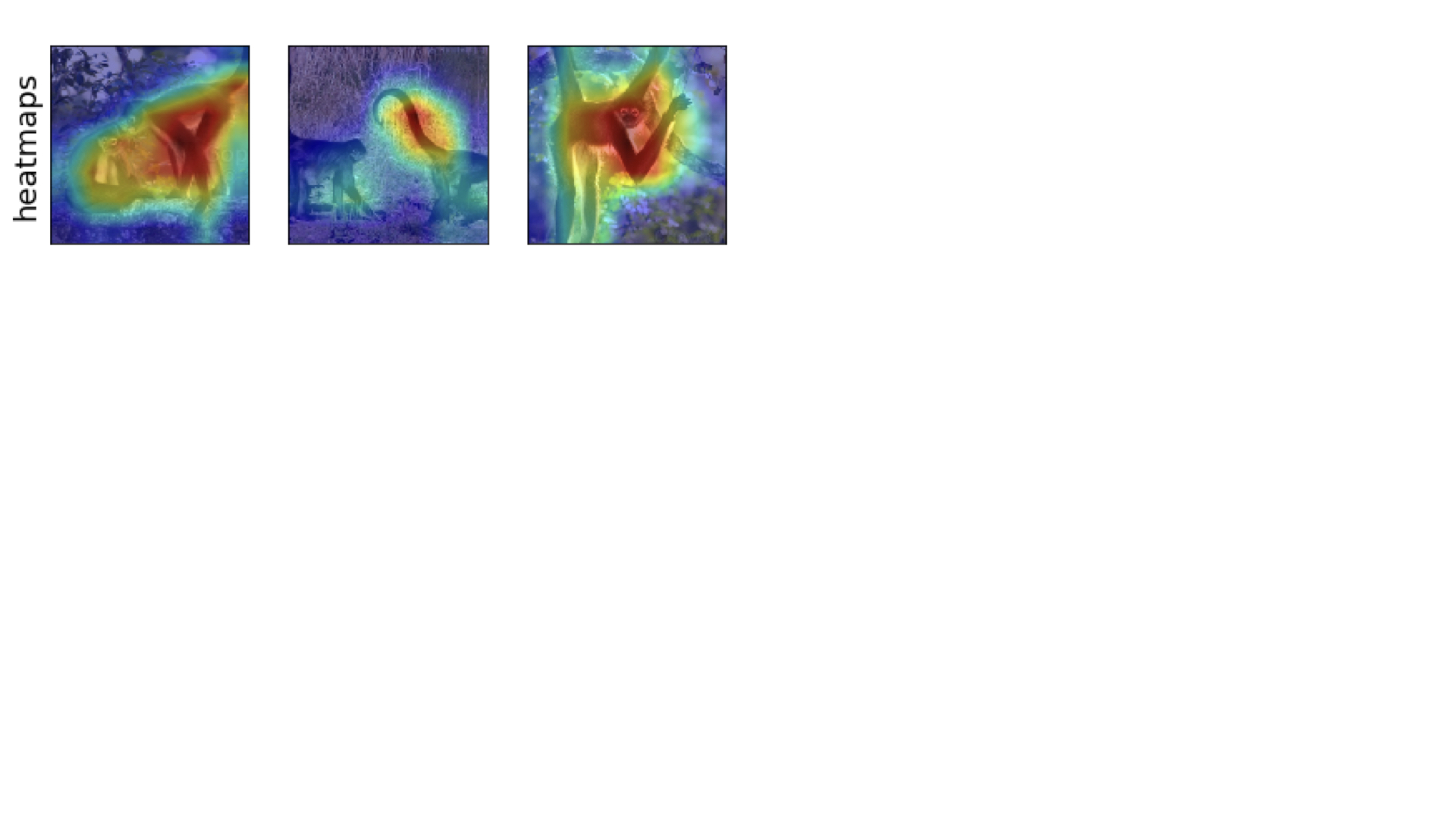}
\end{subfigure}\\
\begin{subfigure}{0.9\linewidth}
\centering
\includegraphics[trim=0cm 26.5cm 33cm 1.5cm, clip, width=\linewidth]{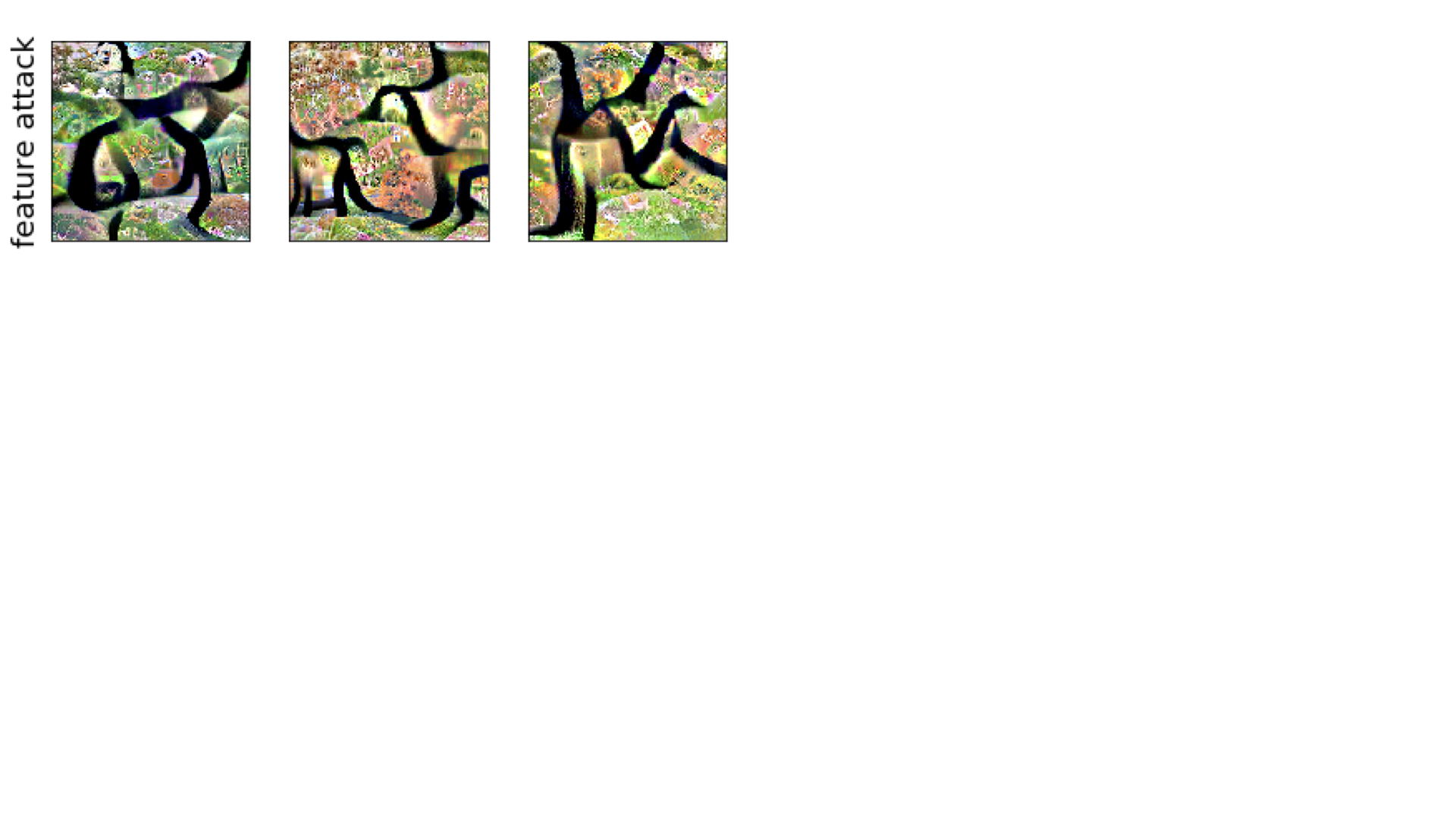}
\end{subfigure}
\caption{Illustration of feature attack. Heatmaps focus on different attributes in different images: body, tail, and face (left to right). Feature Attack (bottom row) resolves the ambiguity. The neuron appears to focus on the limbs and tail.}
\label{fig:attacks_viz}
\end{figure}

\noindent\textbf{Heatmaps:} To address the aforementioned problem, we propose to use heatmaps based on CAM~\cite{ZhouKLOT15} (second row in Figure \ref{fig:heatmaps_viz}), as an additional signal for visualization. For a Resnet-50 model, the representation layer is computed via a global average pooling operation to the tensor in the previous layer. We use the index of the desired neuron to retrieve the slice of the tensor from the previous layer at the same index (we refer to this as the \textit{feature map}). Next, we normalize the feature map between 0 and 1 and resize it to match the input size. Details are in Appendix Section \ref{sec:appendix_heatmap_generation}. Figure \ref{fig:heatmaps_viz} shows how heatmaps can help resolve the ambiguity. \\
\noindent\textbf{Feature Attack:} In some cases, heatmaps may not be sufficient to resolve ambiguity. For example, in Figure \ref{fig:attacks_viz}, it is not clear from the heatmaps whether the neuron is focused on the body, tail, or face of the monkey. Hence, we propose to maximize the neuron in the representation layer with respect to the original image (based on Engstrom \etal \cite{Engstrom2019LearningPR}). We observe that the limbs and tail (on top of green background) are amplified and this resolves the ambiguity. We call the resulting visualization, a \textit{feature attack}. More details are provided in Appendix Section \ref{sec:appendix_feature_attack}. \\
We use the top-$6$ most activating images, corresponding heatmaps, and feature attack images to generate visualizations of the desired feature (Examples in Appendix Section \ref{sec:appendix_mturk_examples}). We showcase the importance of these visualizations for interpretability via a study detailed in Section \ref{sec:mturk_study}.
\section{Failure mode generation}\label{sec:feature_identification}

We start with a set of features extracted from images using a robust model. Then, we explore the failure modes of a neural network (not necessarily the robust model) identified via consideration of the ground-truth labels. Our goal is to identify clusters with high error rates. In pursuit of failure analyses that can be understood with ease, we generate decision trees trained to predict failures (see Figure~\ref{fig:workflow}), determined by checking whether the model prediction is equal to the image label. 

We note that there are challenges with relying on decision trees for explanations. Large numbers of features can make finding a good split difficult due to the curse of dimensionality. From a usability perspective, failures for different classes can vary greatly and practitioners may be interested in understanding them separately~\cite{pandora1809}. Moreover, psychological studies provide evidence that people can hold in memory and understand only a limited number of chunks of information at the same time~\cite{miller1956magical,oberauer2012attention,cowan2001magical}. These challenges motivate designs for \methodname\ and other human-in-the-loop methodologies that avoid complex representations, such as large trees. Thus, we take the following steps:\\
\noindent \textbf{Class-based failure explanations}: We focus analyses and results on two types of class-based groupings: (i) \textit{prediction groupings} and (ii) \textit{label groupings}. For example, \textit{prediction groupings} for the class ``goldfish'', contain all images for which the model under inspection predicts ``goldfish''. A failure analysis for this group enables us to understand false positives.  \textit{Label groupings} for the class ``goldfish'', contain all images for which the ground truth in ImageNet is ``goldfish'' regardless of the model's prediction. A failure analysis for this class provides insights about false negatives.  

\noindent \textbf{Feature selection}: To reduce dimensionality, we compute the mutual information between each feature and failure labels in the group, and select the top-$20$ features, sorted by mutual information values. Mutual information estimates the information gained via the feature on the failure label variable in addition to its prior (i.e., the base error rate). Using mutual information for feature selection aligns with our goal of building decision trees with the features, as the same metric is used to estimate the value of  splits in the tree. \\
\noindent \textbf{Truncation and rule generation}: All trees are truncated to a low depth ($\leq 3$). We choose to explain failure modes by using paths in the tree as summary rules that have an error coverage higher than a given threshold $\tau$ and an error rate of at least $\baseerrorrate + \delta$. Both $\delta$ and $\tau$ are adjustable. 
This procedure is described in Appendix Algorithm \ref{alg:feature_id}.

\subsection{Automatic evaluation of decision tree}
\label{sec:evaluation_metric}
Traditional metrics for evaluating the quality of a decision tree for a classification problem, such as accuracy, precision, and recall, are insufficient when the model is used for description and explanation. Ideally we would want to consider failure modes that include all possible failures in a benchmark. This goal is challenging because of (i) incompleteness in the feature set, (ii) difficulties in finding failure modes that generalize well for many examples at the same time, and, most importantly, (iii) certain failures may happen very rarely in the benchmark. Therefore, we focus on the explanatory properties of failure analyses. 

\begin{figure}[t]
\centering
\includegraphics[width=\linewidth]{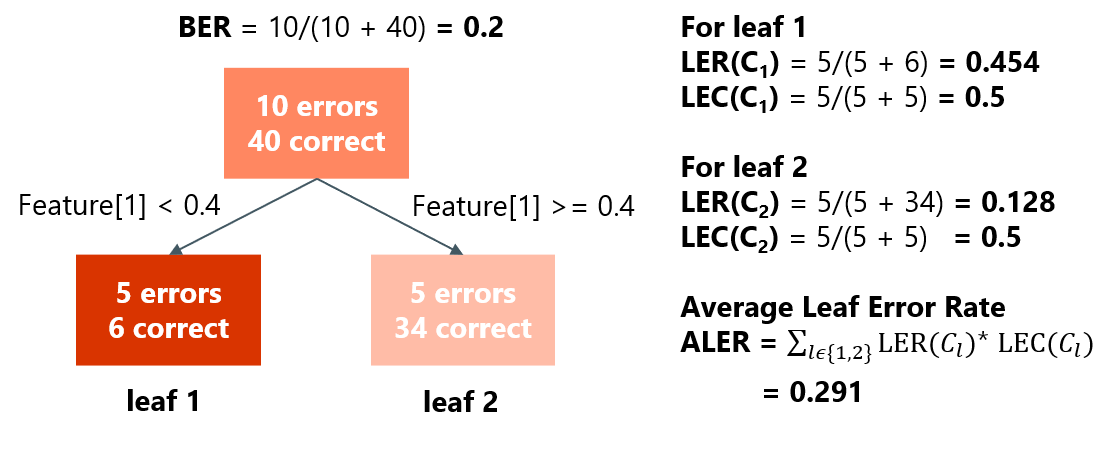}
\caption{Illustration of $\avgtree$ metric. For both leaf nodes, precision and recall are 0 since $\errorrate < 0.5$ however leaf $1$ is more important for failure mode discovery.}
\label{fig:decision_tree}
\end{figure}

For a given leaf node $\leaf$, we can use definitions \ref{def:er} and \ref{def:ec} to define its \textit{leaf error rate} $\errorrate(\cluster_\leaf$),  and \textit{leaf error coverage} $\errorcoverage(\cluster_\leaf$). $\cluster_\leaf$ denotes the cluster of data that falls into leaf $\leaf$. For a decision tree $T$, we then compute the following metric as the \textit{average leaf error rate} ($\avgtree$).

\begin{definition}{$\avgtree$ of a tree $\tree$ is the average error rate across all leaves weighed by the respective error coverage.}
\label{def:leaf_average_precision}
$$ \avgtree(\tree) = \sum_{\leaf \in\ leaves(\tree)} \errorrate({\cluster_\leaf}) \times \errorcoverage({\cluster_\leaf}) $$
\end{definition}
Per the definition of the leaf error coverage, we have $\sum_{\leaf \in\ leaves(\tree)} \errorrate(\cluster_\leaf) = 1$ because all failure instances will be covered by exactly one leaf. 

As an example, consider the simple $1$-depth decision tree given in Figure~\ref{fig:decision_tree}. Since the leaf error rate is less than $0.5$ for both leaf nodes, tree precision and tree recall are both zero. However, since the error rate of leaf $1$ is significantly higher than the base error rate ($0.454 >> 0.2$), leaf $1$ is important as a failure mode description (and more predictive than leaf $2$), which is exactly the notion we seek to capture for describing high concentrations of error. 

The key property that makes $\avgtree$ useful is that if it is equal to some quantity $q$, then there is at least one leaf in the decision tree with leaf error rate greater than $q$: 
\begin{proposition}\label{thm:main_theorem}
	For a tree $\tree$ with $\avgtree(\tree) = q$, there exists at least one leaf $\leaf$, with leaf error rate $\errorrate(\cluster_\leaf) \geq q$.
\end{proposition}
The proposition follows from the fact that all weights (i.e., leaf error coverage) are less than or equal to 1. For the decision tree in Figure \ref{fig:decision_tree}, $\baseerrorrate = 0.291$, which is greater than the base error rate $0.2$ by a margin of $0.091$. This signals the presence of a leaf with error rate of at least $0.291$ which we know is leaf $1$, $\errorrate(\cluster_1)=0.454$.

Since the root node of the tree already comes with a prior on the error rate ($\baseerrorrate$), we are interested in how much more value the discovered failure modes in the tree add when compared to the root. Thus, for the automated evaluation we use $\avgtree - \baseerrorrate$ to measure the increase in error and discrepancy that the tree explains.

This metric also suggests that leaves with higher value of $\errorrate({\cluster_\leaf}) \times \errorcoverage({\cluster_\leaf})$ contribute more to $\avgtree$ and are thus more important for explaining failures. This leads to the following metric for ranking nodes for failure explanation:

\begin{definition}
\label{def:importance_value}
{The Importance Value i.e $\importance(\cluster_\leaf)$ of a leaf node $l$ in a tree $\tree$ is defined as: }
\begin{align*}
\importance(\cluster_\leaf) = \errorrate({\cluster_\leaf}) \times \errorcoverage({\cluster_\leaf})
\end{align*}

\end{definition}
\section{Failure modes discovered by \methodname}\label{sec:failure_modes_maintext}
We now describe some failure modes discovered by \methodname\ when analyzing a standard
Resnet-50 model, using a robust Resnet-50 model for feature extraction (both trained on ImageNet). We use the ImageNet training set (instead of the validation set) for failure analysis due to the larger number of image instances per class (1300 vs. 50 in the validation set). Note that, from a methodology perspective, practitioners may apply the \methodname\ workflow to any benchmark with ample data and failures, but the failure modes that they discover may vary depending on the nature of the benchmark. For ease of exposition, all decision trees have depth one. We selected the leaf node with highest $\importance(\cluster_{\leaf})$ and visualized the feature responsible for the split. More examples of failure modes are in Appendix Section  \ref{sec:failure_modes_standard} (using a standard model) and Section  \ref{sec:failure_modes_robust} (using a robust model). 
\begin{figure}[t]
\centering
\begin{subfigure}{0.9\linewidth}
\centering
\includegraphics[trim=0cm 0cm 15cm 0.9cm, clip, width=\linewidth]{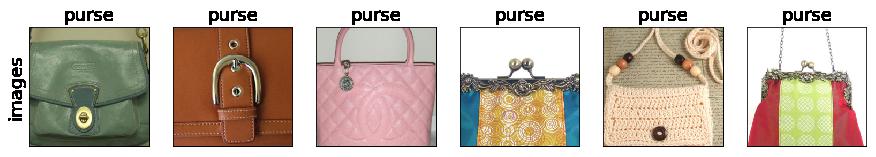}
\end{subfigure}\\
\begin{subfigure}{0.9\linewidth}
\centering
\includegraphics[trim=0cm 0cm 15cm 0.9cm, clip, width=\linewidth]{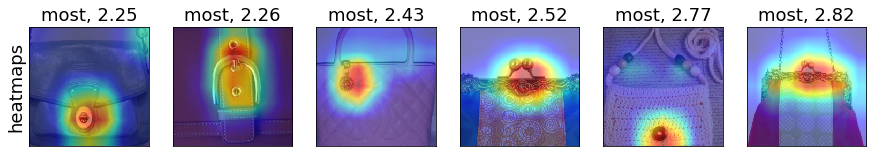}
\end{subfigure}\\
\begin{subfigure}{0.9\linewidth}
\centering
\includegraphics[trim=0cm 0cm 15cm 0.9cm, clip, width=\linewidth]{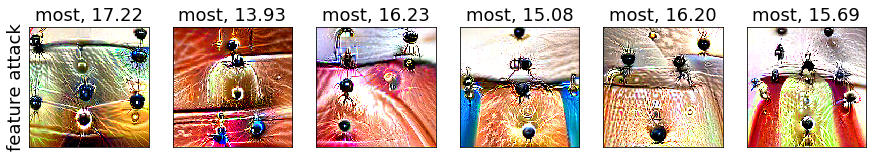}
\end{subfigure}
\begin{subfigure}{0.9\linewidth}
\centering
\includegraphics[trim=0cm 0cm 15cm 0.9cm, clip, width=\linewidth]{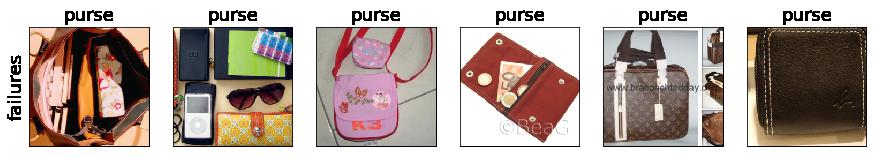}
\end{subfigure}
\caption{For images with \textbf{label purse}, $\baseerrorrate$ = $0.3085$. When feature$[1456] < 0.3641$, error rate increases to $0.4179$ (\textbf{\textcolor{red}{+10.94\%}}) and a fraction of $0.6409$ of all failures are in this node. $\avgtree$ = $0.3433$.}
\label{fig:label_purse}
\end{figure}

\noindent\textbf{Label grouping}: 
In Figure \ref{fig:label_purse}, the top row shows the most highly activating images for a feature identified important for failure. The heatmap and feature attack provide strong evidence that the feature is paying attention to the buckle on purses. The bottom row shows randomly selected failures in this leaf node. We do not observe a buckle in any of these images even though they have ground-truth labels of purse indicating the feature (while correlated, not causal) is important for making correct predictions. In other words, it appears that, whenever the purse image does not include a buckle, it is more difficult for the model to predict it correctly as a purse (error increases by 10.94\%) and the prediction is more likely to be a false negative. 

\noindent\textbf{Prediction grouping}: 
\begin{figure}[t]
\centering
\begin{subfigure}{0.9\linewidth}
\centering
\includegraphics[trim=0.5cm 0cm 15cm 0.9cm, clip, width=\linewidth]{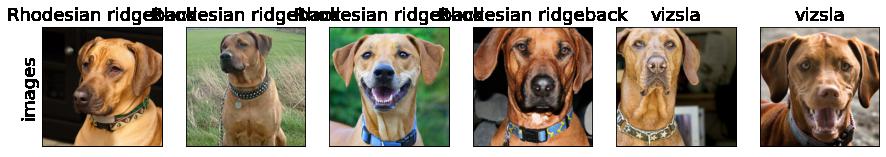}
\end{subfigure}\\
\begin{subfigure}{0.9\linewidth}
\centering
\includegraphics[trim=0cm 0cm 15cm 0.9cm, clip, width=\linewidth]{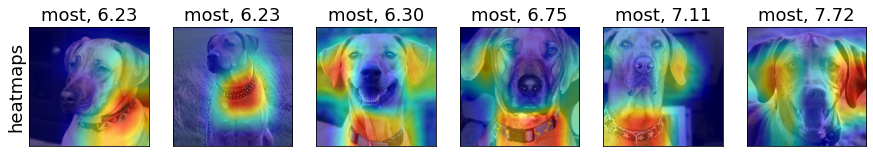}
\end{subfigure}\\
\begin{subfigure}{0.9\linewidth}
\centering
\includegraphics[trim=0cm 0cm 15cm 0.9cm, clip, width=\linewidth]{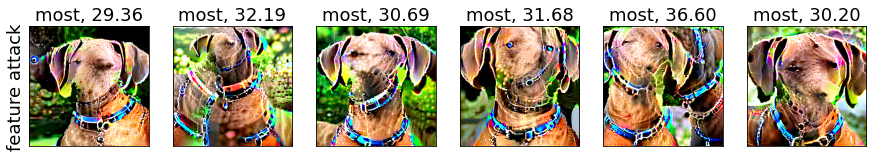}
\end{subfigure}
\begin{subfigure}{0.9\linewidth}
\centering
\includegraphics[trim=0cm 0cm 15cm 0.9cm, clip, width=\linewidth]{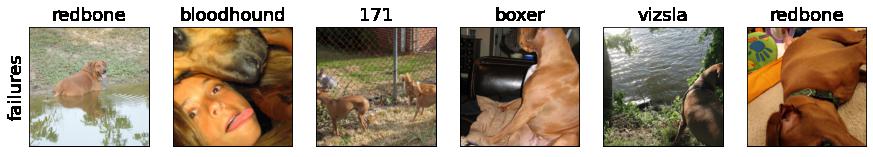}
\end{subfigure}
\caption{For images with \textbf{prediction Rhodesian Ridgeback}, $\baseerrorrate$ = $0.2093$. When feature$[1634] < 1.3779$, error rate increases to $0.3184$ (\textbf{\textcolor{red}{+10.91\%}}) and a fraction of $0.4561$ of all failures are in this node. $\avgtree$ = $0.2337$.}
\label{fig:pred_rhodesian}
\end{figure}
Figure \ref{fig:pred_rhodesian} displays an example where
the heatmap and feature attack provide strong evidence that
the feature indicates a dog collar. In the bottom row (i.e randomly selected failure images), we do not observe a dog collar in any of these images even though the model predicts a Rhodesian Ridgeback for all of them. This indicates that the feature (correlated, not causal) is important for making correct predictions. In other words, whenever the model predicts Rhodesian Ridgeback, and the image does not contain a dog collar, the prediction is more likely to be a false positive (error increases by 10.91\%). 


\section{Experiments with automated evaluation}
\begin{figure}[h]
\centering
\begin{subfigure}{0.49\linewidth}
\centering
\includegraphics[trim=0cm 0cm 0cm 0cm, clip, width=\linewidth]{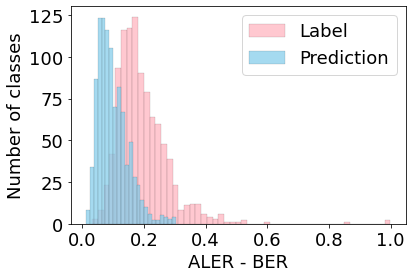}
\caption{Standard model}
\end{subfigure}
\begin{subfigure}{0.49\linewidth}
\centering
\includegraphics[trim=0cm 0cm 0cm 0cm, clip, width=\linewidth]{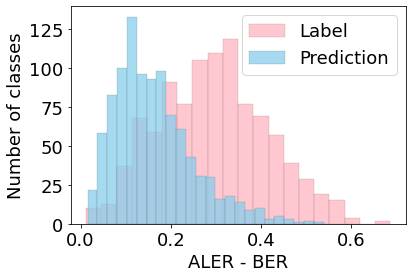}
\caption{Robust model}
\end{subfigure}
\caption{Comparison between grouping strategies using a decision tree of depth 3.}
\label{fig:group_comparison}
\end{figure}

We now report on a study of factors that influence the effectiveness of error analysis: decision tree depth, robustness of model, and grouping strategy. We train decision trees with depths of $1$ and $3$ for each model and grouping strategy. For evaluating a decision tree, we use the metric $\avgtree - \baseerrorrate$ as defined in Section \ref{sec:evaluation_metric}. We also select the leaf with highest importance value $\importance(\cluster_\leaf)$ for each decision tree and evaluate whether the cluster of data in this leaf satisfies the two conditions: $\errorrate(\cluster_\leaf) > \baseerrorrate + \delta$ and $\errorcoverage(\cluster_\leaf) > \tau$, with $\delta=0.1$ and $\tau=0.2$. In Appendix Table \ref{table:appendix_fraction_table}, we report for each model, grouping strategy, and tree depth the fraction of such \textit{valid leaves} across all $1000$ classes that satisfy these conditions.

In summary, we make the following observations: 
\begin{itemize}[leftmargin=7pt]
   \setlength{\itemsep}{2.0pt}
   \setlength{\parskip}{-2pt}
   \setlength{\parsep}{-2pt}
    \item Grouping by ground-truth labels results in better decision trees (by $\avgtree - \baseerrorrate$ score) as compared to prediction grouping for both standard and robust models (Figure \ref{fig:group_comparison}), and also for decision trees with different depths. 
    \item Failure explanation for a robust model results in significantly better score compared to the standard model for both grouping strategies and depths of decision tree (See Appendix Figures \ref{fig:appendix_depth_one_tree}, \ref{fig:appendix_depth_three_tree}).
    \item Increasing depth significantly improves the score for both models as shown in Appendix Figures \ref{fig:appendix_depth_comparison_nonrobust} and \ref{fig:appendix_depth_comparison_robust}.
    \item For the standard model, on all trees (except prediction grouping and depth=$1$ where it is $0.211$) the fraction of classes with at least one valid leaf is at least $0.596$. For the robust model, on all trees the fraction of classes with at least one valid leaf is at least $0.787$.
\end{itemize}

\section{Crowd study on feature interpretability}
\label{sec:mturk_study}

To understand the effectiveness of feature visualizations, we conducted a crowdsourcing study using Amazon Mechanical Turk. For both grouping strategies (prediction and label), we selected the top, middle, and bottom 20 classes (total 60 x 2) based on the error rate of the robust Resnet-50 model and selected the top 10 features with the highest mutual information on failure, resulting in 1200 visualizations and 5971 human answers. In total, we had 312 unique workers, each completing an average of 19.14 tasks. All visualizations were evaluated by five workers, except 29 for which we only acquired four assessments. Workers were paid \$0.5 per hit, with an average salary of $\$$12 per hour. Anonymized data from this study is available in the associated code repository. 

For each visualization, workers were shown three sections: A (most activating images), B (heatmaps), and C (feature attack), and asked to answer the set of questions in Figure \ref{fig:aws_questions} in the Appendix Section~\ref{sec:appendix_mturk_examples}. The questions were designed with two goals: (1) to collect human-generated feature descriptions, and (2) to evaluate the ease with which workers can understand and describe the features.

\begin{figure}[t]
\centering
\begin{subfigure}{0.47\linewidth}
\centering
\includegraphics[trim=0.2cm 0cm 0cm 0cm, clip, width=\linewidth]{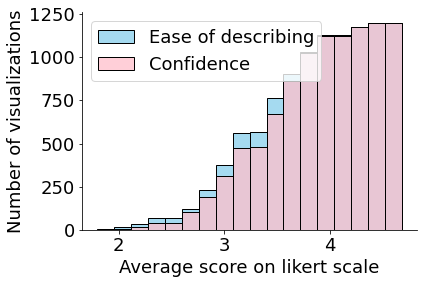}
\caption{Questions 3 and 4}
\label{subfig:conf_easy_cumhist}
\end{subfigure}
\begin{subfigure}{0.47\linewidth}
\centering
\includegraphics[trim=0.2cm 0cm 0cm 0cm, clip, width=\linewidth]{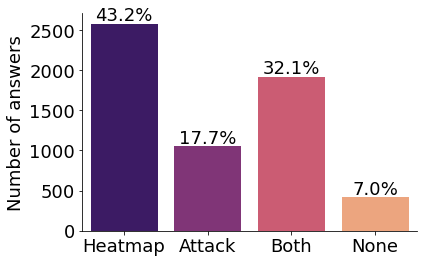}
\caption{Question 5}
\label{subfig:countplot}
\end{subfigure}
\caption{Cumulative distribution of answers to Questions 3, 4, 5 in the crowd study.}
\label{fig:amt_study_plots}
\end{figure}

Figure \ref{subfig:conf_easy_cumhist} shows the cumulative distribution of answers to Questions 3 (Ease of understanding) and 4 (Confidence). We observe that workers are able to describe visualizations with confidence and ease (average likert scale $>$ 3 for both) for a large number of features ($>$ 800). Figure \ref{subfig:countplot} shows workers' preferences of heatmaps versus feature attacks (Question 5). In response to a question about the most useful views, 43.18\% answers (out of 5971) report Section B (heatmap) as most useful, 17.67\% report Section C (feature attack), and 32.14\% report both Section B and C. Only 7\% of the answers reported None. These results provide evidence that heatmaps are most valuable in contributing to the understanding of the meaning of features, and that feature attack visualizations are also valuable individually and together with heatmaps (examples in Sections~\ref{sec:examples_feature_attack_important} and \ref{sec:examples_both_important}). Importantly, only 0.92\% of all 1200 features received None as the majority vote from all 5 workers. This provides further evidence that the methodology is effective in explaining a large number of features.  Note that since crowd workers do not routinely visualize and debug models using such visualizations, our results are likely to be a lower bound on the true effectiveness of the visualizations for engineers.

For further details, Appendix Section \ref{sec:appendix_mturk_examples} shows an analysis on the agreement scores between textual descriptions of different workers as well as concrete examples that workers found easy or difficult to describe.

\section{Study with machine learning practitioners}
\label{sec:MLEng_study}
To evaluate the usefulness of \methodname\ for error analysis and debugging, we conducted user studies with 14 ML practitioners at Microsoft. Participants were recruited via four mailing lists on the topics of ``Machine Learning'' and ``Computer Vision''. Participation requirements included previous experience in applying machine learning on vision tasks. A summary of roles and years of experience is shown in Table~\ref{tab:user_study} in the Appendix Section~\ref{sec:app_user_study}.

\noindent\textbf{Study Protocol:} Each study lasted one hour and started with a description of the \methodname\ workflow and terminology as shown in Figure~\ref{fig:workflow}. We asked participants to imagine facing a situation as follows: 

\noindent\myquote{We will conduct error analysis for a classification model (robust Resnet-50) trained on ImageNet. We ask you to imagine that you are part of a team that will deploy this in production and wants to understand where the model is incorrect and identify action items upon failures.} 

All participants inspected two groupings. The first grouping was randomly assigned from a set of five pre-selected groupings where the most important features for failure explanation were rated as ``easy to describe'' by crowd workers to facilitate onboarding (exact assignment in Table~\ref{tab:user_study_conditions_1}, Appendix Section~\ref{sec:app_user_study}). The second grouping was selected randomly. In each case, participants were presented with a decision tree (of depth 2) describing the failure modes. For each node, they could see the  error rate and coverage, the instances in the node, and the visualization of the feature responsible for the split. To collect feedback on their experience, we asked the following questions: 

\begin{figure}[t]
\centering
\includegraphics[trim=0.5cm 0cm 0.5cm 2cm, clip,width=0.9\linewidth]{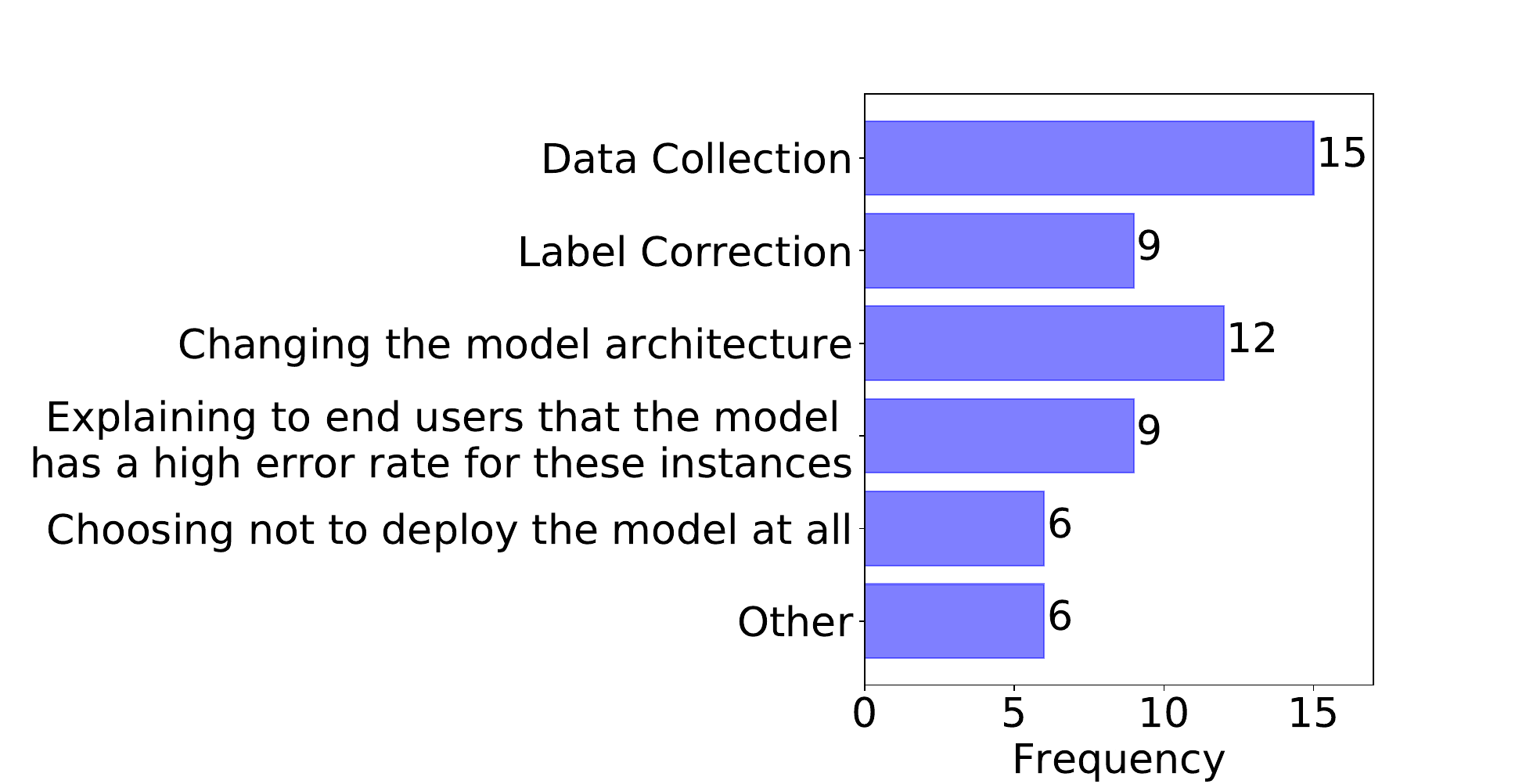}
\caption{Answers of practitioners on identifying action items for mitigating failures discovered by \methodname.}
\label{fig:action_items}
\end{figure}

\noindent\textbf{Q1}: \emph{Can you describe the cluster of images  defined by this feature?}
All participants were able to describe the features of at least one of the two groupings they inspected. 10 out of 14 participants were able to describe features of both groupings. The four participants who could not describe the features of one grouping faced one or both of the following challenges: the feature represented two or more concepts (e.g., sky and wires), or participants had a difficult time understanding the sensitivity of the feature's visual appearance  to its activation value. In such cases, they preferred to see more than five examples for refining their hypothesis. 


\noindent\textbf{Q2}: \emph{Is the feature necessary for the task or do you think it is a spurious correlation?}
Participants were able to identify several surprising spurious correlations (e.g., \myquote{This feature looks like a hair detector, should not be used for Seat Belts.}- P12, \myquote{Seems like the model can do well only when the photos of Water Jugs are taken either professionally or on a clean background.}- P17). Similarly, they could identify when the failure modes were related to necessary but not sufficient features (e.g., \myquote{The model is focusing on the face of the dog. We might also need to make the model look at the legs, hair length, hair color, etc.}- P7). 

\noindent\textbf{Q3}: \emph{What action items would you take for mitigating the errors for this class?} Overall, participants reported that the identified failure explanations gave them ideas about how to continue with mitigating the errors, given sufficient time and resources. Data collection was the most popular action item (Figure~\ref{fig:action_items}) either for mitigating lack of data diversity or addressing statistical biases (e.g. \myquote{I would check if most Tiger Cats in the training data have green background and if so, add more diverse photos.}- P5). Changes in model architecture were most often related to increasing the capacity of the model (e.g., \myquote{The model is using the same features for detecting zebra patterns and spatulas, perhaps it's best to use more than one feature for this.}- P5). 
\begin{figure}[t]
\centering
\includegraphics[trim=3cm 1cm 4.0cm 2cm, width=1\linewidth]{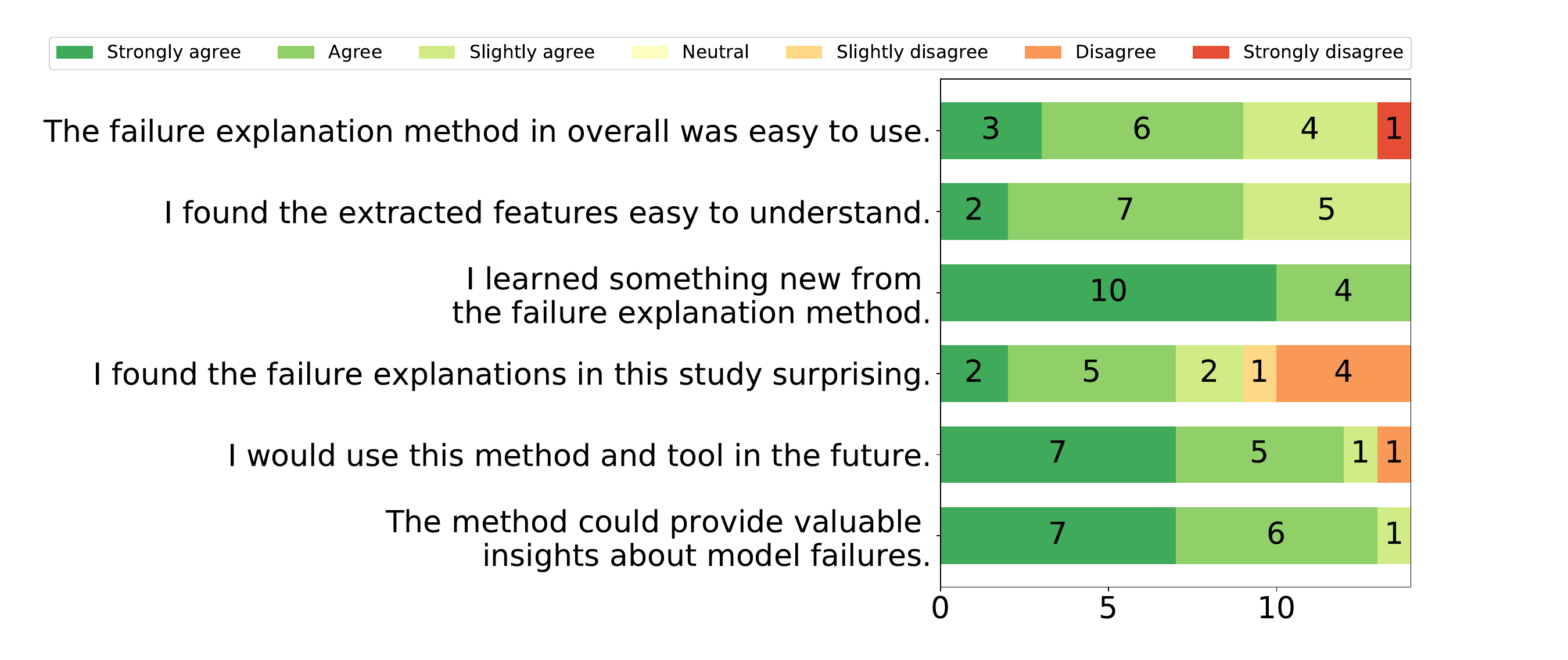}
\caption{Agreement scores of participants on general evaluation of \methodname.}
\label{fig:agreement_scores}
\end{figure}

\noindent\textbf{User satisfaction scores}: The study ended with a survey on whether participants agreed with six statements, completed in private (Figure~\ref{fig:agreement_scores}). We observe overall enthusiasm about the usefulness of the workflow and its ability to surface important failures (e.g. \myquote{The approach can help make DNNs more explainable, making such analysis mandatory before deploying a solution. A step further could be to take a similar test with humans to ensure the model has learned explainable features, before we ship the model.}- P7). 

\section{Conclusion and future work}
\label{sec:conclusion}
We described proposed methods and a workflow aimed at providing insights about critical failures of machine learning models by using features extracted from robust representations. Beyond developing and refining the approach, we performed two sets of studies with human subjects, focusing on the interpretability of features and on the usefulness of the error analysis in realistic settings. The studies provide evidence that practitioners can effectively leverage the methods to interpret the features and to identify most significant clusters in data with errors due to issues, such as systematic spurious correlations. We believe that the error analyses and visualizations embodied in Barlow show promise for supporting cycles of iterative improvement that can identify and mitigate failures without requiring expensive manual metadata collection.

Directions for refining Barlow include extending the range of actions suggested automatically to practitioners and providing richer interactive capabilities. Interactive explorations can include hypothesis testing via the consideration of counterfactuals that make specific, understandable changes to the data and models. Another enabling direction centers on extending the current correlational analyses to causal studies. Extensions that establish causal influences on failure modes will be valuable for debugging non-robust models with robust features. To fairly evaluate counterfactual analyses, it is necessary to establish notions of causal accuracy that reward a model only when it is accurate for the right reasons, as reducing model access to features that are highly predictive but spurious (e.g. food for detecting plates) often will decrease standard measures of accuracy.



\noindent \textbf{Acknowledgments}:
We thank Parham Mohadjer for helping with our studies with human subjects, Nathan Cross for helpful discussions on extending \methodname\ to radiology images, and Soheil Feizi for providing useful feedback on the paper and throughout the project. We thank all participants in our studies; the feedback received helped us to refine the methods. Finally, we acknowledge the support provided by Arda Deniz Eden, our newest team member.

{\small
\bibliographystyle{ieee_fullname}
\bibliography{egbib}
}
\clearpage

\noindent {\bf \LARGE Appendix}

\appendix

\section{Feature extraction}\label{sec:appendix_feature_extraction}
\begin{figure}[t]
\centering
\includegraphics[trim=0cm 9cm 2cm 0.8cm, clip, width=0.9\linewidth]{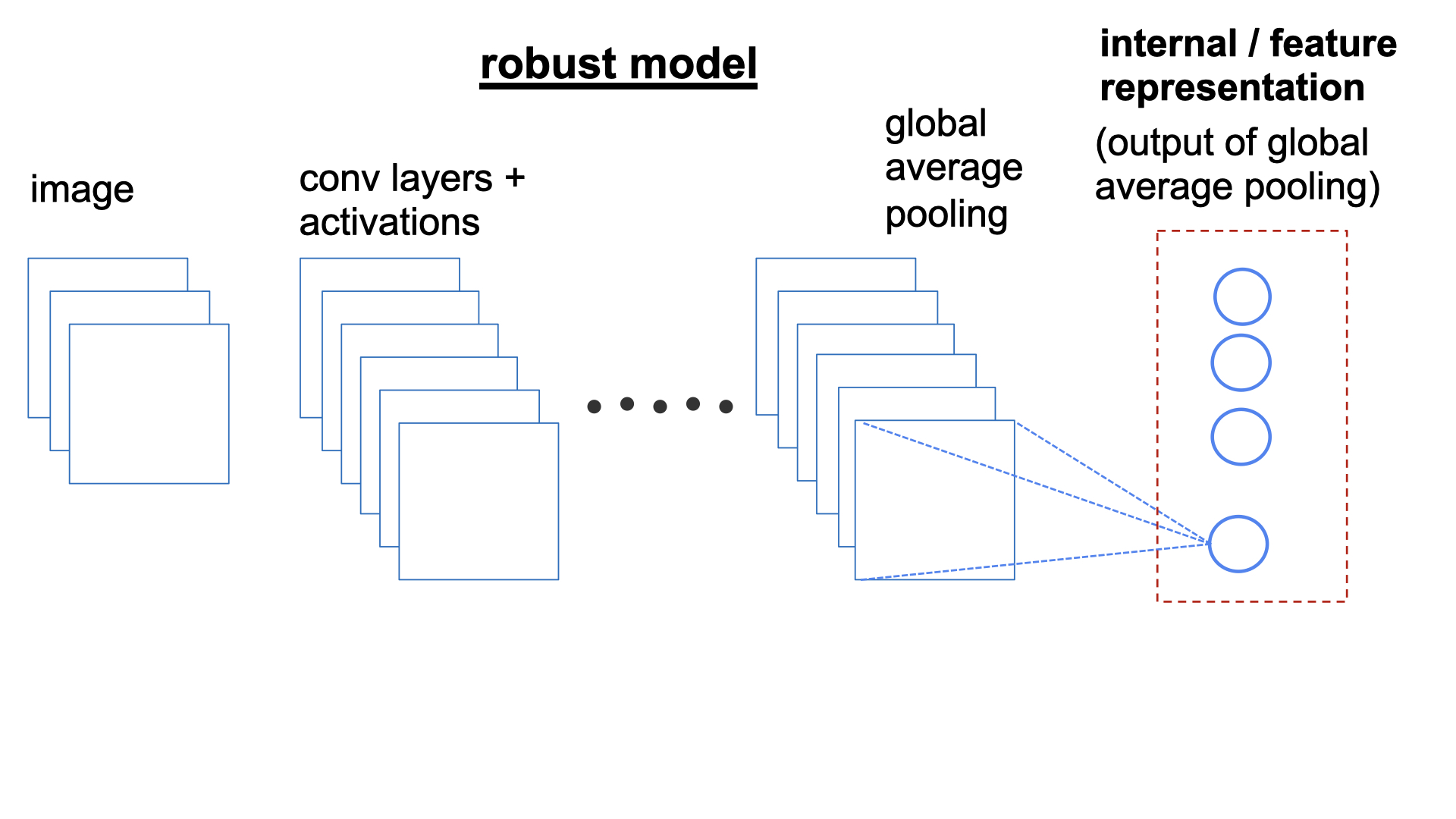}
\caption{Feature extraction.}
\label{fig:feature_extraction}
\end{figure}
Figure \ref{fig:feature_extraction} shows our feature extraction mechanism. We use an adversarially trained Resnet-50 model (threat model is an $l_{2}$ ball of radius $3$). For feature extraction, we use the penultimate layer i.e layer adjacent to the logits layer (also the output of global average pooling layer for a Resnet-50 architecture). In practice, in order to extract these features for a given benchmark, we run each image in the benchmark through the model (in inference time) and use the activation in this layer as feature values.

\section{Heatmap generation}\label{sec:appendix_heatmap_generation}
\begin{figure}[t]
\centering
\includegraphics[trim=0cm 3cm 2cm 0.8cm, clip, width=0.9\linewidth]{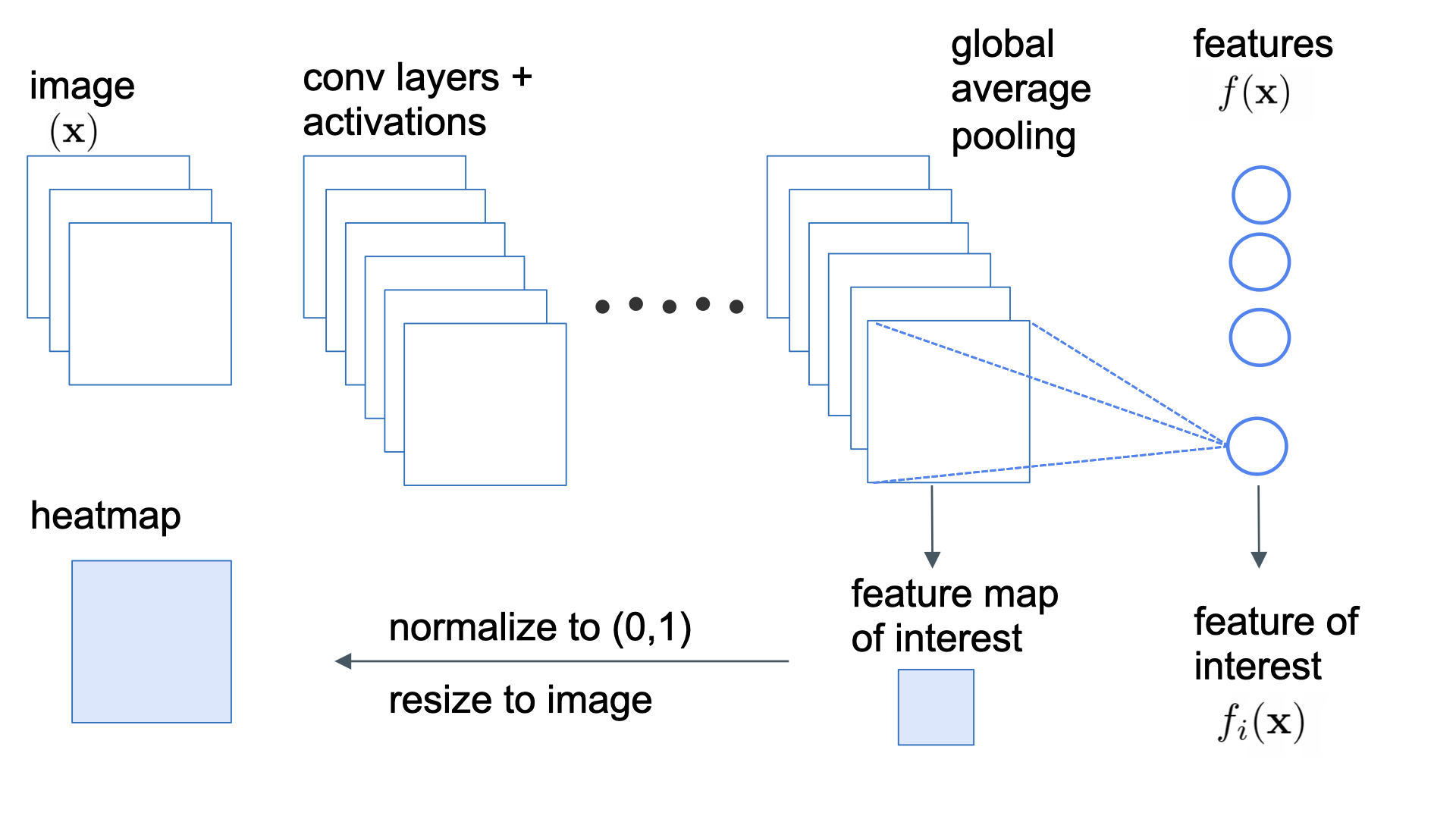}
\caption{Heatmap generation.}
\label{fig:heatmap_generation}
\end{figure}
Figure \ref{fig:heatmap_generation} describes the heatmap generation method. We select the feature map from the output of the tensor of the previous layer (i.e before applying the global average pooling operation). Next, we normalize the feature map between 0 and 1 and resize the feature map to match the image size.

\section{Feature attack}\label{sec:appendix_feature_attack}
\begin{figure}[t]
\centering
\includegraphics[trim=0cm 1cm 2cm 0.8cm, clip, width=0.9\linewidth]{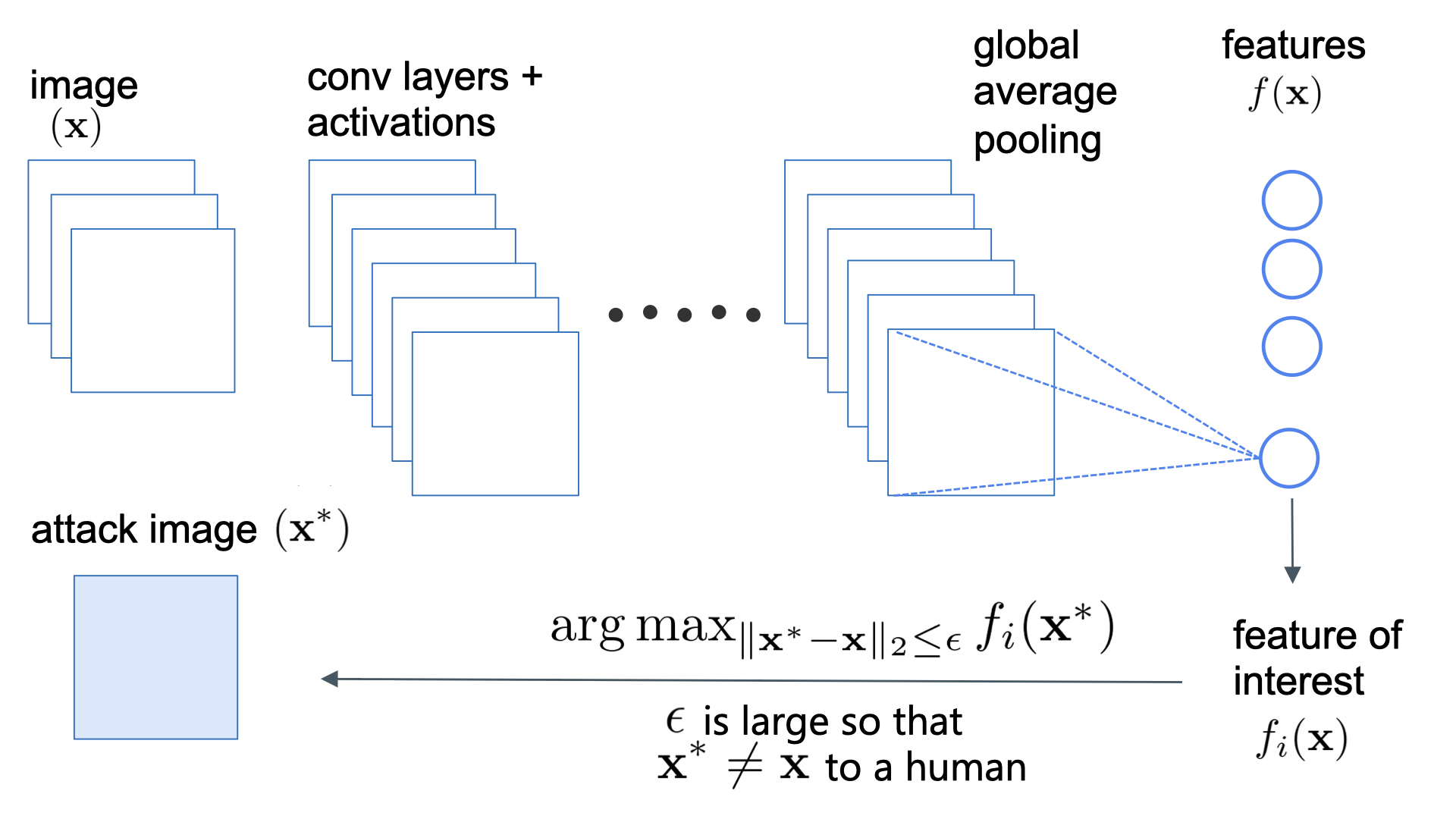}
\caption{Feature attack generation.}
\label{fig:feature_attack_algorithm}
\end{figure}
In Figure \ref{fig:feature_attack_algorithm}, we illustrate the process behind the feature attack. We select the feature we are interested in and optimize the image to maximize its value to generate the visualization. $\epsilon$ is a hyperparameter used to control the amount of change allowed in the image. For optimization, we use gradient ascent with step size = $1$, $\epsilon = 500$ and number of iterations = $500$. 

\newpage
\section{Failure mode generation}
We describe our procedure for generating failure modes in Algorithm \ref{alg:feature_id}. The algorithm can take as an input any cluster of image data $\cluster$. In our experiments, the clusters were defined via image grouping by label and model prediction. However, practitioners may choose to apply the same procedure to clusters of images defined in other ways such as for example pairs of classes that are often confused with each other or unions of prediction and label groupings for the same class.

\begin{algorithm}
\SetAlgoLined
\SetKwInput{KwInput}{Input}
\SetKwInput{KwOutput}{Output}
\SetKwInput{KwReturn}{Return}
\DontPrintSemicolon
  
\KwInput{features: $F$, model: $h$, image cluster: $\cluster$, \\ 
number\_of\_features: $k$, tree\_parameters: $A$,\\
error\_rate\_threshold: $\delta$, error\_coverage\_threshold: $\tau$}
\KwOutput{leaves with high error concentration: $L$}
$L = \emptyset$

$\baseerrorrate =  \errorrate(\cluster)$

$E(x)= \begin{dcases}
        0 & h(x) = y \\
        1 & h(x) \ne y 
        \end{dcases}\quad \forall (x,y) \in \cluster $

$F^*=\emptyset$

\While{$|F^*| < k$}
{
    $F^*=F^* \cup \argmax_{f \in F \setminus F^*} \mathrm{IG}(E;f)$
}

$T$ = train\_decision\_tree($F^*$, $E$, $A$)

\For{$\leaf \in \tree$}
{
    $\mathrm{\textbf{If}}{(\errorrate(\cluster_\leaf) > \baseerrorrate + \delta)\ \mathrm{and}\ (\errorcoverage(\cluster_\leaf) > \tau)}$
    {
        $L = L \cup \{\leaf\}$
    }
}
\KwReturn{$L$}
list of leaves from decision tree $T$ with error rate of at least $\baseerrorrate + \delta$ and error coverage at least $\tau$.
\caption{Failure mode generation procedure.}
\label{alg:feature_id}
\end{algorithm}

\section{Automatic evaluation of decision tree}\label{sec:automatic_eval}
We now report on a study of factors that influence the effectiveness of error analysis: decision tree depth, robustness of model, grouping strategy. We train decision trees with depths of $1$ and $3$ for each model and grouping strategy. For evaluating a decision tree, we use the metric $\avgtree - \baseerrorrate$ as defined in Maintext Section \ref{sec:evaluation_metric}, Definition \ref{def:leaf_average_precision}. We also select the leaf with highest importance value $\importance(\cluster_\leaf)$ for each decision tree (Maintext Definition \ref{def:importance_value}) and evaluate whether the cluster of data in this leaf satisfies the two conditions: $\errorrate(\cluster_\leaf) > \baseerrorrate + \delta$ and $\errorcoverage(\cluster_\leaf) > \tau$, with $\delta=0.1$ and $\tau=0.2$. In Table \ref{table:appendix_fraction_table}, we report for each model, grouping strategy, and tree depth the fraction of such \textit{valid leaves} across all $1000$ classes that satisfy these conditions.

\begin{figure}[t]
\centering
\begin{subfigure}{0.45\linewidth}
\centering
\includegraphics[trim=0cm 0cm 0cm 0cm, clip, width=\linewidth]{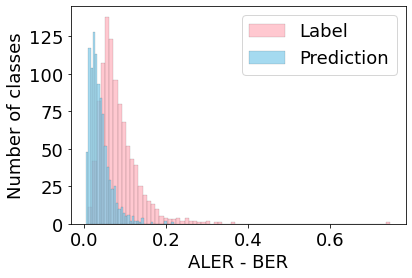}
\end{subfigure}
\begin{subfigure}{0.45\linewidth}
\centering
\includegraphics[trim=0cm 0cm 0cm 0cm, clip, width=\linewidth]{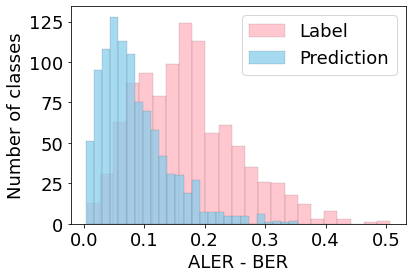}
\end{subfigure}\\
\begin{subfigure}{0.45\linewidth}
\centering
\includegraphics[trim=0cm 0cm 0cm 0cm, clip, width=\linewidth]{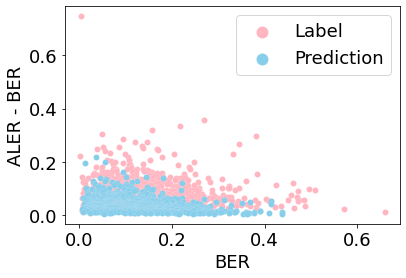}
\caption{Standard model}
\end{subfigure}
\begin{subfigure}{0.45\linewidth}
\centering
\includegraphics[trim=0cm 0cm 0cm 0cm, clip, width=\linewidth]{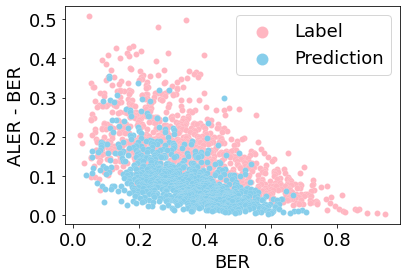}
\caption{Robust model}
\end{subfigure}
\caption{Comparison between grouping strategies using a decision tree of depth 1.}
\label{fig:group_comparison_depth_one}
\end{figure}

\begin{figure}[t]
\centering
\begin{subfigure}{0.45\linewidth}
\centering
\includegraphics[trim=0cm 0cm 0cm 0cm, clip, width=\linewidth]{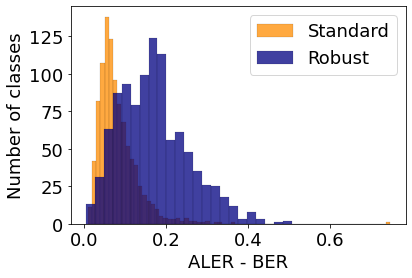}
\end{subfigure}
\begin{subfigure}{0.45\linewidth}
\centering
\includegraphics[trim=0cm 0cm 0cm 0cm, clip, width=\linewidth]{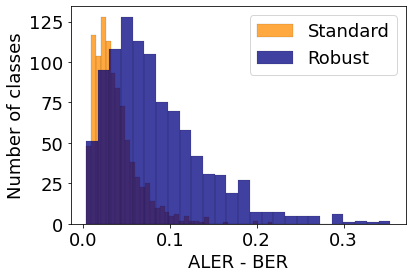}
\end{subfigure}\\
\begin{subfigure}{0.45\linewidth}
\centering
\includegraphics[trim=0cm 0cm 0cm 0cm, clip, width=\linewidth]{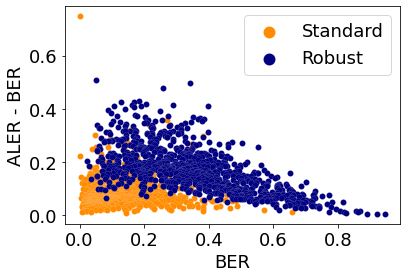}
\caption{Label grouping}
\end{subfigure}
\begin{subfigure}{0.45\linewidth}
\centering
\includegraphics[trim=0cm 0cm 0cm 0cm, clip, width=\linewidth]{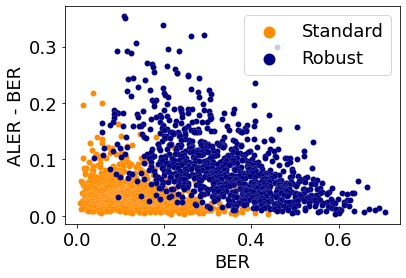}
\caption{Prediction grouping}
\end{subfigure}
\caption{Comparison between standard and robust models using a decision tree of depth 1.}
\label{fig:appendix_depth_one_tree}
\end{figure}

\begin{figure}[t]
\centering
\begin{subfigure}{0.45\linewidth}
\centering
\includegraphics[trim=0cm 0cm 0cm 0cm, clip, width=\linewidth]{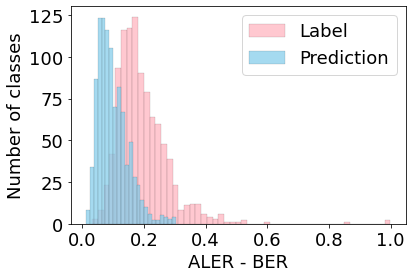}
\end{subfigure}
\begin{subfigure}{0.45\linewidth}
\centering
\includegraphics[trim=0cm 0cm 0cm 0cm, clip, width=\linewidth]{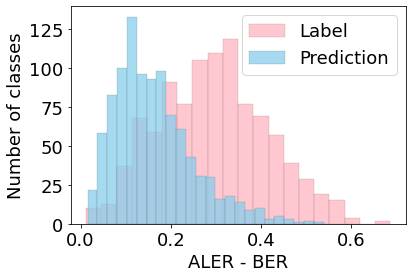}
\end{subfigure}\\
\begin{subfigure}{0.45\linewidth}
\centering
\includegraphics[trim=0cm 0cm 0cm 0cm, clip, width=\linewidth]{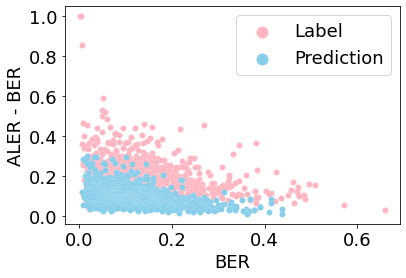}
\caption{Standard model}
\end{subfigure}
\begin{subfigure}{0.45\linewidth}
\centering
\includegraphics[trim=0cm 0cm 0cm 0cm, clip, width=\linewidth]{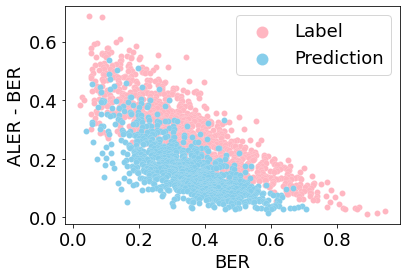}
\caption{Robust model}
\end{subfigure}
\caption{Comparison between grouping strategies using a decision tree of depth 3.}
\label{fig:group_comparison_depth_three}
\end{figure}

\begin{figure}[t]
\centering
\begin{subfigure}{0.45\linewidth}
\centering
\includegraphics[trim=0cm 0cm 0cm 0cm, clip, width=\linewidth]{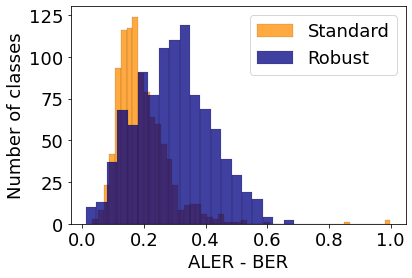}
\end{subfigure}
\begin{subfigure}{0.45\linewidth}
\centering
\includegraphics[trim=0cm 0cm 0cm 0cm, clip, width=\linewidth]{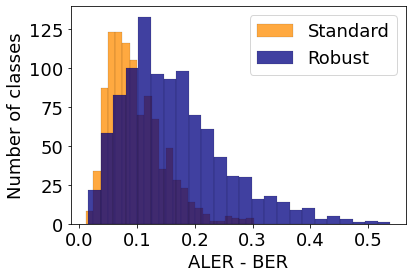}
\end{subfigure}\\
\begin{subfigure}{0.45\linewidth}
\centering
\includegraphics[trim=0cm 0cm 0cm 0cm, clip, width=\linewidth]{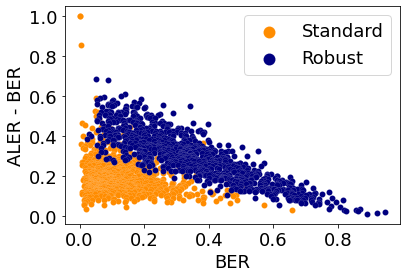}
\caption{Label grouping}
\end{subfigure}
\begin{subfigure}{0.45\linewidth}
\centering
\includegraphics[trim=0cm 0cm 0cm 0cm, clip, width=\linewidth]{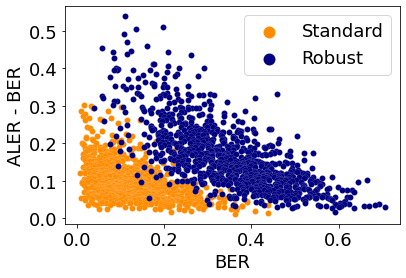}
\caption{Prediction grouping}
\end{subfigure}
\caption{Comparison between standard and robust models using a decision tree of depth 3.}
\label{fig:appendix_depth_three_tree}
\end{figure}

We make the following observations: 
\begin{itemize}[leftmargin=7pt]
   \setlength{\itemsep}{2.0pt}
   \setlength{\parskip}{-2pt}
   \setlength{\parsep}{-2pt}
\item Grouping by ground-truth labels results in better decision trees (by $\avgtree - \baseerrorrate$ score) compared to prediction grouping for both standard and robust models and also for decision trees with different depths. This is true even when $\baseerrorrate$ is similar (See Figures \ref{fig:group_comparison_depth_one} and \ref{fig:group_comparison_depth_three}). \item Failure explanation for a robust model results in significantly better score compared to standard model for both grouping strategies and depths of decision tree. This is again true, even when $\baseerrorrate$ is similar (See Figures \ref{fig:appendix_depth_one_tree} and \ref{fig:appendix_depth_three_tree}). While this observation is intuitive, given that that the extracted features come from the robust model, it serves as an additional motivation for employing robust models in practice. The evaluation shows that such models might simplify the debugging and error analysis processes.
\end{itemize}

\begin{table}[t]
\centering
\begin{tabular}{| c | c | c | c | }
\hline
\textbf{Model} & \textbf{Depth} & \textbf{Grouping} & \textbf{Fraction}\\
\hline
\hline
Standard & 1 & Label & 0.596 \\
\hline
Standard & 1 & Prediction & 0.211 \\
\hline
Standard & 3 & Label & 0.900 \\
\hline
Standard & 3 & Prediction & 0.649 \\
\hline
Robust & 1 & Label & 0.977 \\
\hline
Robust & 1 & Prediction & 0.787 \\
\hline
Robust & 3 & Label & 0.899 \\
\hline
Robust & 3 & Prediction & 0.804 \\
\hline
\end{tabular}
\caption{For each model, grouping strategy and decision tree depth we report the fraction of \textit{valid leaves} across all $1000$ classes, i.e the leaf nodes that satisfy these two conditions: $\errorrate(\cluster_\leaf) > \baseerrorrate + \delta$ and $\errorcoverage(\cluster_\leaf) > \tau$, with $\delta=0.1$ and $\tau=0.2$ in the last column. Semantically, these would be all leaves with an error increase of at least 10\% that cover 20\% of the failures or more.}
\label{table:appendix_fraction_table}
\end{table}
\clearpage

\newpage

\begin{figure}[t]
\centering
\begin{subfigure}{0.45\linewidth}
\centering
\includegraphics[trim=0cm 0cm 0cm 0cm, clip, width=\linewidth]{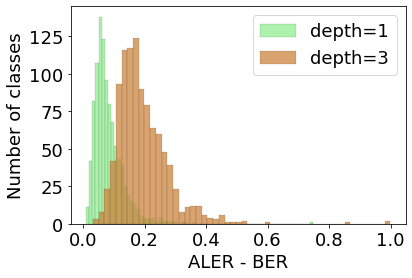}
\end{subfigure}
\begin{subfigure}{0.45\linewidth}
\centering
\includegraphics[trim=0cm 0cm 0cm 0cm, clip, width=\linewidth]{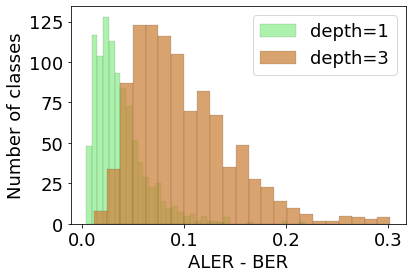}
\end{subfigure}\\
\begin{subfigure}{0.45\linewidth}
\centering
\includegraphics[trim=0cm 0cm 0cm 0cm, clip, width=\linewidth]{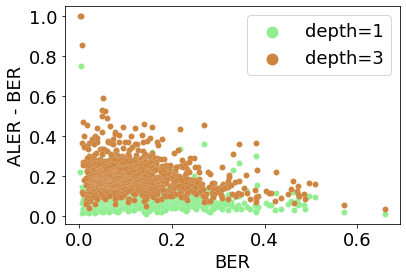}
\caption{Label grouping}
\end{subfigure}
\begin{subfigure}{0.45\linewidth}
\centering
\includegraphics[trim=0cm 0cm 0cm 0cm, clip, width=\linewidth]{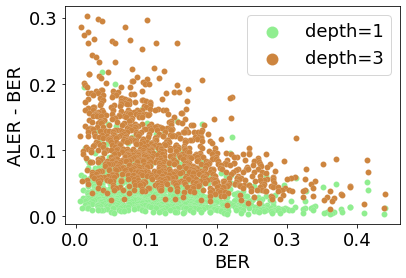}
\caption{Prediction grouping}
\end{subfigure}
\caption{Comparison between decision trees of different depths using a standard model.}
\label{fig:appendix_depth_comparison_nonrobust}
\end{figure}

\begin{figure}[t]
\centering
\begin{subfigure}{0.45\linewidth}
\centering
\includegraphics[trim=0cm 0cm 0cm 0cm, clip, width=\linewidth]{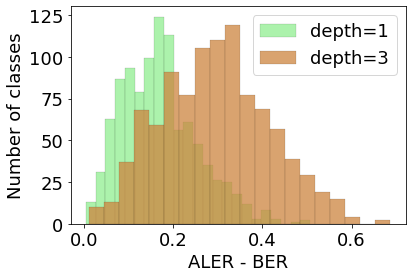}
\end{subfigure}
\begin{subfigure}{0.45\linewidth}
\centering
\includegraphics[trim=0cm 0cm 0cm 0cm, clip, width=\linewidth]{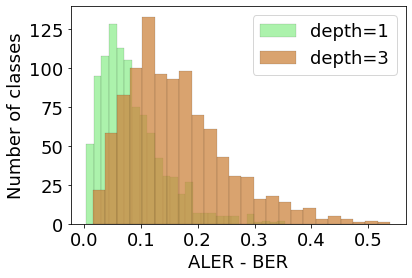}
\end{subfigure}\\
\begin{subfigure}{0.45\linewidth}
\centering
\includegraphics[trim=0cm 0cm 0cm 0cm, clip, width=\linewidth]{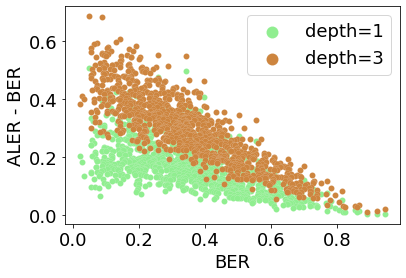}
\caption{Label grouping}
\end{subfigure}
\begin{subfigure}{0.45\linewidth}
\centering
\includegraphics[trim=0cm 0cm 0cm 0cm, clip, width=\linewidth]{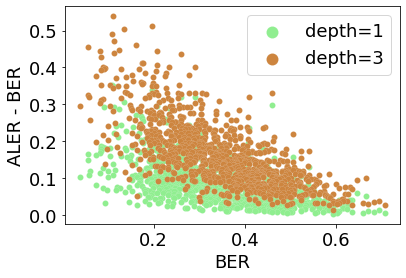}
\caption{Prediction grouping}
\end{subfigure}
\caption{Comparison between decision trees of different depths using a robust model.}
\label{fig:appendix_depth_comparison_robust}
\end{figure}

\clearpage
\section{Failure modes discovered by \methodname}\label{sec:failure_modes_large}
In this section, we describe several failure modes discovered by \methodname. For experiments in subsection \ref{sec:failure_modes_standard}, we analyze the errors of a standard Resnet-50 model for failure analysis and for subsection \ref{sec:failure_modes_robust}, we inspect a robust Resnet-50 model. In both cases, we use a robust Resnet-50 model for feature extraction. All models were pretrained on ImageNet. We use the ImageNet training set (instead of the validation set) for failure analysis due to the larger number of instances and failures. For ease of exposition, all decision trees have depth one. We select the leaf node with highest Importance Value (i.e $\importance$ as defined in Definition \ref{def:importance_value}) for visualizing the failure mode. Since the tree has depth one, we can visualize the one feature that defines this leaf node. 

All feature visualizations are organized as follows. The topmost row shows the most activating images. The second row shows the heatmaps The third row shows feature attack images. Finally, the bottom row shows randomly selected failure examples in the leaf node. 

For all tables, $\baseerrorrate$ denotes the Base Error Rate, $\errorrate$ denotes Error Rate, $\errorcoverage$ denotes Error Coverage for the leaf with highest Importance Value and $\avgtree$ denotes Average Leaf Error Rate.

\subsection{Failure explanation for a standard model}\label{sec:failure_modes_standard}
\subsubsection{Grouping by label}
Results are in Table \ref{table:standard_label_grouping}.

\begin{figure*}[t]
\centering
\begin{subfigure}{0.9\linewidth}
\centering
\includegraphics[trim=0cm 0cm 0cm 0.9cm, clip, width=\linewidth]{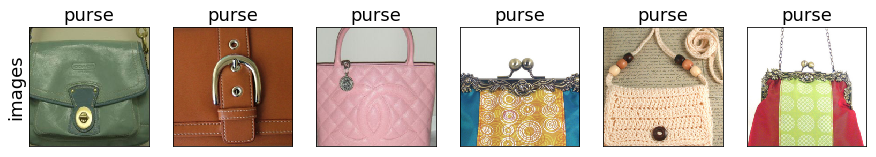}
\end{subfigure}\\
\begin{subfigure}{0.9\linewidth}
\centering
\includegraphics[trim=0cm 0cm 0cm 0.9cm, clip, width=\linewidth]{viz/purse_heatmaps}
\end{subfigure}\\
\begin{subfigure}{0.9\linewidth}
\centering
\includegraphics[trim=0cm 0cm 0cm 0.9cm, clip, width=\linewidth]{viz/purse_attacks}
\end{subfigure}\\
\begin{subfigure}{0.9\linewidth}
\centering
\includegraphics[trim=0cm 0cm 0cm 0.9cm, clip, width=\linewidth]{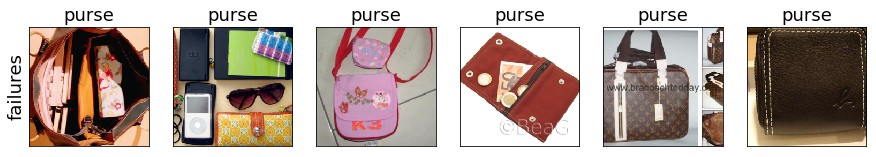}
\end{subfigure}
\caption{Visualization of feature[1456]. For images with \textbf{label purse}, when feature$[1456] < 0.3641$, error rate increases to $0.4179$ ( \textcolor{red}{\textbf{+10.94\%}}).}
\label{fig:appendix_label_purse}
\end{figure*}

\begin{table*}
\centering
\begin{tabular}{| p{1.7cm} || p{1cm} | p{1.5cm} | p{1cm} | p{1cm} | p{1cm} | p{1cm} | p{1.9cm} | p{3cm} |}
\hline
\textbf{Class name} & \textbf{Feature index} & \textbf{Decision rule} & \textbf{\baseerrorratetext} & \textbf{\errorratetext} & \textbf{\errorcoveragetext} & \textbf{\avgtreetext} &\textbf{Feature} \newline \textbf{visualization} & \textbf{Feature name\newline (from visualization)}\\
\hline
purse & $1456$ & $< 0.3641$ & $0.3085$ & $0.4179$ & $0.6409$ & $0.3433$ & Figure \ref{fig:appendix_label_purse} & buckle \\
\hline
monastery & $995$ & $< 0.1428$ & $0.3861$ & $0.6345$ & $0.3543$ & $0.4301$ & Figure \ref{fig:appendix_label_monastery} & greenery \\
\hline
maillot & $1364$ & $> 0.7066$ & $0.6592$ & $0.7564$ & $0.4819$ & $0.6696$ & Figure \ref{fig:appendix_label_maillot} & water \\
\hline
monitor & $1679$ & $< 0.8030$ & $0.4731$ & $0.6061$ & $0.7431$ & $0.5247$ & Figure \ref{fig:appendix_label_monitor} & black rectangles \\
\hline
tiger cat & $544$ & $< 0.2036$ & $0.4969$ & $0.8754$ & $0.4458$ & $0.5946$ & Figure \ref{fig:appendix_label_tiger_cat} & face close up \\
\hline
titi & $1911$ & $< 0.7329$ & $0.4131$ & $0.5240$ & $0.8138$ & $0.4664$ & Figure \ref{fig:appendix_label_titi} & brown color,\newline green background \\
\hline
lotion & 776 & $< 0.3313$ & $0.3624$ & $0.4797$ & $0.6920$ & $0.4040$ & Figure \ref{fig:appendix_label_lotion} & fluffy cream\newline color/texture \\
\hline
pitcher & 1378 & $< 0.7671$ & $0.3438$ & $0.6253$ & $0.5526$ & $0.4444 $ & Figure \ref{fig:appendix_label_pitcher} & handle \\
\hline
hog & 1611 & $< 0.0578$ & $0.3315$ & $0.6842$ & $0.7842$ & $0.5615 $ & Figure \ref{fig:appendix_label_hog} & pinkish animal \\
\hline
trench coat & 1264 & $< 0.6915$ & $0.1339$ & $0.3196$ & $0.8227$ & $0.2693 $ & Figure \ref{fig:appendix_label_trench_coat} & light color coat \\
\hline
baseball & 1081 & $< 0.5461$ & $0.1069$ & $0.3034$ & $0.9712$ & $0.2948 $ & Figure \ref{fig:appendix_label_baseball} & baseball stitch\newline pattern \\
\hline
\end{tabular}
\caption{Results on a standard Resnet-50 model using grouping by label.}
\label{table:standard_label_grouping}
\end{table*}

\begin{figure*}[t]
\centering
\begin{subfigure}{0.9\linewidth}
\centering
\includegraphics[trim=0cm 0cm 0cm 0.9cm, clip, width=\linewidth]{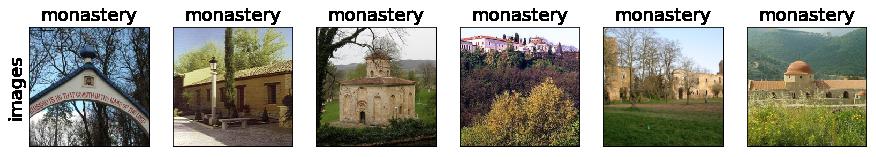}
\end{subfigure}\\
\begin{subfigure}{0.9\linewidth}
\centering
\includegraphics[trim=0cm 0cm 0cm 0.9cm, clip, width=\linewidth]{viz/monastery_heatmaps}
\end{subfigure}\\
\begin{subfigure}{0.9\linewidth}
\centering
\includegraphics[trim=0cm 0cm 0cm 0.9cm, clip, width=\linewidth]{viz/monastery_attacks}
\end{subfigure}\\
\begin{subfigure}{0.9\linewidth}
\centering
\includegraphics[trim=0cm 0cm 0cm 0.9cm, clip, width=\linewidth]{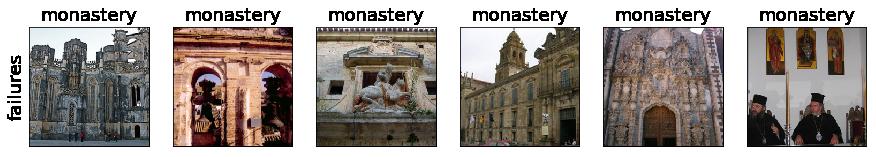}
\end{subfigure}
\caption{Visualization of feature$[995]$. For images with \textbf{label monastery}, when feature$[995] < 0.1428$, error rate increases to $0.6345$ (\textcolor{red}{\textbf{+24.84\%}}).}
\label{fig:appendix_label_monastery}
\end{figure*}

\begin{figure*}[t]
\centering
\begin{subfigure}{0.9\linewidth}
\centering
\includegraphics[trim=0cm 0cm 0cm 0.9cm, clip, width=\linewidth]{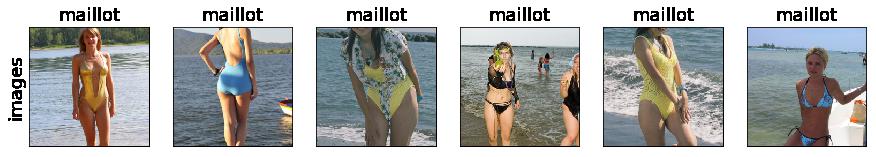}
\end{subfigure}\\
\begin{subfigure}{0.9\linewidth}
\centering
\includegraphics[trim=0cm 0cm 0cm 0.9cm, clip, width=\linewidth]{viz/maillot_heatmaps}
\end{subfigure}\\
\begin{subfigure}{0.9\linewidth}
\centering
\includegraphics[trim=0cm 0cm 0cm 0.9cm, clip, width=\linewidth]{viz/maillot_attacks}
\end{subfigure}\\
\begin{subfigure}{0.9\linewidth}
\centering
\includegraphics[trim=0cm 0cm 0cm 0.9cm, clip, width=\linewidth]{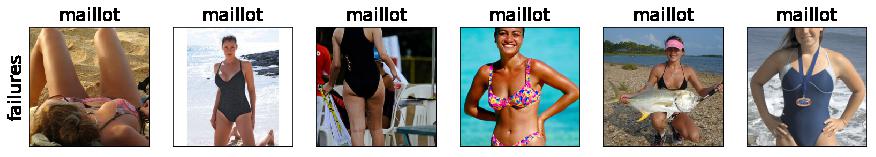}
\end{subfigure}
\caption{Visualization of feature[1365]. For images with \textbf{label maillot}, when feature$[1365] > 0.7066$, error rate increases to $0.7564$ (\textcolor{red}{\textbf{+9.72\%}}).}
\label{fig:appendix_label_maillot}
\end{figure*}

\begin{figure*}[t]
\centering
\begin{subfigure}{0.9\linewidth}
\centering
\includegraphics[trim=0cm 0cm 0cm 0.9cm, clip, width=\linewidth]{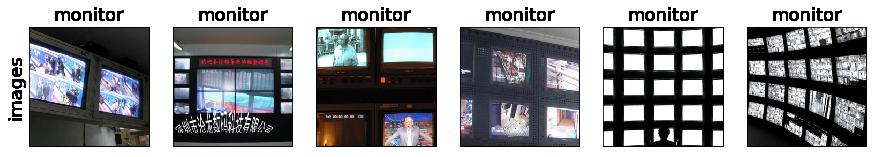}
\end{subfigure}\\
\begin{subfigure}{0.9\linewidth}
\centering
\includegraphics[trim=0cm 0cm 0cm 0.9cm, clip, width=\linewidth]{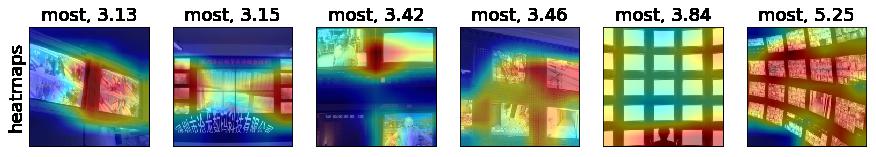}
\end{subfigure}\\
\begin{subfigure}{0.9\linewidth}
\centering
\includegraphics[trim=0cm 0cm 0cm 0.9cm, clip, width=\linewidth]{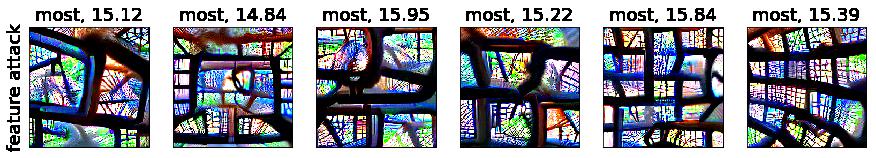}
\end{subfigure}\\
\begin{subfigure}{0.9\linewidth}
\centering
\includegraphics[trim=0cm 0cm 0cm 0.9cm, clip, width=\linewidth]{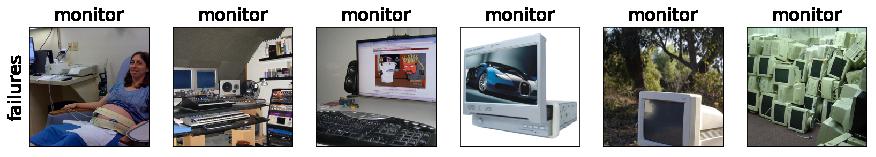}
\end{subfigure}
\caption{Visualization of feature[1679]. For images with \textbf{label monitor}, when feature$[1679] < 0.8030$, error rate increases to $0.6061$ (\textcolor{red}{\textbf{+13.00\%}}).}
\label{fig:appendix_label_monitor}
\end{figure*}

\begin{figure*}[t]
\centering
\begin{subfigure}{0.9\linewidth}
\centering
\includegraphics[trim=0cm 0cm 0cm 0.9cm, clip, width=\linewidth]{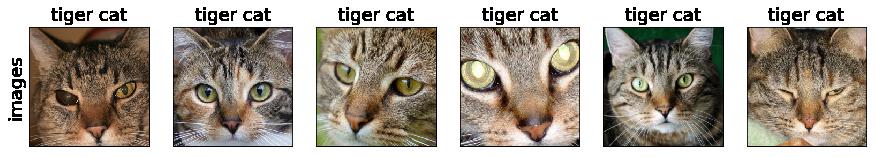}
\end{subfigure}\\
\begin{subfigure}{0.9\linewidth}
\centering
\includegraphics[trim=0cm 0cm 0cm 0.9cm, clip, width=\linewidth]{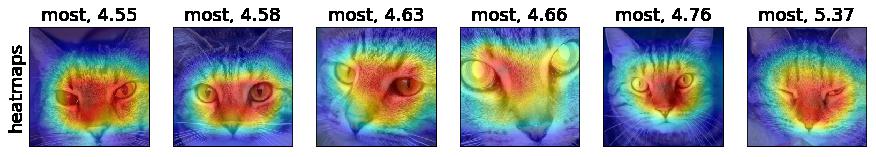}
\end{subfigure}\\
\begin{subfigure}{0.9\linewidth}
\centering
\includegraphics[trim=0cm 0cm 0cm 0.9cm, clip, width=\linewidth]{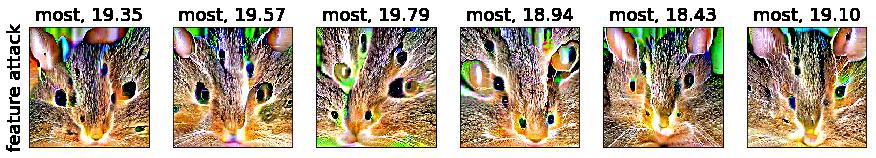}
\end{subfigure}\\
\begin{subfigure}{0.9\linewidth}
\centering
\includegraphics[trim=0cm 0cm 0cm 0.9cm, clip, width=\linewidth]{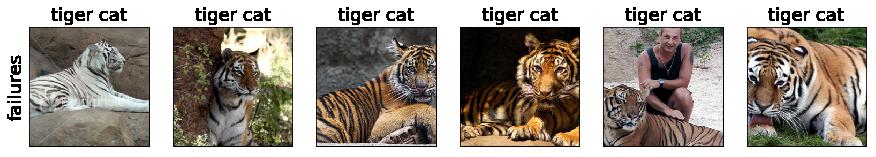}
\end{subfigure}
\caption{Visualization of feature[544]. For images with \textbf{label tiger cat}, when feature$[544] < 0.2036$, error rate increases to $0.8754$ (\textcolor{red}{\textbf{+37.85\%}}).}
\label{fig:appendix_label_tiger_cat}
\end{figure*}

\begin{figure*}[t]
\centering
\begin{subfigure}{0.9\linewidth}
\centering
\includegraphics[trim=0cm 0cm 0cm 0.9cm, clip, width=\linewidth]{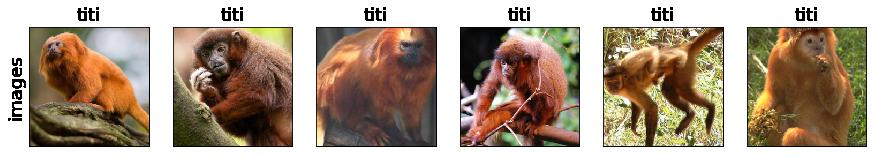}
\end{subfigure}\\
\begin{subfigure}{0.9\linewidth}
\centering
\includegraphics[trim=0cm 0cm 0cm 0.9cm, clip, width=\linewidth]{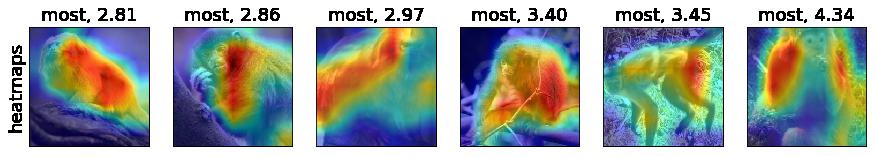}
\end{subfigure}\\
\begin{subfigure}{0.9\linewidth}
\centering
\includegraphics[trim=0cm 0cm 0cm 0.9cm, clip, width=\linewidth]{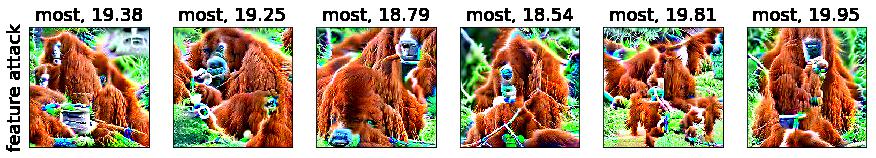}
\end{subfigure}\\
\begin{subfigure}{0.9\linewidth}
\centering
\includegraphics[trim=0cm 0cm 0cm 0.9cm, clip, width=\linewidth]{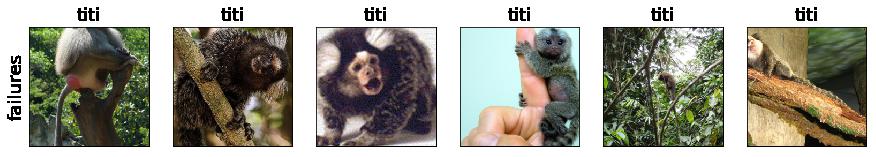}
\end{subfigure}
\caption{Visualization of feature[1911]. For images with \textbf{label titi}, when feature$[1911] < 0.7329$, error rate increases to $0.5240$ ($\textcolor{red}{\textbf{+11.09\%}}$).}
\label{fig:appendix_label_titi}
\end{figure*}

\begin{figure*}[t]
\centering
\begin{subfigure}{0.9\linewidth}
\centering
\includegraphics[trim=0cm 0cm 0cm 0.9cm, clip, width=\linewidth]{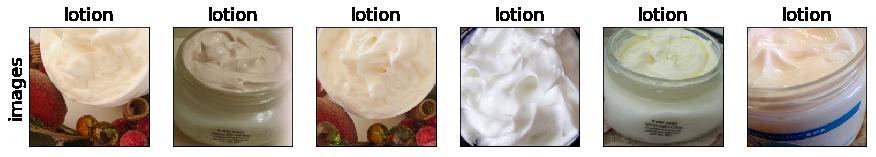}
\end{subfigure}\\
\begin{subfigure}{0.9\linewidth}
\centering
\includegraphics[trim=0cm 0cm 0cm 0.9cm, clip, width=\linewidth]{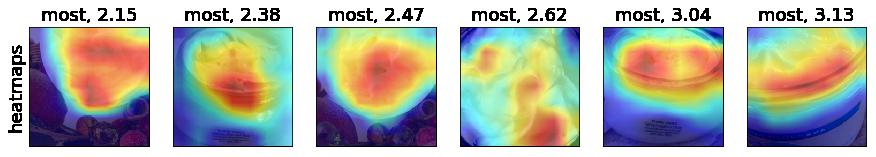}
\end{subfigure}\\
\begin{subfigure}{0.9\linewidth}
\centering
\includegraphics[trim=0cm 0cm 0cm 0.9cm, clip, width=\linewidth]{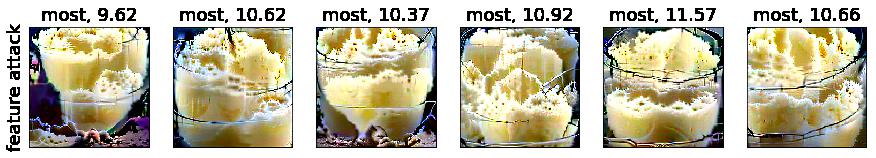}
\end{subfigure}\\
\begin{subfigure}{0.9\linewidth}
\centering
\includegraphics[trim=0cm 0cm 0cm 0.9cm, clip, width=\linewidth]{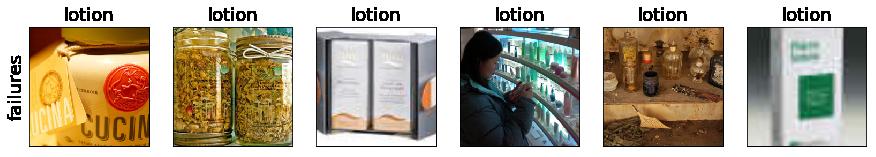}
\end{subfigure}
\caption{Visualization of feature[776]. For images with \textbf{label lotion}, when feature$[776] < 0.3313$, error rate increases to $0.4797$ (\textcolor{red}{\textbf{+11.73\%}}).}
\label{fig:appendix_label_lotion}
\end{figure*}

\begin{figure*}[t]
\centering
\begin{subfigure}{0.9\linewidth}
\centering
\includegraphics[trim=0cm 0cm 0cm 0.9cm, clip, width=\linewidth]{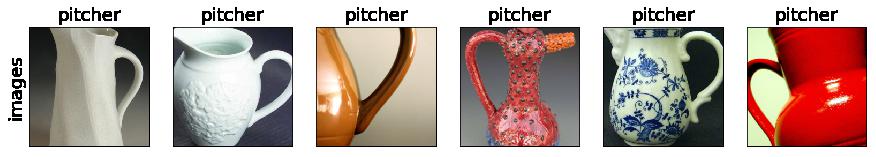}
\end{subfigure}\\
\begin{subfigure}{0.9\linewidth}
\centering
\includegraphics[trim=0cm 0cm 0cm 0.9cm, clip, width=\linewidth]{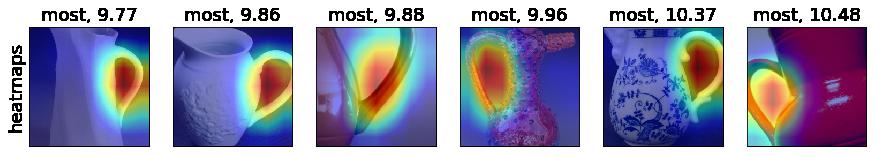}
\end{subfigure}\\
\begin{subfigure}{0.9\linewidth}
\centering
\includegraphics[trim=0cm 0cm 0cm 0.9cm, clip, width=\linewidth]{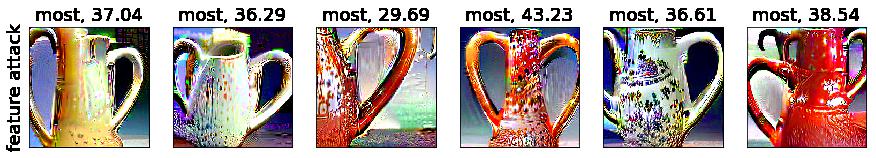}
\end{subfigure}\\
\begin{subfigure}{0.9\linewidth}
\centering
\includegraphics[trim=0cm 0cm 0cm 0.9cm, clip, width=\linewidth]{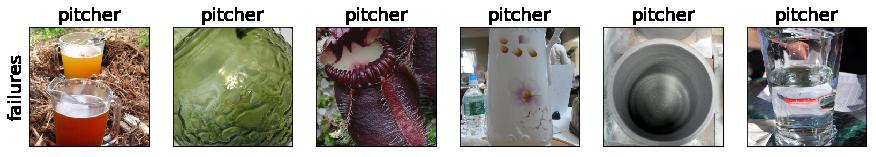}
\end{subfigure}
\caption{Visualization of feature[1378]. For images with \textbf{label pitcher}, when feature$[1378] < 0.7671$, error rate increases to $0.6253$ (\textcolor{red}{\textbf{+28.15\%}}).}
\label{fig:appendix_label_pitcher}
\end{figure*}

\begin{figure*}[t]
\centering
\begin{subfigure}{0.9\linewidth}
\centering
\includegraphics[trim=0cm 0cm 0cm 0.9cm, clip, width=\linewidth]{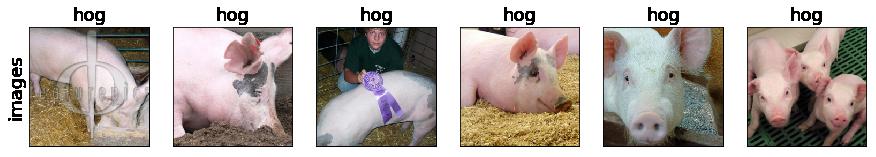}
\end{subfigure}\\
\begin{subfigure}{0.9\linewidth}
\centering
\includegraphics[trim=0cm 0cm 0cm 0.9cm, clip, width=\linewidth]{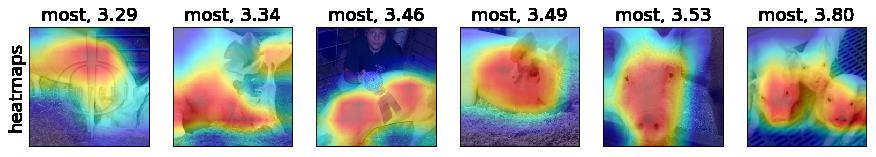}
\end{subfigure}\\
\begin{subfigure}{0.9\linewidth}
\centering
\includegraphics[trim=0cm 0cm 0cm 0.9cm, clip, width=\linewidth]{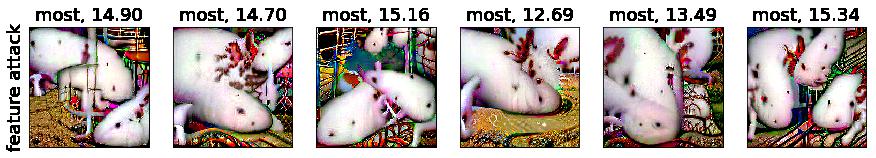}
\end{subfigure}\\
\begin{subfigure}{0.9\linewidth}
\centering
\includegraphics[trim=0cm 0cm 0cm 0.9cm, clip, width=\linewidth]{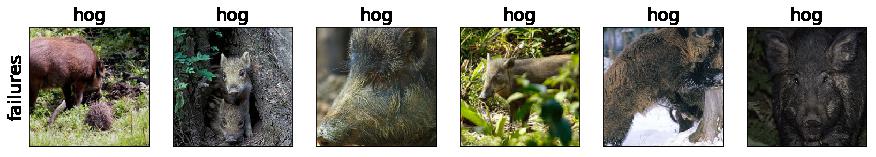}
\end{subfigure}
\caption{Visualization of feature[1611]. For images with \textbf{label hog}, when feature$[1611] < 0.0578$, error rate increases to $0.6842$ (\textcolor{red}{\textbf{+35.27\%}}).}
\label{fig:appendix_label_hog}
\end{figure*}

\begin{figure*}[t]
\centering
\begin{subfigure}{0.9\linewidth}
\centering
\includegraphics[trim=0cm 0cm 0cm 0.9cm, clip, width=\linewidth]{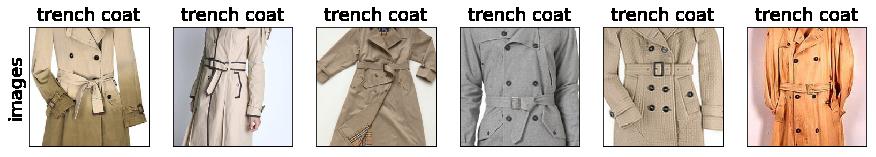}
\end{subfigure}\\
\begin{subfigure}{0.9\linewidth}
\centering
\includegraphics[trim=0cm 0cm 0cm 0.9cm, clip, width=\linewidth]{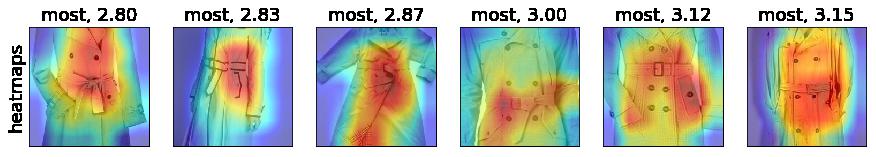}
\end{subfigure}\\
\begin{subfigure}{0.9\linewidth}
\centering
\includegraphics[trim=0cm 0cm 0cm 0.9cm, clip, width=\linewidth]{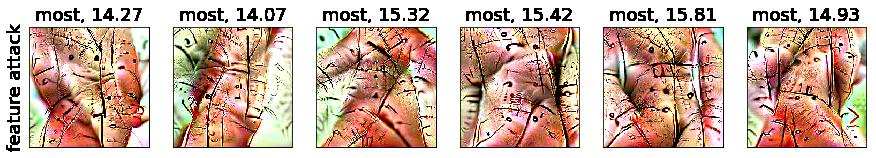}
\end{subfigure}\\
\begin{subfigure}{0.9\linewidth}
\centering
\includegraphics[trim=0cm 0cm 0cm 0.9cm, clip, width=\linewidth]{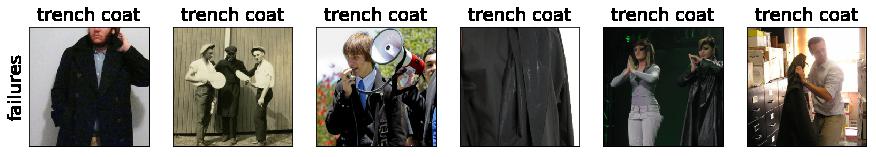}
\end{subfigure}
\caption{Visualization of feature[1264]. For images with \textbf{label trench coat}, when feature$[1264] < 0.6915$, error rate increases to $0.3196$ (\textcolor{red}{\textbf{+18.57\%}}).}
\label{fig:appendix_label_trench_coat}
\end{figure*}

\begin{figure*}[t]
\centering
\begin{subfigure}{0.9\linewidth}
\centering
\includegraphics[trim=0cm 0cm 0cm 0.9cm, clip, width=\linewidth]{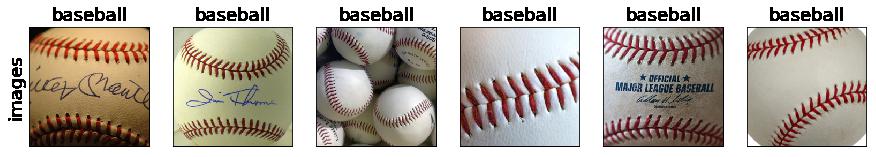}
\end{subfigure}\\
\begin{subfigure}{0.9\linewidth}
\centering
\includegraphics[trim=0cm 0cm 0cm 0.9cm, clip, width=\linewidth]{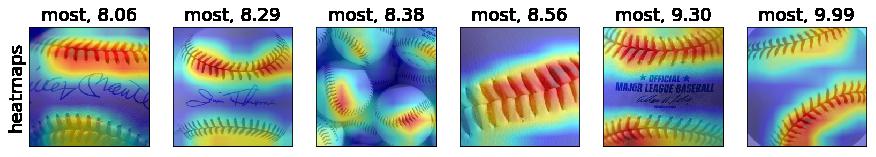}
\end{subfigure}\\
\begin{subfigure}{0.9\linewidth}
\centering
\includegraphics[trim=0cm 0cm 0cm 0.9cm, clip, width=\linewidth]{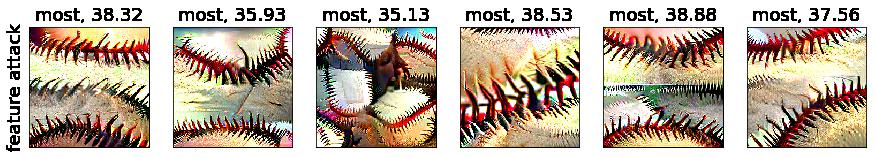}
\end{subfigure}\\
\begin{subfigure}{0.9\linewidth}
\centering
\includegraphics[trim=0cm 0cm 0cm 0.9cm, clip, width=\linewidth]{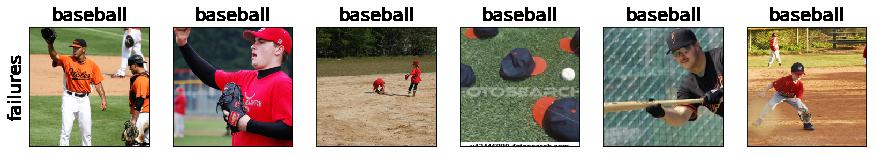}
\end{subfigure}
\caption{Visualization of feature[1081]. For images with \textbf{label baseball}, when feature$[1081] < 0.5461$, error rate increases to $0.3034$ (\textcolor{red}{\textbf{+19.65\%}}).}
\label{fig:appendix_label_baseball}
\end{figure*}

\subsubsection{Grouping by prediction}
Results are in Table \ref{table:standard_pred_grouping}.

\begin{figure*}[t]
\centering
\begin{subfigure}{0.9\linewidth}
\centering
\includegraphics[trim=0cm 0cm 0cm 0.9cm, clip, width=\linewidth]{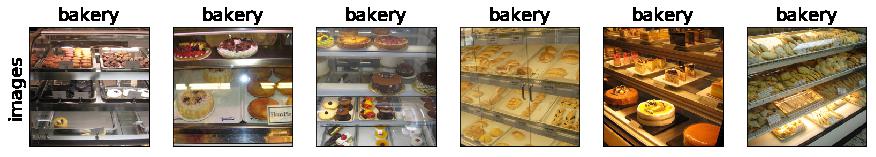}
\end{subfigure}\\
\begin{subfigure}{0.9\linewidth}
\centering
\includegraphics[trim=0cm 0cm 0cm 0.9cm, clip, width=\linewidth]{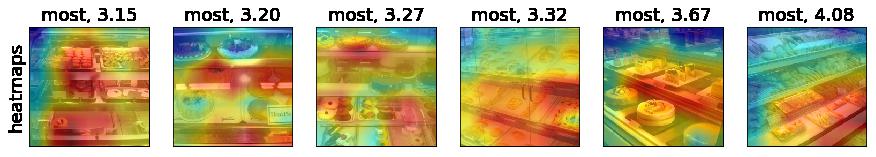}
\end{subfigure}\\
\begin{subfigure}{0.9\linewidth}
\centering
\includegraphics[trim=0cm 0cm 0cm 0.9cm, clip, width=\linewidth]{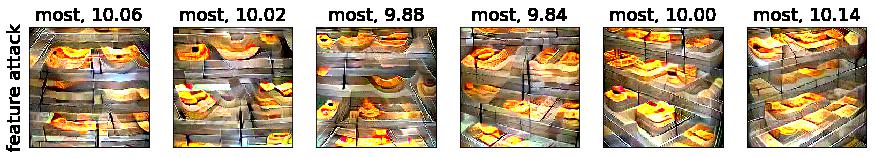}
\end{subfigure}\\
\begin{subfigure}{0.9\linewidth}
\centering
\includegraphics[trim=0cm 0cm 0cm 0.9cm, clip, width=\linewidth]{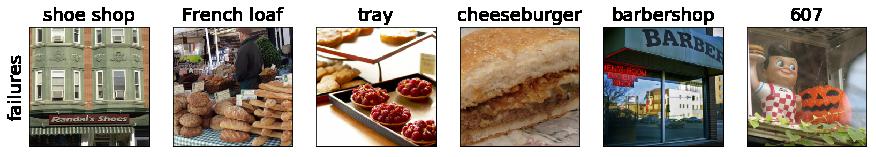}
\end{subfigure}
\caption{Visualization of feature[443]. For images with \textbf{prediction bakery}, when feature$[443] < 1.1382$, error rate increases to $0.3875$ (\textcolor{red}{\textbf{+11.80\%}}).}
\label{fig:appendix_standard_pred_bakery}
\end{figure*}

\begin{table*}
\centering
\begin{tabular}{| p{1.7cm} || p{1cm} | p{1.5cm} | p{1cm} | p{1cm} | p{1cm} | p{1cm} | p{1.9cm} | p{3cm} |}
\hline
\textbf{Class name} & \textbf{Feature index} & \textbf{Decision rule} & \textbf{\baseerrorratetext} & \textbf{\errorratetext} & \textbf{\errorcoveragetext} & \textbf{\avgtreetext} &\textbf{Feature}\newline \textbf{visualization} & \textbf{Feature name\newline (from visualization)}\\
\hline
bakery & 443 & $ < 1.1382$ & $0.2695$ & $0.3875$ & $0.7793$ & $0.3307$ & Figure \ref{fig:appendix_standard_pred_bakery} & shelves with sweets \\
\hline
polaroid camera & 793 & $ < 0.8166 $ & $0.1141$ & $0.2713$ &  $0.8671$ & $0.2384$ & Figure \ref{fig:appendix_standard_pred_polaroid_camera} & close-up view\newline of camera \\
\hline
saluki & 1395 & $ < 0.3263 $ & $0.1122$ & $0.2287$ & $0.5772$ & $0.1600$ & Figure \ref{fig:appendix_standard_pred_saluki} & long and hairy\newline dog ears \\
\hline
trailer truck & 1451 & $ < 0.2181 $ & $0.1121$ & $0.2184$ & $0.5036$ & $0.1472$ & Figure \ref{fig:appendix_standard_pred_trailer_truck} & white  truck \\
\hline
apiary & 1909 & $ < 0.5646$ & $0.1057$ & $0.2341$ & $0.8041$ & $0.1969$ & Figure \ref{fig:appendix_standard_pred_apiary} & white boxes \\
\hline
anemone fish & 262 & $ < 0.1792 $ & $0.1056$ & $0.2573$ & $0.4247$ & $ 0.1516$ & Figure \ref{fig:appendix_standard_pred_anemone_fish} & red fish \\
\hline
theater\newline curtain & 1063 & $ < 0.9047$ & $0.1049$ & $0.2482$ & $0.5072$ & $ 0.1583$ & Figure \ref{fig:appendix_standard_pred_theater_curtain} & red curtain \\
\hline
forklift & 943 & $ < 1.1721$ & $0.1047$ & $0.2379$ & $0.8889$ & $0.2136$ & Figure \ref{fig:appendix_standard_pred_forklift} & orange car\\
\hline
french \newline bulldog & 404 & $ < 0.2946$ & $0.1022$ & $0.2103$ & $0.3712$ & $0.1273$ & Figure \ref{fig:appendix_standard_pred_french_bulldog} & dog nose \\
\hline
syringe & 638 & $ < 0.2325 $ & $0.2020$ & $0.3519$ & $0.4894$ & $0.2455$ & Figure \ref{fig:appendix_standard_pred_syringe} & measurements \\
\hline
rhodesian ridgeback & 1634 & $ < 1.3779 $ & $0.2093$ & $0.3184$ & $0.4561$ & $0.2337$ & Figure \ref{fig:appendix_standard_pred_rhodesian_ridgeback} & dog collar \\
\hline
\end{tabular}
\caption{Results on a standard Resnet-50 model using grouping by prediction.}
\label{table:standard_pred_grouping}
\end{table*}

\begin{figure*}[t]
\centering
\begin{subfigure}{0.9\linewidth}
\centering
\includegraphics[trim=0cm 0cm 0cm 0.9cm, clip, width=\linewidth]{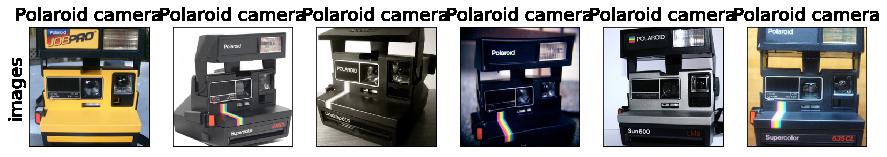}
\end{subfigure}\\
\begin{subfigure}{0.9\linewidth}
\centering
\includegraphics[trim=0cm 0cm 0cm 0.9cm, clip, width=\linewidth]{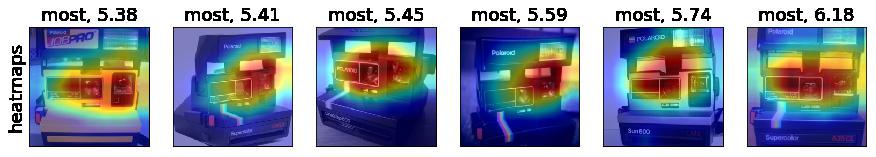}
\end{subfigure}\\
\begin{subfigure}{0.9\linewidth}
\centering
\includegraphics[trim=0cm 0cm 0cm 0.9cm, clip, width=\linewidth]{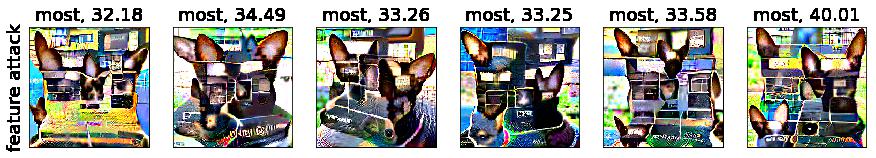}
\end{subfigure}\\
\begin{subfigure}{0.9\linewidth}
\centering
\includegraphics[trim=0cm 0cm 0cm 0.9cm, clip, width=\linewidth]{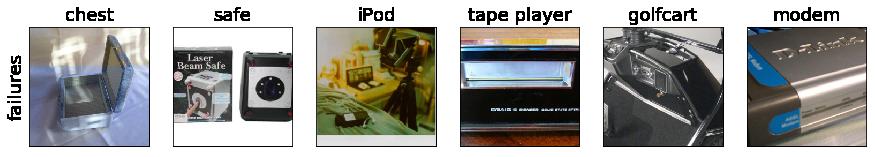}
\end{subfigure}
\caption{Visualization of feature[793]. For images with \textbf{prediction polaroid camera}, when feature$[793] < 0.8166$, error rate increases to $0.2713$ (\textcolor{red}{\textbf{+15.72\%}}).}
\label{fig:appendix_standard_pred_polaroid_camera}
\end{figure*}

\begin{figure*}[t]
\centering
\begin{subfigure}{0.9\linewidth}
\centering
\includegraphics[trim=0cm 0cm 0cm 0.9cm, clip, width=\linewidth]{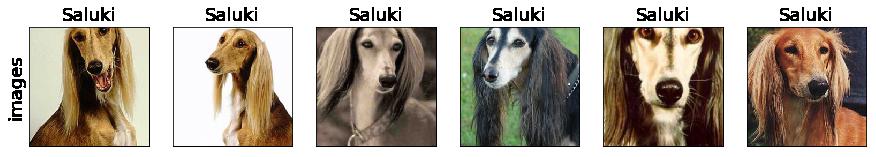}
\end{subfigure}\\
\begin{subfigure}{0.9\linewidth}
\centering
\includegraphics[trim=0cm 0cm 0cm 0.9cm, clip, width=\linewidth]{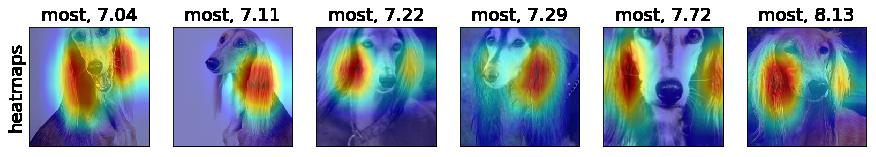}
\end{subfigure}\\
\begin{subfigure}{0.9\linewidth}
\centering
\includegraphics[trim=0cm 0cm 0cm 0.9cm, clip, width=\linewidth]{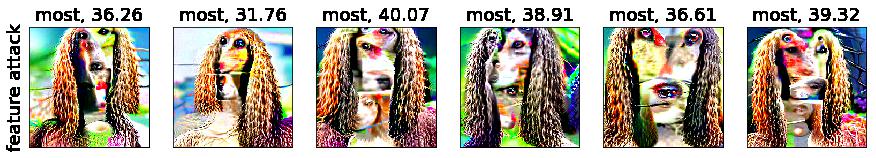}
\end{subfigure}\\
\begin{subfigure}{0.9\linewidth}
\centering
\includegraphics[trim=0cm 0cm 0cm 0.9cm, clip, width=\linewidth]{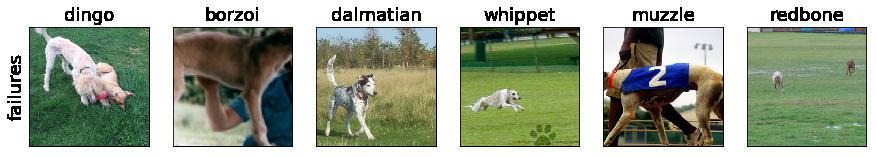}
\end{subfigure}
\caption{Visualization of feature[1395]. For images with \textbf{prediction saluki}, when feature$[1395] < 0.3263$, error rate increases to $0.2287$ (\textcolor{red}{\textbf{+11.65\%}}).}
\label{fig:appendix_standard_pred_saluki}
\end{figure*}

\begin{figure*}[t]
\centering
\begin{subfigure}{0.9\linewidth}
\centering
\includegraphics[trim=0cm 0cm 0cm 0.9cm, clip, width=\linewidth]{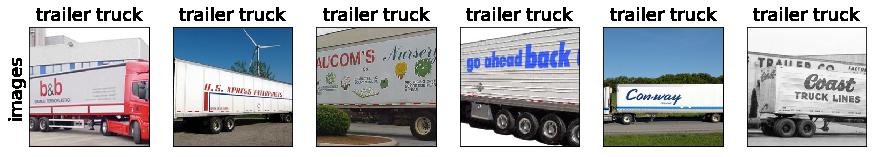}
\end{subfigure}\\
\begin{subfigure}{0.9\linewidth}
\centering
\includegraphics[trim=0cm 0cm 0cm 0.9cm, clip, width=\linewidth]{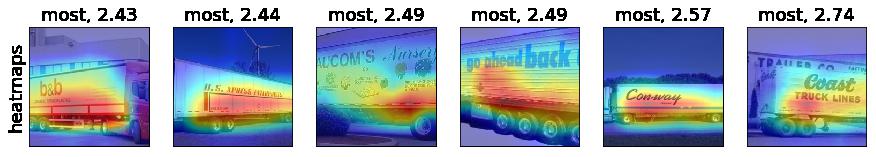}
\end{subfigure}\\
\begin{subfigure}{0.9\linewidth}
\centering
\includegraphics[trim=0cm 0cm 0cm 0.9cm, clip, width=\linewidth]{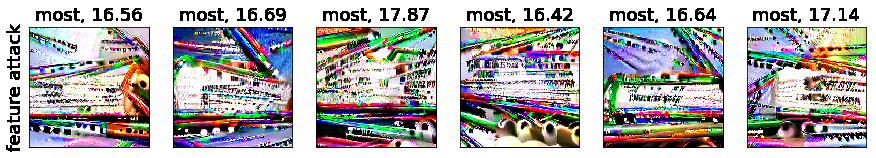}
\end{subfigure}\\
\begin{subfigure}{0.9\linewidth}
\centering
\includegraphics[trim=0cm 0cm 0cm 0.9cm, clip, width=\linewidth]{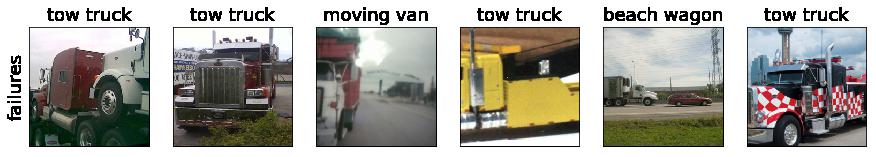}
\end{subfigure}
\caption{Visualization of feature[1451]. For images with \textbf{prediction trailer truck}, when feature$[1451] < 0.2181$, error rate increases to $0.2184$ (\textcolor{red}{\textbf{+10.63\%}}).}
\label{fig:appendix_standard_pred_trailer_truck}
\end{figure*}

\begin{figure*}[t]
\centering
\begin{subfigure}{0.9\linewidth}
\centering
\includegraphics[trim=0cm 0cm 0cm 0.9cm, clip, width=\linewidth]{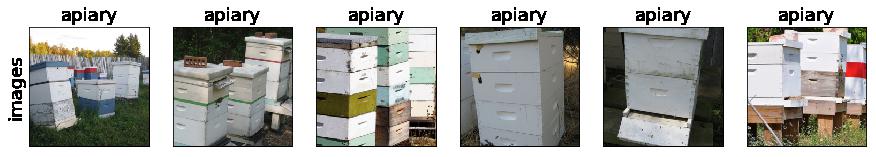}
\end{subfigure}\\
\begin{subfigure}{0.9\linewidth}
\centering
\includegraphics[trim=0cm 0cm 0cm 0.9cm, clip, width=\linewidth]{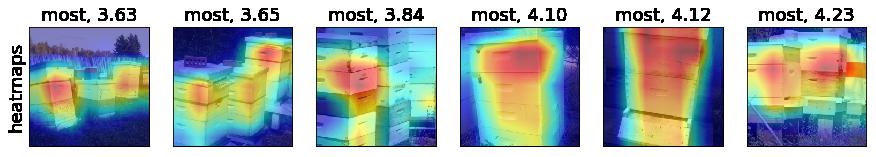}
\end{subfigure}\\
\begin{subfigure}{0.9\linewidth}
\centering
\includegraphics[trim=0cm 0cm 0cm 0.9cm, clip, width=\linewidth]{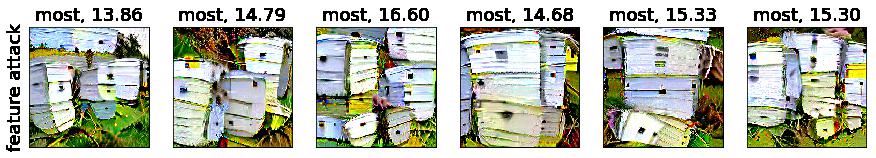}
\end{subfigure}\\
\begin{subfigure}{0.9\linewidth}
\centering
\includegraphics[trim=0cm 0cm 0cm 0.9cm, clip, width=\linewidth]{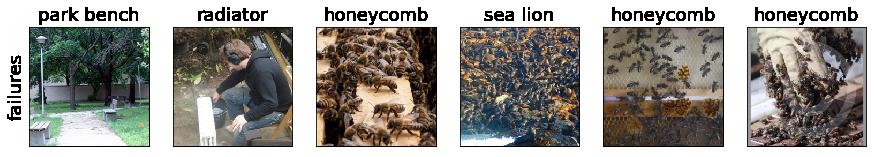}
\end{subfigure}
\caption{Visualization of feature[1909]. For images with \textbf{prediction apiary}, when feature$[1909] < 0.5646$, error rate increases to $0.2371$ (\textcolor{red}{\textbf{+13.14\%}}).}
\label{fig:appendix_standard_pred_apiary}
\end{figure*}

\begin{figure*}[t]
\centering
\begin{subfigure}{0.9\linewidth}
\centering
\includegraphics[trim=0cm 0cm 0cm 0.9cm, clip, width=\linewidth]{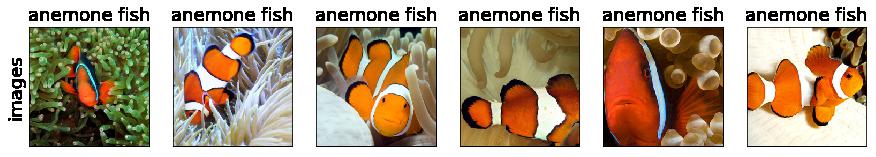}
\end{subfigure}\\
\begin{subfigure}{0.9\linewidth}
\centering
\includegraphics[trim=0cm 0cm 0cm 0.9cm, clip, width=\linewidth]{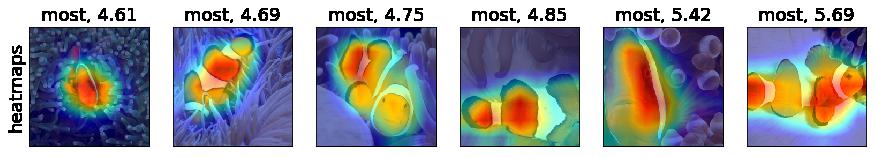}
\end{subfigure}\\
\begin{subfigure}{0.9\linewidth}
\centering
\includegraphics[trim=0cm 0cm 0cm 0.9cm, clip, width=\linewidth]{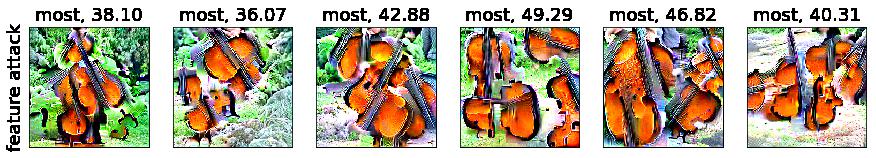}
\end{subfigure}\\
\begin{subfigure}{0.9\linewidth}
\centering
\includegraphics[trim=0cm 0cm 0cm 0.9cm, clip, width=\linewidth]{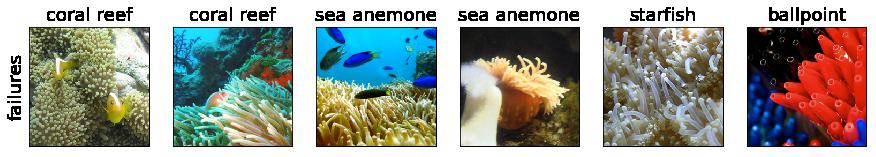}
\end{subfigure}
\caption{Visualization of feature[262]. For images with \textbf{prediction anemone fish}, when feature$[262] < 0.1792$, error rate increases to $0.2573$ (\textcolor{red}{\textbf{+15.17\%}}).}
\label{fig:appendix_standard_pred_anemone_fish}
\end{figure*}

\begin{figure*}[t]
\centering
\begin{subfigure}{0.9\linewidth}
\centering
\includegraphics[trim=0cm 0cm 0cm 0.9cm, clip, width=\linewidth]{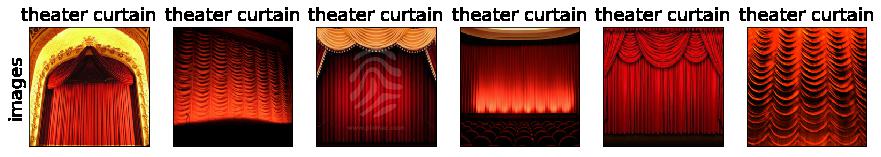}
\end{subfigure}\\
\begin{subfigure}{0.9\linewidth}
\centering
\includegraphics[trim=0cm 0cm 0cm 0.9cm, clip, width=\linewidth]{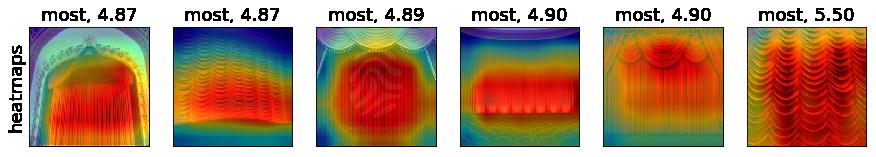}
\end{subfigure}\\
\begin{subfigure}{0.9\linewidth}
\centering
\includegraphics[trim=0cm 0cm 0cm 0.9cm, clip, width=\linewidth]{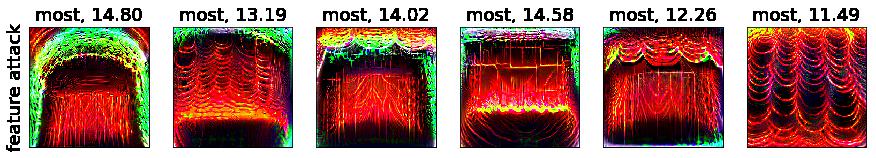}
\end{subfigure}\\
\begin{subfigure}{0.9\linewidth}
\centering
\includegraphics[trim=0cm 0cm 0cm 0.9cm, clip, width=\linewidth]{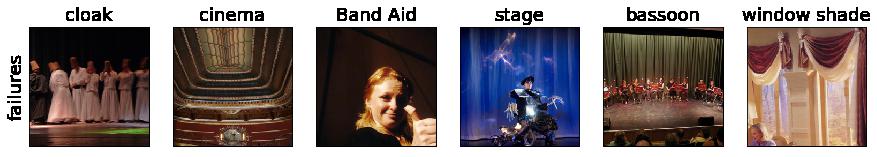}
\end{subfigure}
\caption{Visualization of feature[1063]. For images with \textbf{prediction theater curtain}, when feature$[1063] < 0.9047$, error rate increases to $0.2482$ (\textcolor{red}{\textbf{+14.33\%}}).}
\label{fig:appendix_standard_pred_theater_curtain}
\end{figure*}

\begin{figure*}[t]
\centering
\begin{subfigure}{0.9\linewidth}
\centering
\includegraphics[trim=0cm 0cm 0cm 0.9cm, clip, width=\linewidth]{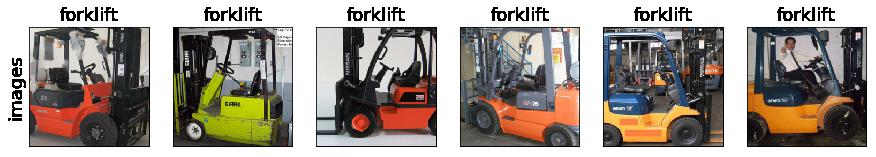}
\end{subfigure}\\
\begin{subfigure}{0.9\linewidth}
\centering
\includegraphics[trim=0cm 0cm 0cm 0.9cm, clip, width=\linewidth]{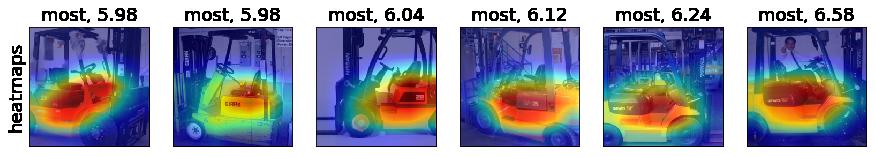}
\end{subfigure}\\
\begin{subfigure}{0.9\linewidth}
\centering
\includegraphics[trim=0cm 0cm 0cm 0.9cm, clip, width=\linewidth]{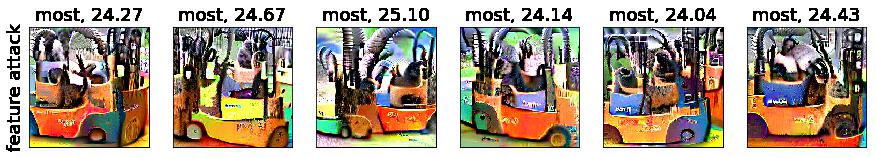}
\end{subfigure}\\
\begin{subfigure}{0.9\linewidth}
\centering
\includegraphics[trim=0cm 0cm 0cm 0.9cm, clip, width=\linewidth]{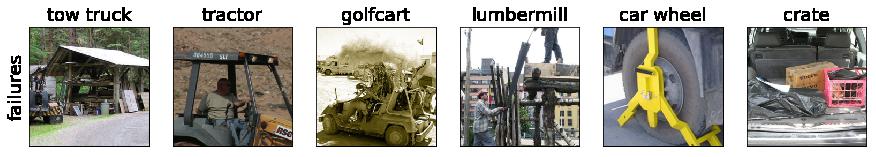}
\end{subfigure}
\caption{Visualization of feature[943]. For images with \textbf{prediction forklift}, when feature$[943] < 1.1721$, error rate increases to $0.2379$ (\textcolor{red}{\textbf{+13.32\%}}).}
\label{fig:appendix_standard_pred_forklift}
\end{figure*}

\begin{figure*}[t]
\centering
\begin{subfigure}{0.9\linewidth}
\centering
\includegraphics[trim=0cm 0cm 0cm 0.9cm, clip, width=\linewidth]{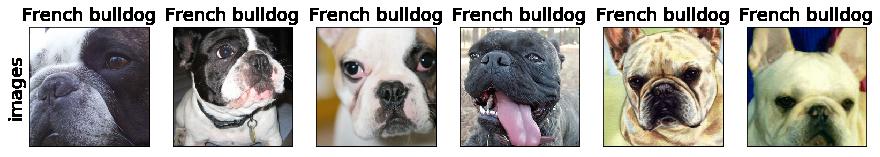}
\end{subfigure}\\
\begin{subfigure}{0.9\linewidth}
\centering
\includegraphics[trim=0cm 0cm 0cm 0.9cm, clip, width=\linewidth]{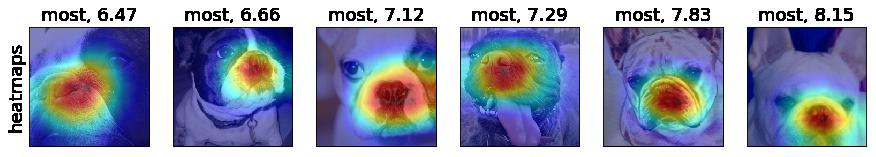}
\end{subfigure}\\
\begin{subfigure}{0.9\linewidth}
\centering
\includegraphics[trim=0cm 0cm 0cm 0.9cm, clip, width=\linewidth]{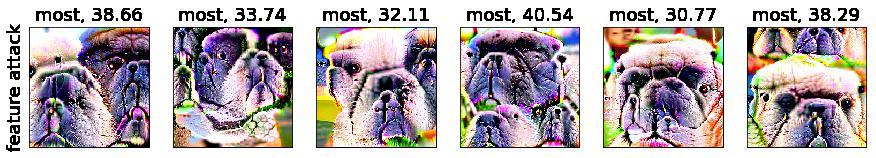}
\end{subfigure}\\
\begin{subfigure}{0.9\linewidth}
\centering
\includegraphics[trim=0cm 0cm 0cm 0.9cm, clip, width=\linewidth]{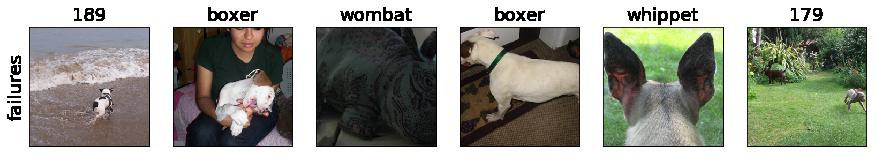}
\end{subfigure}
\caption{Visualization of feature[404]. For images with \textbf{prediction french bulldog}, when feature$[404] < 0.2946$, error rate increases to $0.2103$ (\textcolor{red}{\textbf{+10.81\%}}).}
\label{fig:appendix_standard_pred_french_bulldog}
\end{figure*}

\begin{figure*}[t]
\centering
\begin{subfigure}{0.9\linewidth}
\centering
\includegraphics[trim=0cm 0cm 0cm 0.9cm, clip, width=\linewidth]{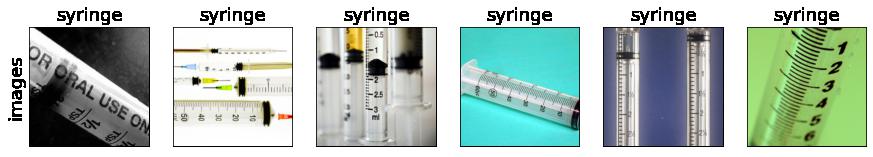}
\end{subfigure}\\
\begin{subfigure}{0.9\linewidth}
\centering
\includegraphics[trim=0cm 0cm 0cm 0.9cm, clip, width=\linewidth]{viz/syringe_heatmaps}
\end{subfigure}\\
\begin{subfigure}{0.9\linewidth}
\centering
\includegraphics[trim=0cm 0cm 0cm 0.9cm, clip, width=\linewidth]{viz/syringe_attacks}
\end{subfigure}\\
\begin{subfigure}{0.9\linewidth}
\centering
\includegraphics[trim=0cm 0cm 0cm 0.9cm, clip, width=\linewidth]{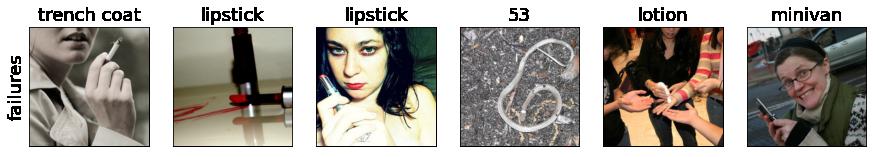}
\end{subfigure}
\caption{Visualization of feature[638]. For images with \textbf{prediction syringe}, when feature$[638] < 0.2325$, error rate increases to $0.3519$ (\textcolor{red}{\textbf{+14.99\%}}).}
\label{fig:appendix_standard_pred_syringe}
\end{figure*}

\begin{figure*}[t]
\centering
\begin{subfigure}{0.9\linewidth}
\centering
\includegraphics[trim=0.5cm 0cm 0cm 0.9cm, clip, width=\linewidth]{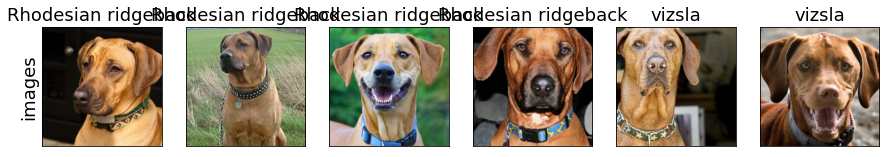}
\end{subfigure}\\
\begin{subfigure}{0.9\linewidth}
\centering
\includegraphics[trim=0cm 0cm 0cm 0.9cm, clip, width=\linewidth]{viz/rhodesian_ridgeback_heatmaps}
\end{subfigure}\\
\begin{subfigure}{0.9\linewidth}
\centering
\includegraphics[trim=0cm 0cm 0cm 0.9cm, clip, width=\linewidth]{viz/rhodesian_ridgeback_attacks}
\end{subfigure}\\
\begin{subfigure}{0.9\linewidth}
\centering
\includegraphics[trim=0cm 0cm 0cm 0.9cm, clip, width=\linewidth]{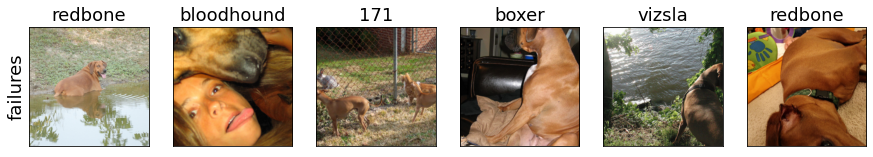}
\end{subfigure}
\caption{Visualization of feature[1634]. For images with \textbf{prediction rhodesian ridgeback}, when feature$[1634] < 1.3779$, error rate increases to $0.3184$ (\textcolor{red}{\textbf{+10.91\%}}).}
\label{fig:appendix_standard_pred_rhodesian_ridgeback}
\end{figure*}

\subsection{Failure explanation for a robust model}\label{sec:failure_modes_robust}
\subsubsection{Grouping by label}
Results are in Table \ref{table:robust_label_grouping}.

\begin{figure*}[t]
\centering
\begin{subfigure}{0.9\linewidth}
\centering
\includegraphics[trim=0cm 0cm 0cm 0.9cm, clip, width=\linewidth]{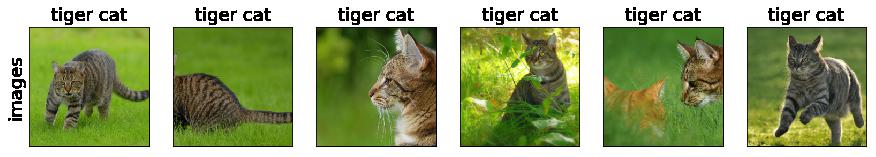}
\end{subfigure}\\
\begin{subfigure}{0.9\linewidth}
\centering
\includegraphics[trim=0cm 0cm 0cm 0.9cm, clip, width=\linewidth]{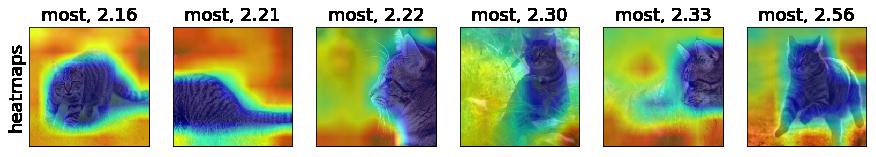}
\end{subfigure}\\
\begin{subfigure}{0.9\linewidth}
\centering
\includegraphics[trim=0cm 0cm 0cm 0.9cm, clip, width=\linewidth]{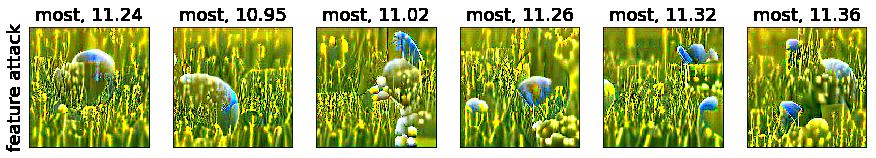}
\end{subfigure}\\
\begin{subfigure}{0.9\linewidth}
\centering
\includegraphics[trim=0cm 0cm 0cm 0.9cm, clip, width=\linewidth]{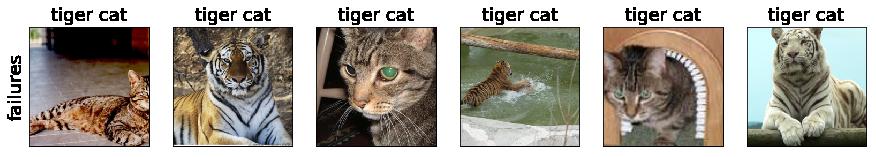}
\end{subfigure}
\caption{Visualization of feature[1864]. For images with \textbf{label tiger cat}, when feature$[1864] < 0.4673$, error rate increases to $0.8786$ (\textcolor{red}{\textbf{+10.71\%}}).}
\label{fig:appendix_robust_label_tiger_cat}
\end{figure*}

\begin{table*}
\centering
\begin{tabular}{| p{1.7cm} || p{1cm} | p{1.5cm} | p{1cm} | p{1cm} | p{1cm} | p{1cm} | p{1.9cm} | p{3cm} |}
\hline
\textbf{Class name} & \textbf{Feature index} & \textbf{Decision rule} & \textbf{\baseerrorratetext} & \textbf{\errorratetext} & \textbf{\errorcoveragetext} & \textbf{\avgtreetext} &\textbf{Feature} \newline \textbf{visualization} & \textbf{Feature name\newline (from visualization)}\\
\hline
tiger cat & 1864 & $ < 0.4673 $ & $0.7715$ & $0.8786$ &  $0.9521$ & $0.8473$ & Figure \ref{fig:appendix_robust_label_tiger_cat} & green background\\
\hline
lighter & 380 & $ < 1.2961$ & $0.7285$ & $0.8608$ & $0.9335$ & $0.8188$ & Figure \ref{fig:appendix_robust_label_lighter} & flame \\
\hline
purse & 486 & $ < 0.1915 $ & $0.7277$ & $0.9258$ & $0.5011$ & $0.7627$ & Figure \ref{fig:appendix_robust_label_purse} & strings \\
\hline
chihuahua & 198 & $ < 0.7873 $ & $0.7269$ & $0.9332$ & $0.7386$ & $0.8062$ & Figure \ref{fig:appendix_robust_label_chihuahua} & close-up face \\
\hline
rifle & 522 & $ < 1.4414$ & $0.7223$ & $0.9558$ & $0.5985$ & $0.7846$ & Figure \ref{fig:appendix_robust_label_rifle} & trigger \\
\hline
crayfish & 1729 & $ < 1.0135 $ & $0.7154$ & $0.9294$ & $0.5806$ & $ 0.7671$ & Figure \ref{fig:appendix_robust_label_crayfish} & red fish skeleton \\
\hline
cougar & 1469 & $ < 0.6376$ & $0.3438$ & $0.6074$ & $0.9172$ & $0.5620$ & Figure \ref{fig:appendix_robust_label_cougar} & cougar nose \\
\hline
butternut squash & 1905 & $ < 0.6324$ & $0.3438$ & $0.6034$ & $0.7830$ & $0.5017$ & Figure \ref{fig:appendix_robust_label_butternut_squash} & orange round edge\\
\hline
sea\newline cucumber & 1147 & $ < 1.067$ & $0.3438$ & $0.6042$ & $0.7718$ & $0.4983$ & Figure \ref{fig:appendix_robust_label_sea_cucumber} & green round shape \\
\hline
zucchini & 752 & $ < 1.0602 $ & $0.3431$ & $0.6445$ & $0.8049$ & $0.5416$ & Figure \ref{fig:appendix_robust_label_zucchini} & green pipe \\
\hline
table lamp & 1740 & $ < 2.5241 $ & $0.3423$ & $0.7251$ & $0.7528$ & $0.5783$ & Figure \ref{fig:appendix_robust_label_table_lamp} & horizontal edge at\newline bottom of table lamp \\
\hline
\end{tabular}
\caption{Results on a robust Resnet-50 model using grouping by label.}
\label{table:robust_label_grouping}
\end{table*}

\begin{figure*}[t]
\centering
\begin{subfigure}{0.9\linewidth}
\centering
\includegraphics[trim=0cm 0cm 0cm 0.9cm, clip, width=\linewidth]{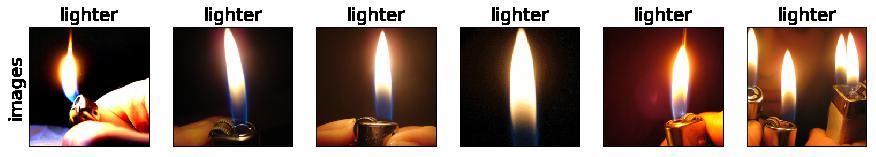}
\end{subfigure}\\
\begin{subfigure}{0.9\linewidth}
\centering
\includegraphics[trim=0cm 0cm 0cm 0.9cm, clip, width=\linewidth]{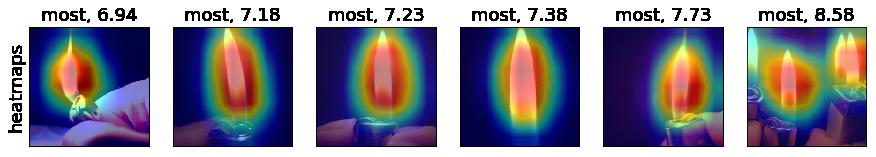}
\end{subfigure}\\
\begin{subfigure}{0.9\linewidth}
\centering
\includegraphics[trim=0cm 0cm 0cm 0.9cm, clip, width=\linewidth]{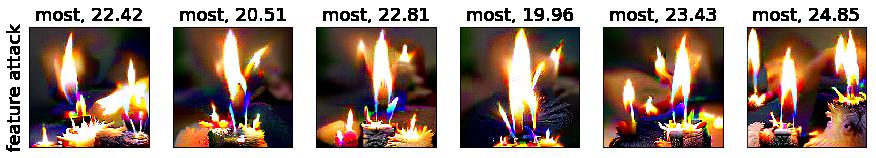}
\end{subfigure}\\
\begin{subfigure}{0.9\linewidth}
\centering
\includegraphics[trim=0cm 0cm 0cm 0.9cm, clip, width=\linewidth]{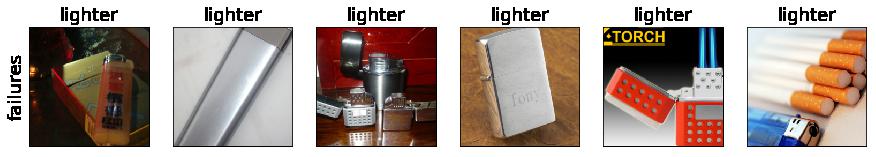}
\end{subfigure}
\caption{Visualization of feature[380]. For images with \textbf{label lighter}, when feature$[380] < 1.2961$, error rate increases to $0.8608$ (\textcolor{red}{\textbf{+13.23\%}}).}
\label{fig:appendix_robust_label_lighter}
\end{figure*}

\begin{figure*}[t]
\centering
\begin{subfigure}{0.9\linewidth}
\centering
\includegraphics[trim=0cm 0cm 0cm 0.9cm, clip, width=\linewidth]{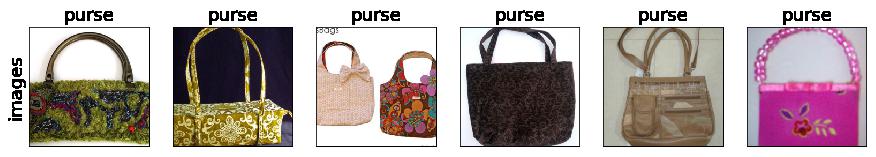}
\end{subfigure}\\
\begin{subfigure}{0.9\linewidth}
\centering
\includegraphics[trim=0cm 0cm 0cm 0.9cm, clip, width=\linewidth]{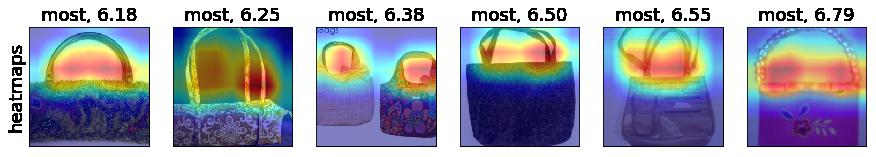}
\end{subfigure}\\
\begin{subfigure}{0.9\linewidth}
\centering
\includegraphics[trim=0cm 0cm 0cm 0.9cm, clip, width=\linewidth]{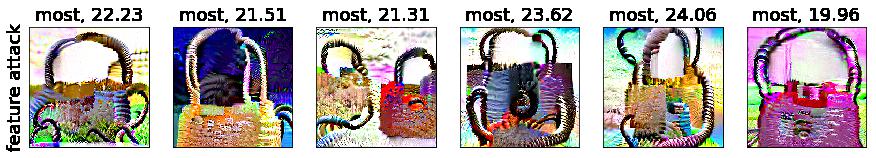}
\end{subfigure}\\
\begin{subfigure}{0.9\linewidth}
\centering
\includegraphics[trim=0cm 0cm 0cm 0.9cm, clip, width=\linewidth]{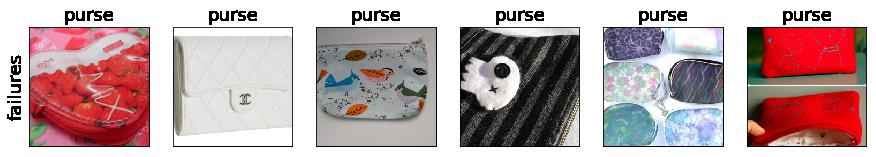}
\end{subfigure}
\caption{Visualization of feature[486]. For images with \textbf{label purse}, when feature$[486] < 0.1915$, error rate increases to $0.9258$ (\textcolor{red}{\textbf{+19.81\%}}).}
\label{fig:appendix_robust_label_purse}
\end{figure*}

\begin{figure*}[t]
\centering
\begin{subfigure}{0.9\linewidth}
\centering
\includegraphics[trim=0cm 0cm 0cm 0.9cm, clip, width=\linewidth]{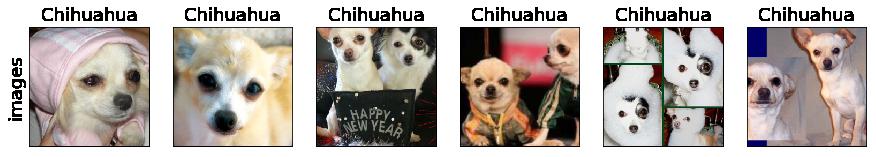}
\end{subfigure}\\
\begin{subfigure}{0.9\linewidth}
\centering
\includegraphics[trim=0cm 0cm 0cm 0.9cm, clip, width=\linewidth]{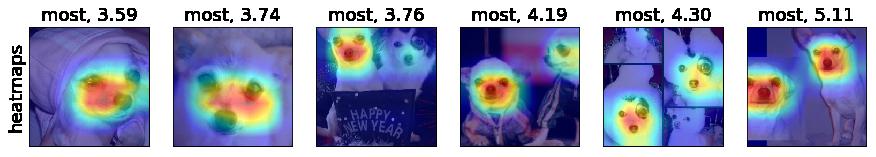}
\end{subfigure}\\
\begin{subfigure}{0.9\linewidth}
\centering
\includegraphics[trim=0cm 0cm 0cm 0.9cm, clip, width=\linewidth]{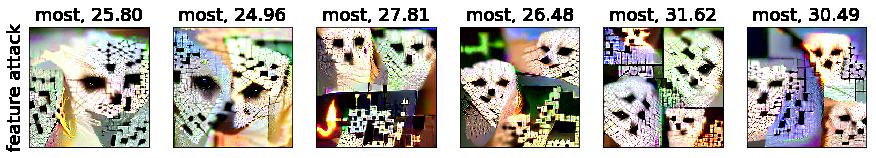}
\end{subfigure}\\
\begin{subfigure}{0.9\linewidth}
\centering
\includegraphics[trim=0cm 0cm 0cm 0.9cm, clip, width=\linewidth]{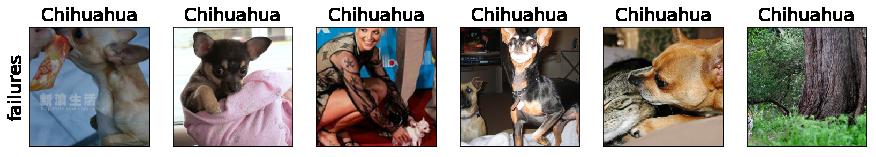}
\end{subfigure}
\caption{Visualization of feature[198]. For images with \textbf{label chihuahua}, when feature$[198] < 0.7873$, error rate increases to $0.9332$ (\textcolor{red}{\textbf{+20.63\%}}).}
\label{fig:appendix_robust_label_chihuahua}
\end{figure*}

\begin{figure*}[t]
\centering
\begin{subfigure}{0.9\linewidth}
\centering
\includegraphics[trim=0cm 0cm 0cm 0.9cm, clip, width=\linewidth]{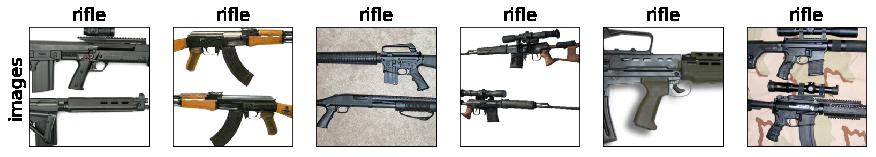}
\end{subfigure}\\
\begin{subfigure}{0.9\linewidth}
\centering
\includegraphics[trim=0cm 0cm 0cm 0.9cm, clip, width=\linewidth]{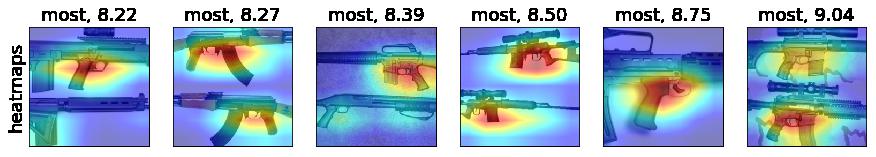}
\end{subfigure}\\
\begin{subfigure}{0.9\linewidth}
\centering
\includegraphics[trim=0cm 0cm 0cm 0.9cm, clip, width=\linewidth]{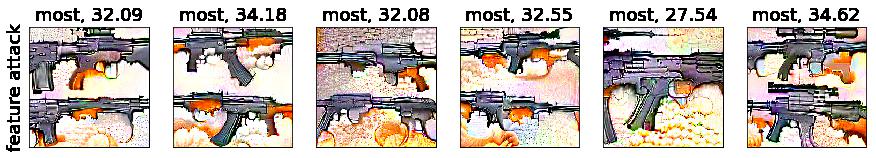}
\end{subfigure}\\
\begin{subfigure}{0.9\linewidth}
\centering
\includegraphics[trim=0cm 0cm 0cm 0.9cm, clip, width=\linewidth]{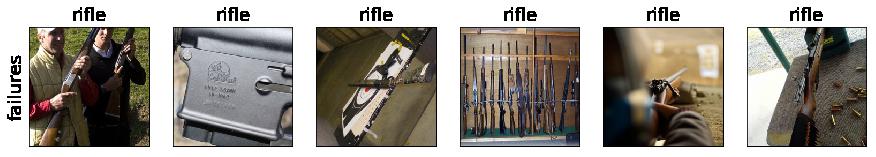}
\end{subfigure}
\caption{Visualization of feature[522]. For images with \textbf{label rifle}, when feature$[522] < 1.4414$, error rate increases to $0.9558$ (\textcolor{red}{\textbf{+23.35\%}}).}
\label{fig:appendix_robust_label_rifle}
\end{figure*}

\begin{figure*}[t]
\centering
\begin{subfigure}{0.9\linewidth}
\centering
\includegraphics[trim=0cm 0cm 0cm 0.9cm, clip, width=\linewidth]{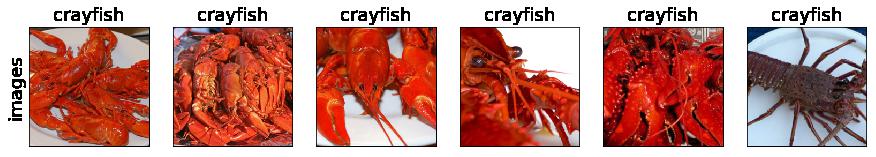}
\end{subfigure}\\
\begin{subfigure}{0.9\linewidth}
\centering
\includegraphics[trim=0cm 0cm 0cm 0.9cm, clip, width=\linewidth]{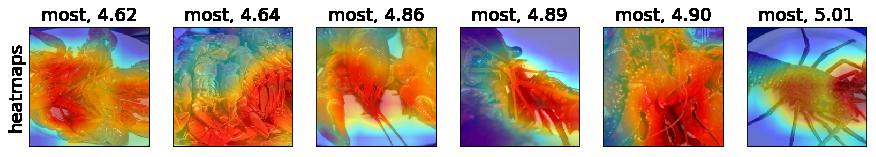}
\end{subfigure}\\
\begin{subfigure}{0.9\linewidth}
\centering
\includegraphics[trim=0cm 0cm 0cm 0.9cm, clip, width=\linewidth]{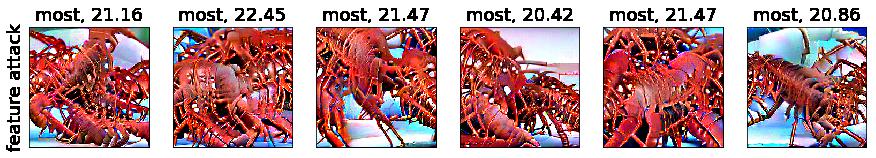}
\end{subfigure}\\
\begin{subfigure}{0.9\linewidth}
\centering
\includegraphics[trim=0cm 0cm 0cm 0.9cm, clip, width=\linewidth]{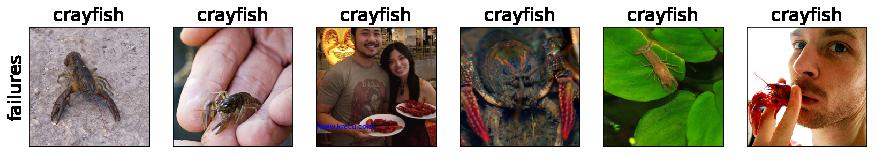}
\end{subfigure}
\caption{Visualization of feature[1729]. For images with \textbf{label crayfish}, when feature$[1729] < 1.0135$, error rate increases to $0.9294$ (\textcolor{red}{\textbf{+21.40\%}}).}
\label{fig:appendix_robust_label_crayfish}
\end{figure*}

\begin{figure*}[t]
\centering
\begin{subfigure}{0.9\linewidth}
\centering
\includegraphics[trim=0cm 0cm 0cm 0.9cm, clip, width=\linewidth]{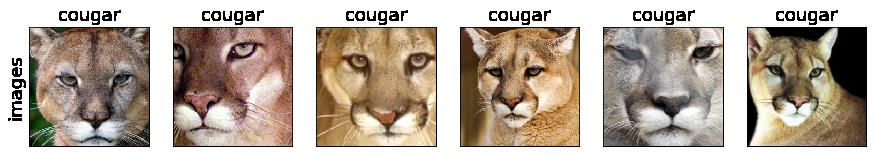}
\end{subfigure}\\
\begin{subfigure}{0.9\linewidth}
\centering
\includegraphics[trim=0cm 0cm 0cm 0.9cm, clip, width=\linewidth]{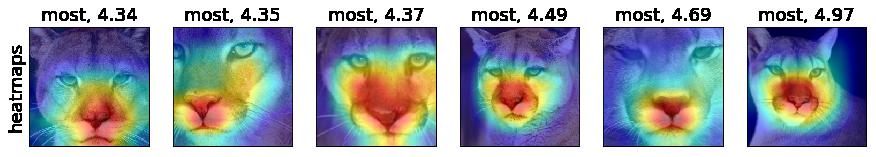}
\end{subfigure}\\
\begin{subfigure}{0.9\linewidth}
\centering
\includegraphics[trim=0cm 0cm 0cm 0.9cm, clip, width=\linewidth]{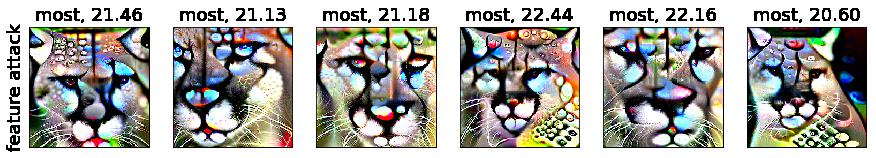}
\end{subfigure}\\
\begin{subfigure}{0.9\linewidth}
\centering
\includegraphics[trim=0cm 0cm 0cm 0.9cm, clip, width=\linewidth]{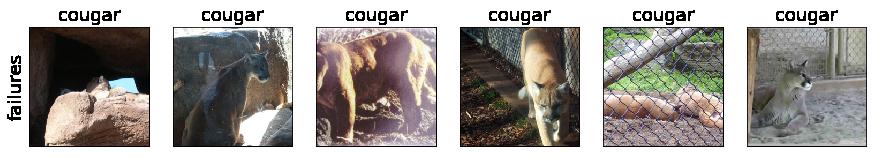}
\end{subfigure}
\caption{Visualization of feature[1469]. For images with \textbf{label cougar}, when feature$[1469] < 0.6376$, error rate increases to $0.6074$ (\textcolor{red}{\textbf{+26.36\%}}).}
\label{fig:appendix_robust_label_cougar}
\end{figure*}

\begin{figure*}[t]
\centering
\begin{subfigure}{0.9\linewidth}
\centering
\includegraphics[trim=0cm 0cm 0cm 0.9cm, clip, width=\linewidth]{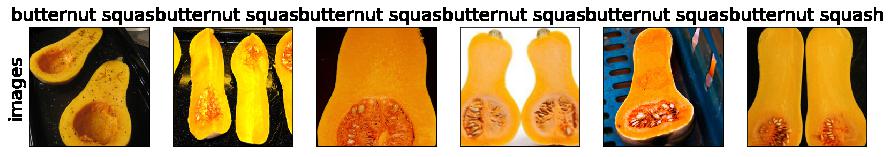}
\end{subfigure}\\
\begin{subfigure}{0.9\linewidth}
\centering
\includegraphics[trim=0cm 0cm 0cm 0.9cm, clip, width=\linewidth]{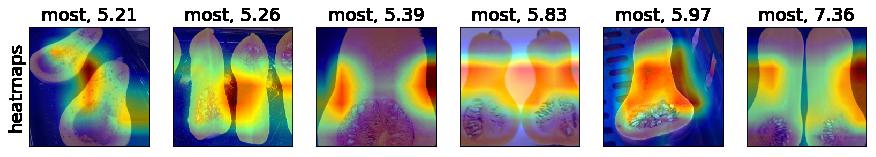}
\end{subfigure}\\
\begin{subfigure}{0.9\linewidth}
\centering
\includegraphics[trim=0cm 0cm 0cm 0.9cm, clip, width=\linewidth]{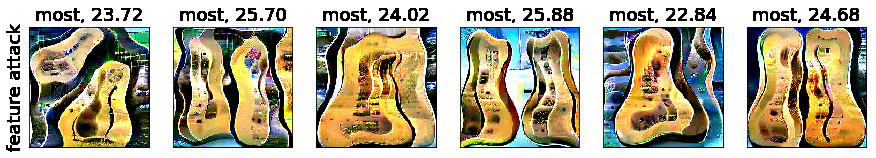}
\end{subfigure}\\
\begin{subfigure}{0.9\linewidth}
\centering
\includegraphics[trim=0cm 0cm 0cm 0.9cm, clip, width=\linewidth]{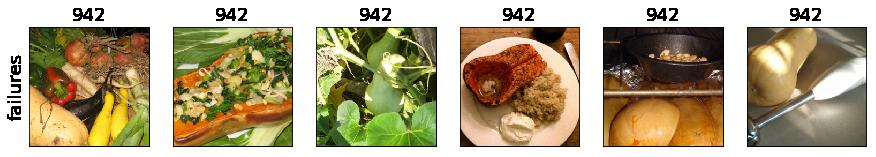}
\end{subfigure}
\caption{Visualization of feature[1905]. For images with \textbf{label butternut squash}, when feature$[1905] < 0.6324$, error rate increases to $0.6034$ (\textcolor{red}{\textbf{+25.96\%}}).}
\label{fig:appendix_robust_label_butternut_squash}
\end{figure*}

\begin{figure*}[t]
\centering
\begin{subfigure}{0.9\linewidth}
\centering
\includegraphics[trim=0cm 0cm 0cm 0.9cm, clip, width=\linewidth]{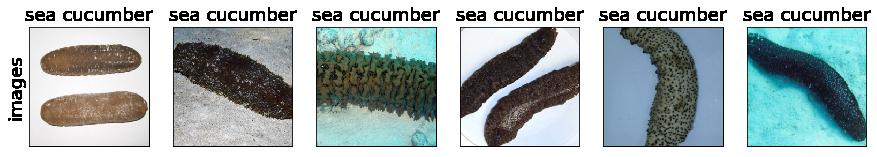}
\end{subfigure}\\
\begin{subfigure}{0.9\linewidth}
\centering
\includegraphics[trim=0cm 0cm 0cm 0.9cm, clip, width=\linewidth]{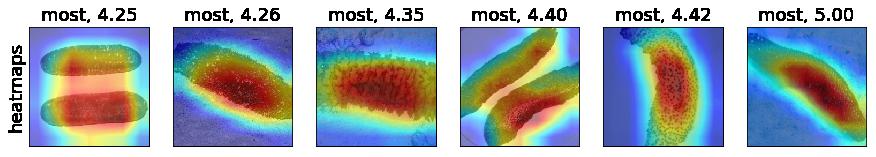}
\end{subfigure}\\
\begin{subfigure}{0.9\linewidth}
\centering
\includegraphics[trim=0cm 0cm 0cm 0.9cm, clip, width=\linewidth]{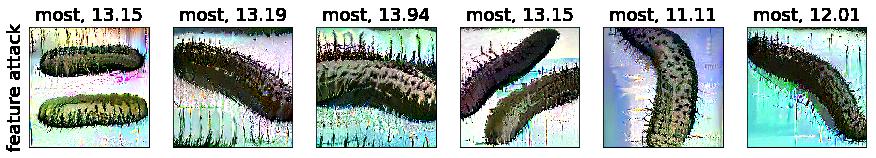}
\end{subfigure}\\
\begin{subfigure}{0.9\linewidth}
\centering
\includegraphics[trim=0cm 0cm 0cm 0.9cm, clip, width=\linewidth]{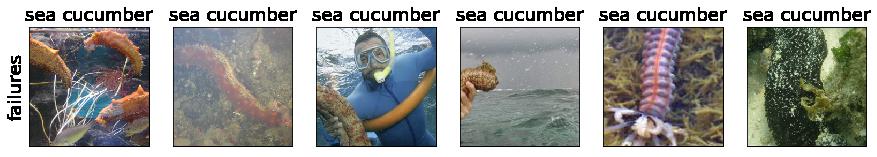}
\end{subfigure}
\caption{Visualization of feature[1147]. For images with \textbf{label sea cucumber}, when feature$[1147] < 1.067$, error rate increases to $0.6042$ (\textcolor{red}{\textbf{+26.04\%}}).}
\label{fig:appendix_robust_label_sea_cucumber}
\end{figure*}

\begin{figure*}[t]
\centering
\begin{subfigure}{0.9\linewidth}
\centering
\includegraphics[trim=0cm 0cm 0cm 0.9cm, clip, width=\linewidth]{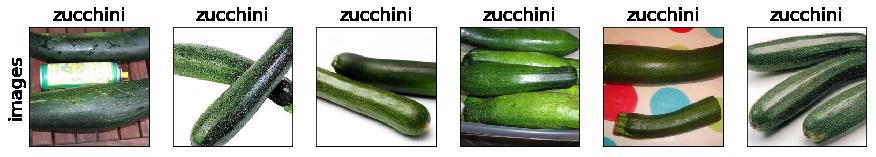}
\end{subfigure}\\
\begin{subfigure}{0.9\linewidth}
\centering
\includegraphics[trim=0cm 0cm 0cm 0.9cm, clip, width=\linewidth]{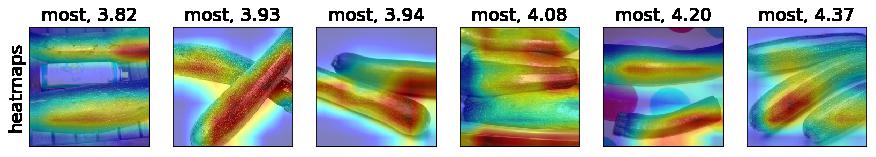}
\end{subfigure}\\
\begin{subfigure}{0.9\linewidth}
\centering
\includegraphics[trim=0cm 0cm 0cm 0.9cm, clip, width=\linewidth]{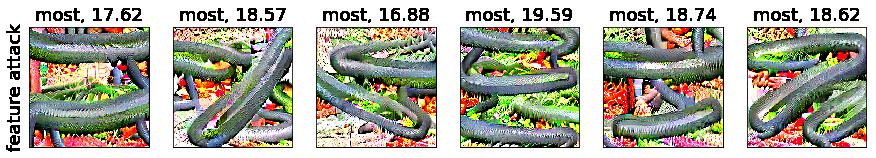}
\end{subfigure}\\
\begin{subfigure}{0.9\linewidth}
\centering
\includegraphics[trim=0cm 0cm 0cm 0.9cm, clip, width=\linewidth]{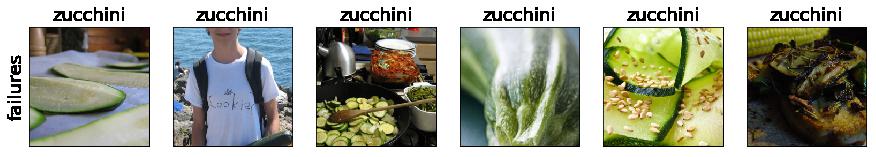}
\end{subfigure}
\caption{Visualization of feature[752]. For images with \textbf{label zucchini}, when feature$[752] < 1.0602$, error rate increases to $0.6445$ (\textcolor{red}{\textbf{+30.14\%}}).}
\label{fig:appendix_robust_label_zucchini}
\end{figure*}

\begin{figure*}[t]
\centering
\begin{subfigure}{0.9\linewidth}
\centering
\includegraphics[trim=0cm 0cm 0cm 0.9cm, clip, width=\linewidth]{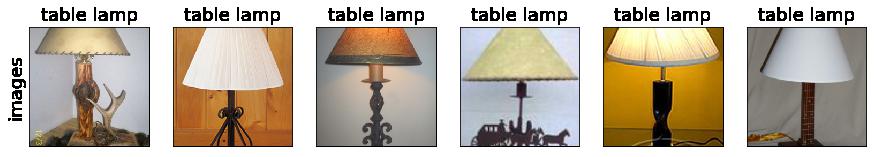}
\end{subfigure}\\
\begin{subfigure}{0.9\linewidth}
\centering
\includegraphics[trim=0cm 0cm 0cm 0.9cm, clip, width=\linewidth]{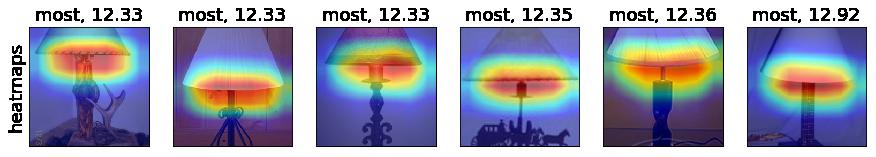}
\end{subfigure}\\
\begin{subfigure}{0.9\linewidth}
\centering
\includegraphics[trim=0cm 0cm 0cm 0.9cm, clip, width=\linewidth]{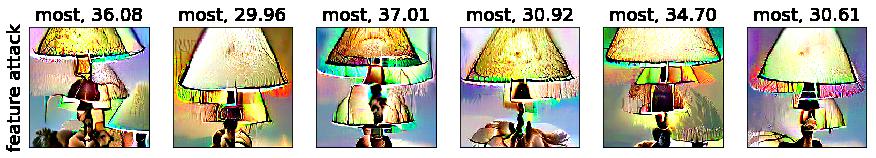}
\end{subfigure}\\
\begin{subfigure}{0.9\linewidth}
\centering
\includegraphics[trim=0cm 0cm 0cm 0.9cm, clip, width=\linewidth]{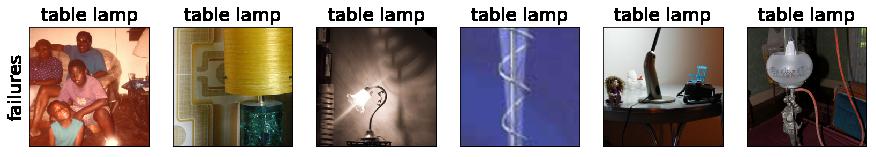}
\end{subfigure}
\caption{Visualization of feature[752]. For images with \textbf{label table lamp}, when feature$[1740] < 2.5241$, error rate increases to $0.7251$ (\textcolor{red}{\textbf{+38.28\%}}).}
\label{fig:appendix_robust_label_table_lamp}
\end{figure*}

\subsubsection{Grouping by prediction}
Results are in Table \ref{table:robust_pred_grouping}.

\begin{figure*}[t]
\centering
\begin{subfigure}{0.9\linewidth}
\centering
\includegraphics[trim=0cm 0cm 0cm 0.9cm, clip, width=\linewidth]{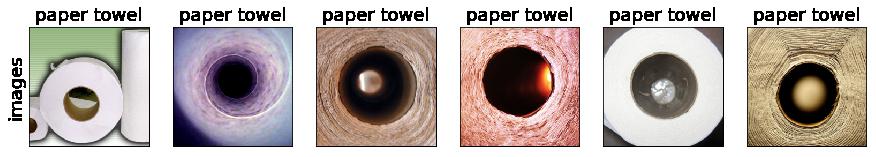}
\end{subfigure}\\
\begin{subfigure}{0.9\linewidth}
\centering
\includegraphics[trim=0cm 0cm 0cm 0.9cm, clip, width=\linewidth]{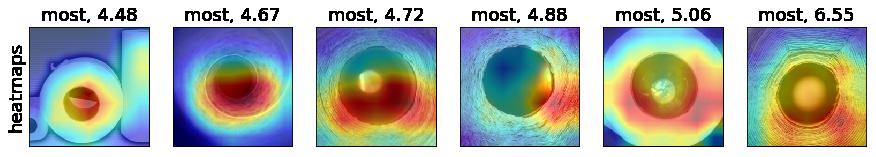}
\end{subfigure}\\
\begin{subfigure}{0.9\linewidth}
\centering
\includegraphics[trim=0cm 0cm 0cm 0.9cm, clip, width=\linewidth]{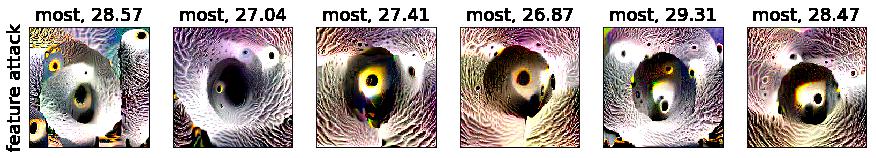}
\end{subfigure}\\
\begin{subfigure}{0.9\linewidth}
\centering
\includegraphics[trim=0cm 0cm 0cm 0.9cm, clip, width=\linewidth]{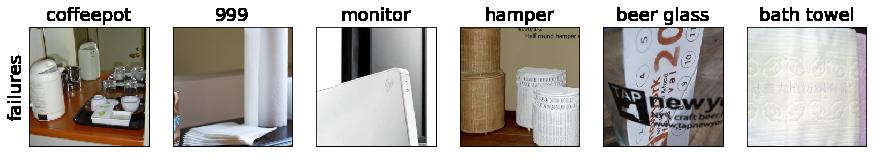}
\end{subfigure}
\caption{Visualization of feature[1412]. For images with \textbf{prediction paper towel}, when feature$[1412] < 1.2665$, error rate increases to $0.7825$ (\textcolor{red}{\textbf{+15.15\%}}).}
\label{fig:appendix_robust_pred_paper_towel}
\end{figure*}

\begin{table*}
\centering
\begin{tabular}{| p{1.7cm} || p{1cm} | p{1.5cm} | p{1cm} | p{1cm} | p{1cm} | p{1cm} | p{1.9cm} | p{3cm} |}
\hline
\textbf{Class name} & \textbf{Feature index} & \textbf{Decision rule} & \textbf{\baseerrorratetext} & \textbf{\errorratetext} & \textbf{\errorcoveragetext} & \textbf{\avgtreetext} &\textbf{Feature} \newline \textbf{visualization} & \textbf{Feature name\newline (from visualization)}\\
\hline
paper towel & 1412 & $ < 1.2665 $ & $0.6310$ & $0.7825$ &  $0.6566$ & $0.6720$ & Figure \ref{fig:appendix_robust_pred_paper_towel} & cylindrical hole\\
\hline
seat belt & 1493 & $ < 1.0624$ & $0.5983$ & $0.7402$ & $0.5816$ & $0.6282$ & Figure \ref{fig:appendix_robust_pred_seat_belt} & window \\
\hline
crutch & 502 & $ < 0.7458 $ & $0.5842$ & $0.7302$ & $0.7233$ & $0.6343$ & Figure \ref{fig:appendix_robust_pred_crutch} & rods \\
\hline
lumbermill & 56 & $ < 0.5817 $ & $0.5625$ & $0.7049$ & $0.4362$ & $0.5817$ & Figure \ref{fig:appendix_robust_pred_lumbermill} & tracks \\
\hline
bassoon & 1104 & $ < 0.7026$ & $0.5621$ & $0.7490$ & $0.6474$ & $0.6208$ & Figure \ref{fig:appendix_robust_pred_bassoon} & hands and cylindrical bassoon \\
\hline
impala & 918 & $ < 0.9435 $ & $0.3535$ & $0.6298$ & $0.5917$ & $ 0.4609$ & Figure \ref{fig:appendix_robust_pred_impala} & close-up face \\
\hline
boxer & 404 & $ < 1.6458$ & $0.3527$ & $0.4991$ & $0.7671$ & $0.4246$ & Figure \ref{fig:appendix_robust_pred_boxer} & dog nose \\
\hline
samoyed & 1694 & $ < 0.7492$ & $0.3487$ & $0.5304$ & $0.6247$ & $0.4147$ & Figure \ref{fig:appendix_robust_pred_samoyed} & close-up dog face\\
\hline
milk can & 676 & $ < 1.1286$ & $0.3530$ & $0.6284$ & $0.6300$ & $0.4707$ & Figure \ref{fig:appendix_robust_pred_milk_can} & horizontal edges \\
\hline
gasmask & 835 & $ < 0.9034 $ & $0.3521$ & $0.6216$ & $0.5736$ & $0.4514$ & Figure \ref{fig:appendix_robust_pred_gasmask} & round patches \\
\hline
king crab & 952 & $ < 2.9012 $ & $0.3487$ & $0.5991$ & $0.5770$ & $0.4396$ & Figure \ref{fig:appendix_robust_pred_king_crab} & crab tentacles \\
\hline
\end{tabular}
\caption{Results on a robust Resnet-50 model using grouping by prediction.}
\label{table:robust_pred_grouping}
\end{table*}

\begin{figure*}[t]
\centering
\begin{subfigure}{0.9\linewidth}
\centering
\includegraphics[trim=0cm 0cm 0cm 0.9cm, clip, width=\linewidth]{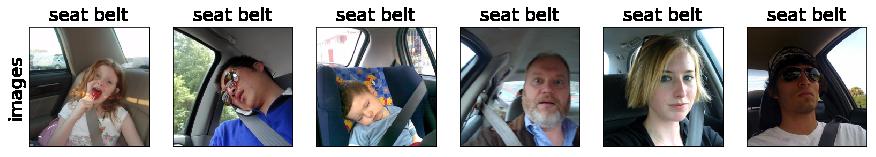}
\end{subfigure}\\
\begin{subfigure}{0.9\linewidth}
\centering
\includegraphics[trim=0cm 0cm 0cm 0.9cm, clip, width=\linewidth]{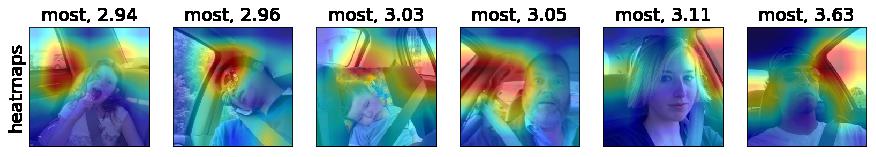}
\end{subfigure}\\
\begin{subfigure}{0.9\linewidth}
\centering
\includegraphics[trim=0cm 0cm 0cm 0.9cm, clip, width=\linewidth]{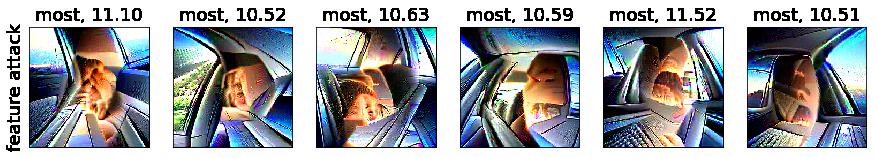}
\end{subfigure}\\
\begin{subfigure}{0.9\linewidth}
\centering
\includegraphics[trim=0cm 0cm 0cm 0.9cm, clip, width=\linewidth]{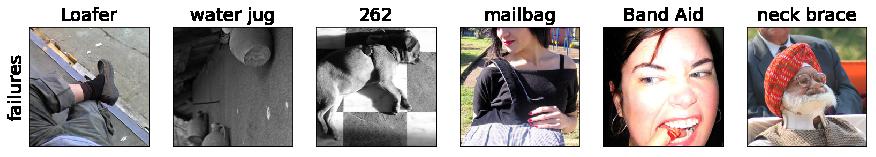}
\end{subfigure}
\caption{Visualization of feature[1493]. For images with \textbf{prediction seat belt}, when feature$[1493] < 1.0624$, error rate increases to $0.7402$ (\textcolor{red}{\textbf{+14.19\%}}).}
\label{fig:appendix_robust_pred_seat_belt}
\end{figure*}

\begin{figure*}
\centering
\begin{subfigure}{0.9\linewidth}
\centering
\includegraphics[trim=0cm 0cm 0cm 0.9cm, clip, width=\linewidth]{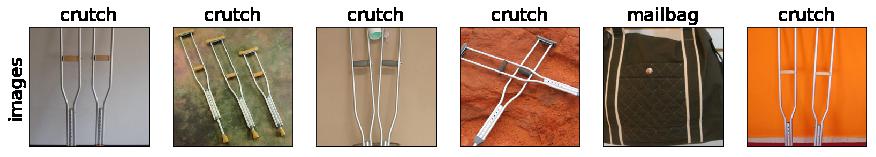}
\end{subfigure}\\
\begin{subfigure}{0.9\linewidth}
\centering
\includegraphics[trim=0cm 0cm 0cm 0.9cm, clip, width=\linewidth]{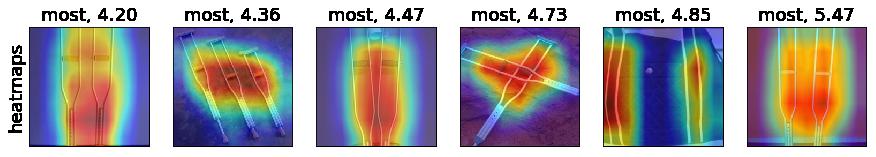}
\end{subfigure}\\
\begin{subfigure}{0.9\linewidth}
\centering
\includegraphics[trim=0cm 0cm 0cm 0.9cm, clip, width=\linewidth]{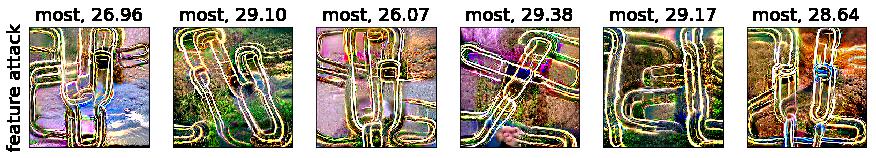}
\end{subfigure}\\
\begin{subfigure}{0.9\linewidth}
\centering
\includegraphics[trim=0cm 0cm 0cm 0.9cm, clip, width=\linewidth]{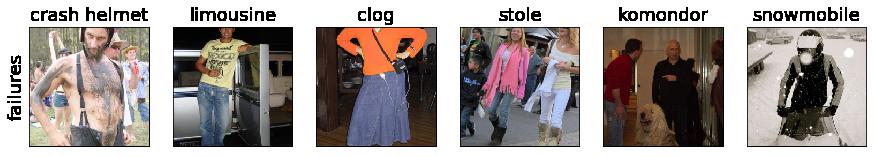}
\end{subfigure}
\caption{Visualization of feature[502]. For images with \textbf{prediction crutch}, when feature$[502] < 0.7458$, error rate increases to $0.7302$ (\textcolor{red}{\textbf{+14.60\%}}).}
\label{fig:appendix_robust_pred_crutch}
\end{figure*}

\begin{figure*}
\centering
\begin{subfigure}{0.9\linewidth}
\centering
\includegraphics[trim=0cm 0cm 0cm 0.9cm, clip, width=\linewidth]{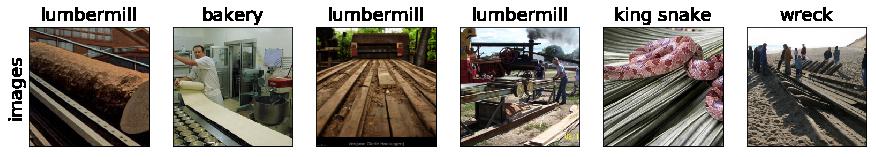}
\end{subfigure}\\
\begin{subfigure}{0.9\linewidth}
\centering
\includegraphics[trim=0cm 0cm 0cm 0.9cm, clip, width=\linewidth]{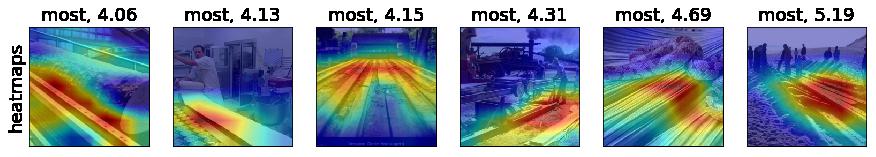}
\end{subfigure}\\
\begin{subfigure}{0.9\linewidth}
\centering
\includegraphics[trim=0cm 0cm 0cm 0.9cm, clip, width=\linewidth]{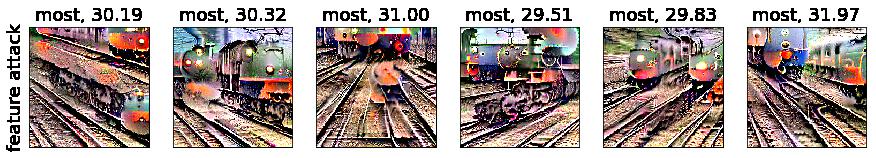}
\end{subfigure}\\
\begin{subfigure}{0.9\linewidth}
\centering
\includegraphics[trim=0cm 0cm 0cm 0.9cm, clip, width=\linewidth]{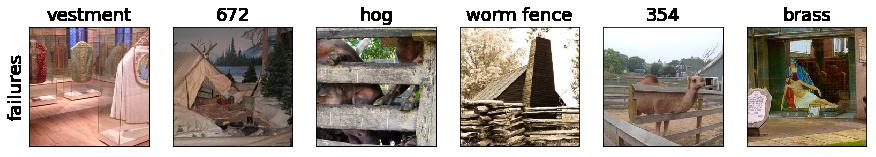}
\end{subfigure}
\caption{Visualization of feature[56]. For images with \textbf{prediction lumberhill}, when feature$[56] < 0.5817$, error rate increases to $0.7049$ (\textcolor{red}{\textbf{+14.24\%}}).}
\label{fig:appendix_robust_pred_lumbermill}
\end{figure*}

\begin{figure*}
\centering
\begin{subfigure}{0.9\linewidth}
\centering
\includegraphics[trim=0cm 0cm 0cm 0.9cm, clip, width=\linewidth]{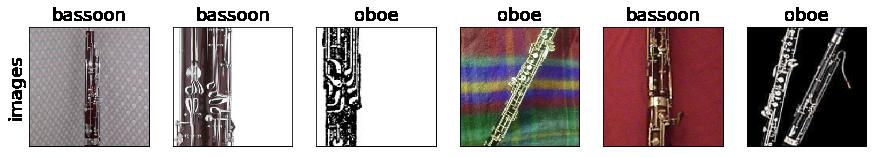}
\end{subfigure}\\
\begin{subfigure}{0.9\linewidth}
\centering
\includegraphics[trim=0cm 0cm 0cm 0.9cm, clip, width=\linewidth]{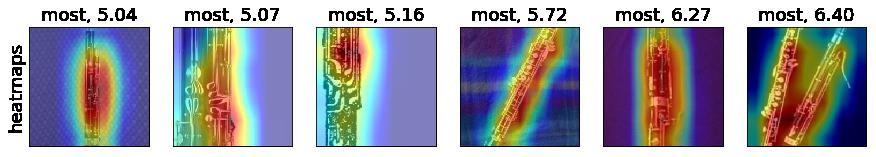}
\end{subfigure}\\
\begin{subfigure}{0.9\linewidth}
\centering
\includegraphics[trim=0cm 0cm 0cm 0.9cm, clip, width=\linewidth]{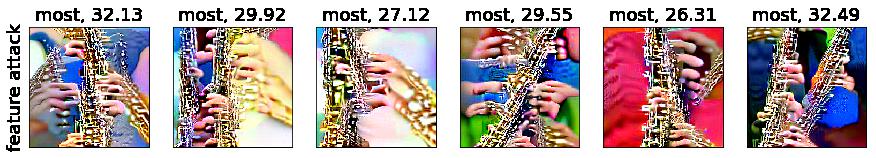}
\end{subfigure}\\
\begin{subfigure}{0.9\linewidth}
\centering
\includegraphics[trim=0cm 0cm 0cm 0.9cm, clip, width=\linewidth]{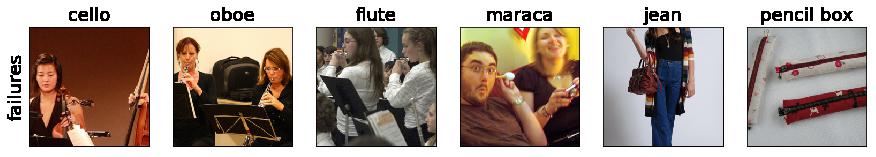}
\end{subfigure}
\caption{Visualization of feature[1104]. For images with \textbf{prediction bassoon}, when feature$[1104] < 0.7026$, error rate increases to $0.7490$ (\textcolor{red}{\textbf{+18.49\%}}).}
\label{fig:appendix_robust_pred_bassoon}
\end{figure*}

\begin{figure*}
\centering
\begin{subfigure}{0.9\linewidth}
\centering
\includegraphics[trim=0cm 0cm 0cm 0.9cm, clip, width=\linewidth]{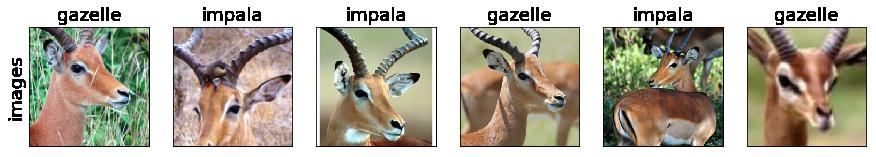}
\end{subfigure}\\
\begin{subfigure}{0.9\linewidth}
\centering
\includegraphics[trim=0cm 0cm 0cm 0.9cm, clip, width=\linewidth]{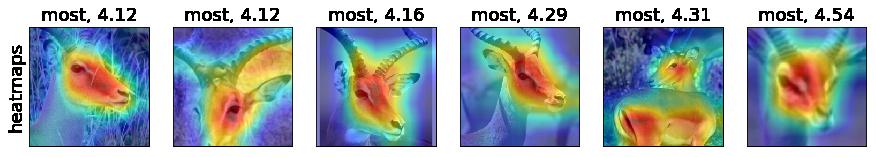}
\end{subfigure}\\
\begin{subfigure}{0.9\linewidth}
\centering
\includegraphics[trim=0cm 0cm 0cm 0.9cm, clip, width=\linewidth]{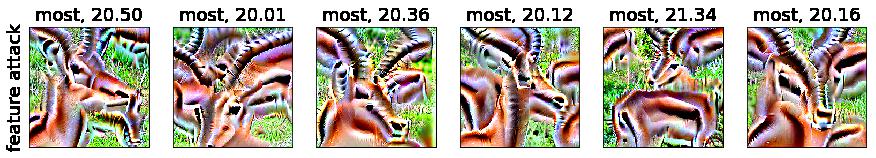}
\end{subfigure}\\
\begin{subfigure}{0.9\linewidth}
\centering
\includegraphics[trim=0cm 0cm 0cm 0.9cm, clip, width=\linewidth]{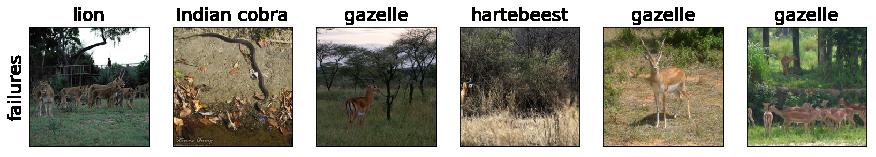}
\end{subfigure}
\caption{Visualization of feature[918]. For images with \textbf{prediction impala}, when feature$[918] < 0.9435$, error rate increases to $0.6298$ (\textcolor{red}{\textbf{+27.63\%}}).}
\label{fig:appendix_robust_pred_impala}
\end{figure*}

\begin{figure*}
\centering
\begin{subfigure}{0.9\linewidth}
\centering
\includegraphics[trim=0cm 0cm 0cm 0.9cm, clip, width=\linewidth]{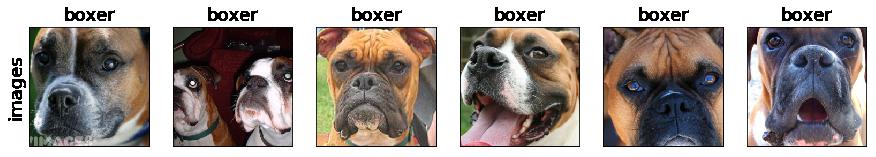}
\end{subfigure}\\
\begin{subfigure}{0.9\linewidth}
\centering
\includegraphics[trim=0cm 0cm 0cm 0.9cm, clip, width=\linewidth]{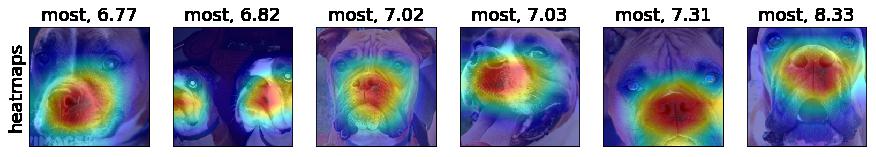}
\end{subfigure}\\
\begin{subfigure}{0.9\linewidth}
\centering
\includegraphics[trim=0cm 0cm 0cm 0.9cm, clip, width=\linewidth]{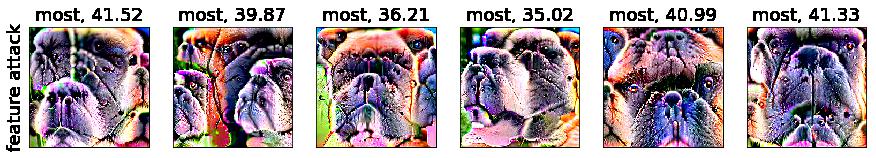}
\end{subfigure}\\
\begin{subfigure}{0.9\linewidth}
\centering
\includegraphics[trim=0cm 0cm 0cm 0.9cm, clip, width=\linewidth]{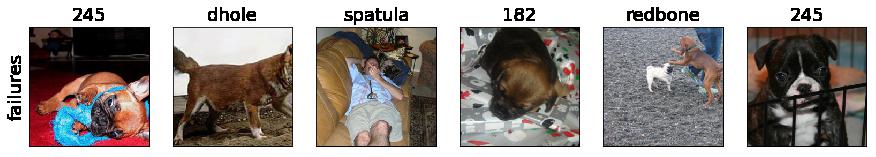}
\end{subfigure}
\caption{Visualization of feature[404]. For images with \textbf{prediction boxer}, when feature$[404] < 1.6458$, error rate increases to $0.4991$ (\textcolor{red}{\textbf{+14.64\%}}).}
\label{fig:appendix_robust_pred_boxer}
\end{figure*}

\begin{figure*}
\centering
\begin{subfigure}{0.9\linewidth}
\centering
\includegraphics[trim=0cm 0cm 0cm 0.9cm, clip, width=\linewidth]{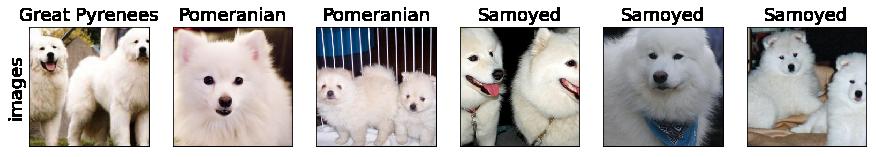}
\end{subfigure}\\
\begin{subfigure}{0.9\linewidth}
\centering
\includegraphics[trim=0cm 0cm 0cm 0.9cm, clip, width=\linewidth]{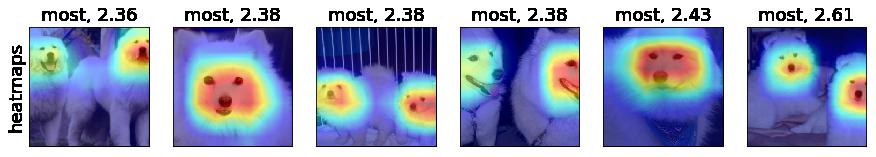}
\end{subfigure}\\
\begin{subfigure}{0.9\linewidth}
\centering
\includegraphics[trim=0cm 0cm 0cm 0.9cm, clip, width=\linewidth]{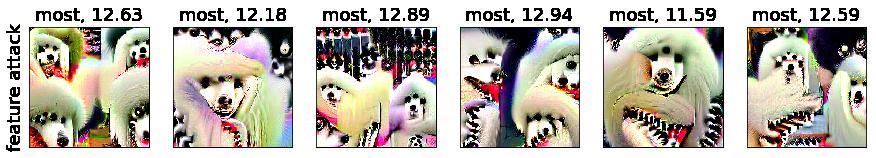}
\end{subfigure}\\
\begin{subfigure}{0.9\linewidth}
\centering
\includegraphics[trim=0cm 0cm 0cm 0.9cm, clip, width=\linewidth]{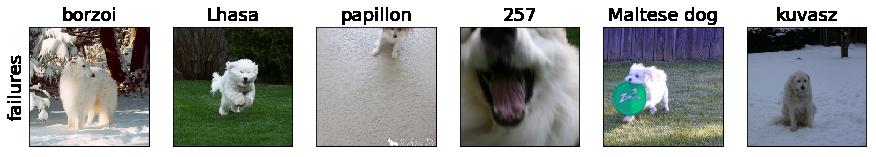}
\end{subfigure}
\caption{Visualization of feature[1694]. For images with \textbf{prediction samoyed}, when feature$[1694] < 0.7492$, error rate increases to $0.5304$ (\textcolor{red}{\textbf{+18.17\%}}).}
\label{fig:appendix_robust_pred_samoyed}
\end{figure*}

\begin{figure*}
\centering
\begin{subfigure}{0.9\linewidth}
\centering
\includegraphics[trim=0cm 0cm 0cm 0.9cm, clip, width=\linewidth]{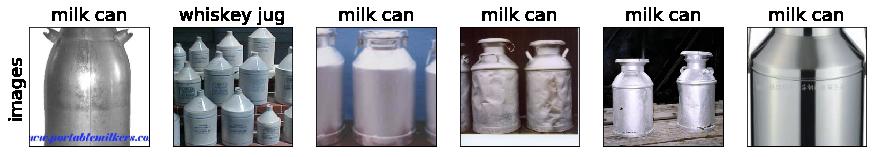}
\end{subfigure}\\
\begin{subfigure}{0.9\linewidth}
\centering
\includegraphics[trim=0cm 0cm 0cm 0.9cm, clip, width=\linewidth]{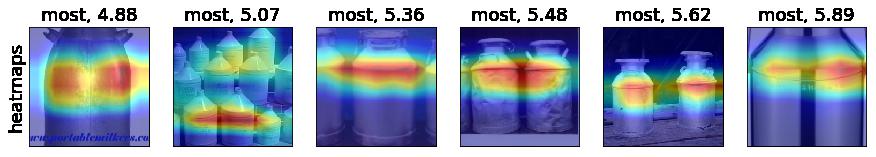}
\end{subfigure}\\
\begin{subfigure}{0.9\linewidth}
\centering
\includegraphics[trim=0cm 0cm 0cm 0.9cm, clip, width=\linewidth]{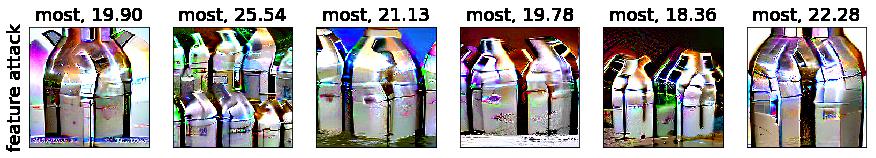}
\end{subfigure}\\
\begin{subfigure}{0.9\linewidth}
\centering
\includegraphics[trim=0cm 0cm 0cm 0.9cm, clip, width=\linewidth]{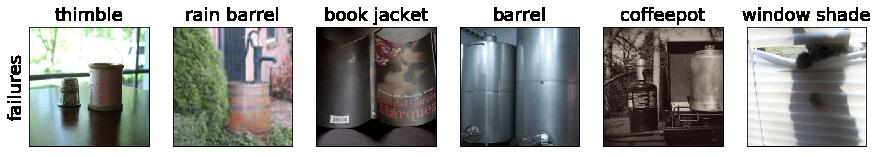}
\end{subfigure}
\caption{Visualization of feature[676]. For images with \textbf{prediction milk can}, when feature$[676] < 1.1286$, error rate increases to $0.6284$ (\textcolor{red}{\textbf{+27.54\%}}).}
\label{fig:appendix_robust_pred_milk_can}
\end{figure*}

\begin{figure*}
\centering
\begin{subfigure}{0.9\linewidth}
\centering
\includegraphics[trim=0cm 0cm 0cm 0.9cm, clip, width=\linewidth]{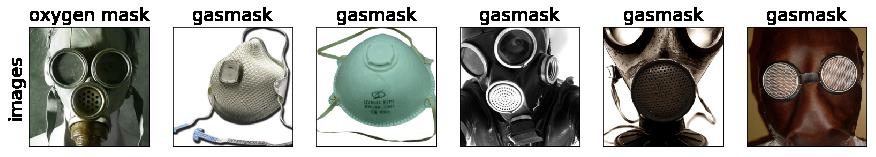}
\end{subfigure}\\
\begin{subfigure}{0.9\linewidth}
\centering
\includegraphics[trim=0cm 0cm 0cm 0.9cm, clip, width=\linewidth]{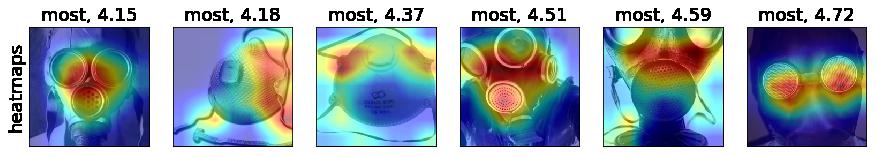}
\end{subfigure}\\
\begin{subfigure}{0.9\linewidth}
\centering
\includegraphics[trim=0cm 0cm 0cm 0.9cm, clip, width=\linewidth]{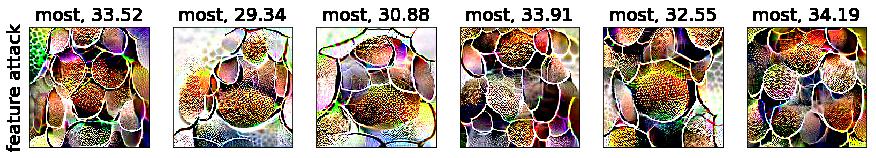}
\end{subfigure}\\
\begin{subfigure}{0.9\linewidth}
\centering
\includegraphics[trim=0cm 0cm 0cm 0.9cm, clip, width=\linewidth]{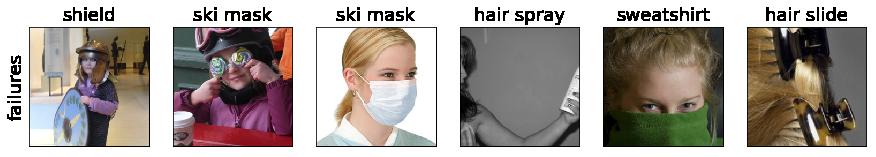}
\end{subfigure}
\caption{Visualization of feature[835]. For images with \textbf{prediction gasmask}, when feature$[835] < 0.9034$, error rate increases to $0.6216$ (\textcolor{red}{\textbf{+26.95\%}}).}
\label{fig:appendix_robust_pred_gasmask}

\end{figure*}

\begin{figure*}
\centering
\begin{subfigure}{0.9\linewidth}
\centering
\includegraphics[trim=0cm 0cm 0cm 0.9cm, clip, width=\linewidth]{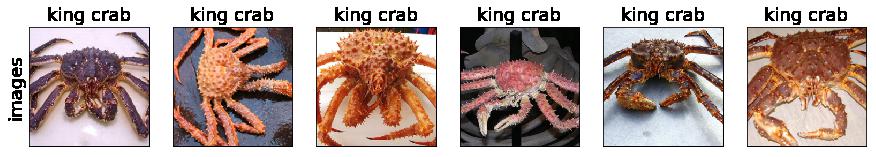}
\end{subfigure}\\
\begin{subfigure}{0.9\linewidth}
\centering
\includegraphics[trim=0cm 0cm 0cm 0.9cm, clip, width=\linewidth]{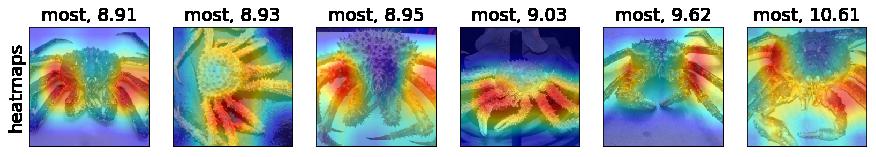}
\end{subfigure}\\
\begin{subfigure}{0.9\linewidth}
\centering
\includegraphics[trim=0cm 0cm 0cm 0.9cm, clip, width=\linewidth]{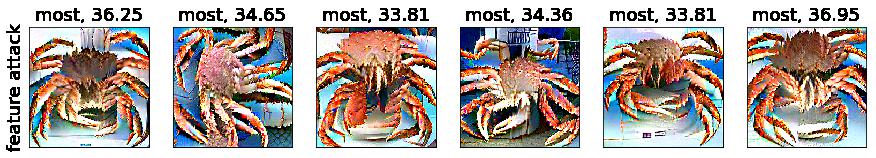}
\end{subfigure}\\
\begin{subfigure}{0.9\linewidth}
\centering
\includegraphics[trim=0cm 0cm 0cm 0.9cm, clip, width=\linewidth]{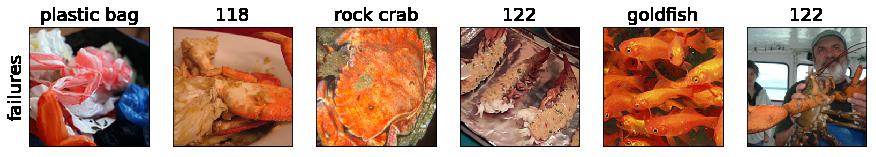}
\end{subfigure}
\caption{Visualization of feature[952]. For images with \textbf{prediction king crab}, when feature$[952] < 2.9012$, error rate increases to $0.5991$ (\textcolor{red}{\textbf{+25.04\%}}).}
\label{fig:appendix_robust_pred_king_crab}
\end{figure*}

\clearpage
\newpage 

\onecolumn

\section{Examples from Crowd study}\label{sec:appendix_mturk_examples}

\begin{figure*}[t]
\centering
\includegraphics[width=\linewidth]{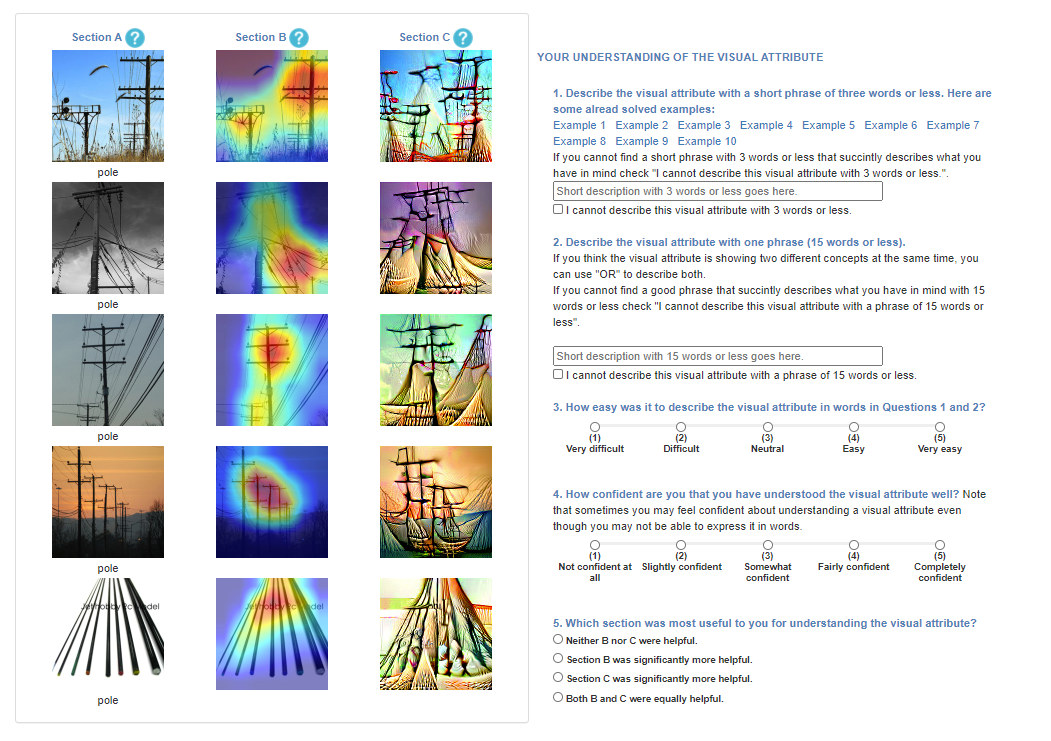}
\caption{Amazon Mechanical Turk questionnaire.}
\label{fig:aws_questions}
\end{figure*}

\begin{figure}[t]
\centering
\includegraphics[trim=0.5cm 0cm 0.5cm 1.2cm, clip,width=0.4\linewidth]{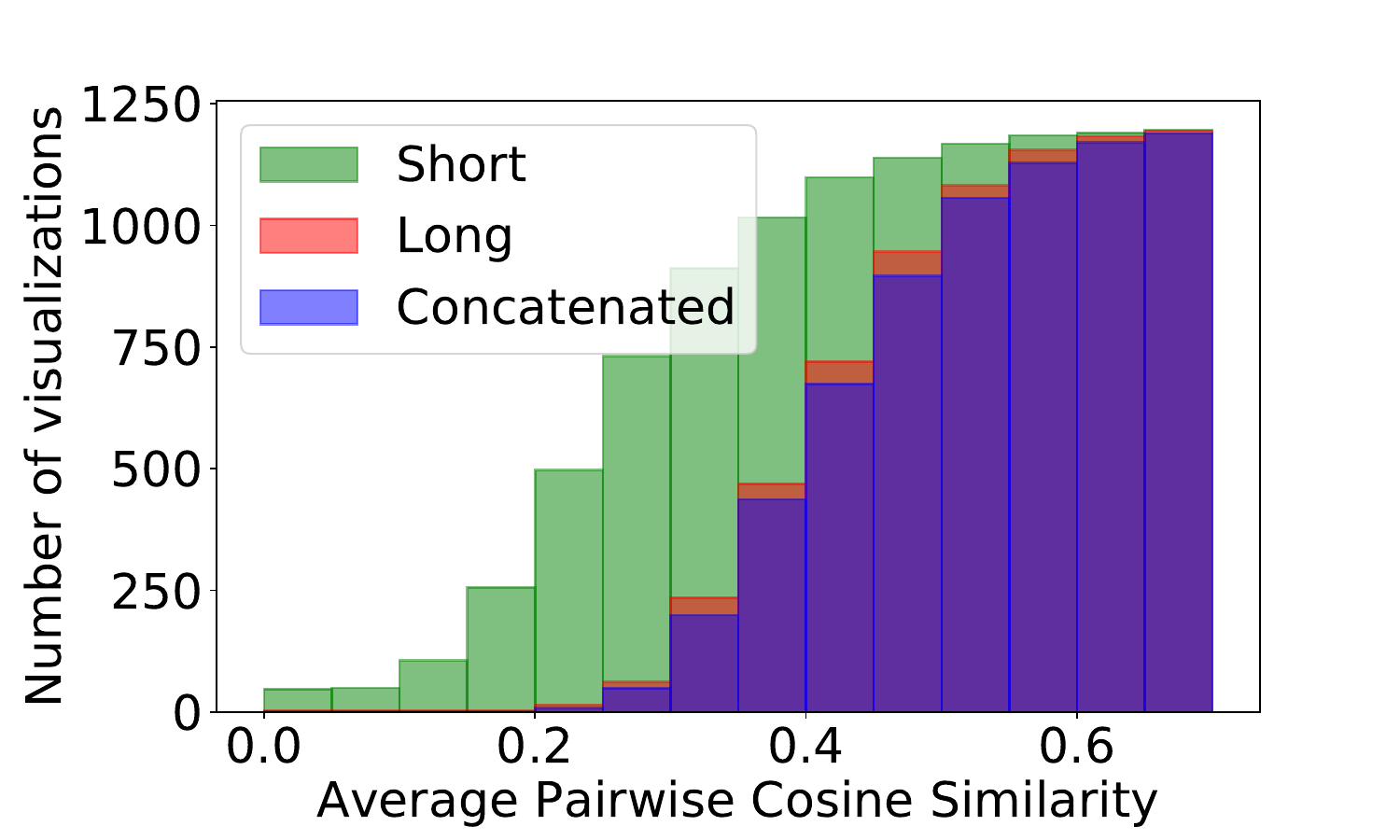}
\caption{Cumulative distribution of worker agreement on the textual feature descriptions in the crowd study.}
\label{fig:amt_study_agreement}
\end{figure}

The questionnaire for the Crowd study is shown in Figure \ref{fig:aws_questions}. For further investigation on the quality of the answers given in Question 1 and 2 of the questionnaire (short and long description), we also compute agreement scores between the answers. Figure~\ref{fig:amt_study_agreement} shows the cumulative distribution of worker agreement on the textual feature descriptions (i.e., short $\leq$ 3-word descriptions, long $\leq$ 15-word descriptions, and concatenated). We use the Word2Vec embedding (trained on the Google News corpus) to compute word vectors. The vector of each description is computed as the average of the vectors of all words in the description that are not stop words. We then compute worker inter-agreement as the pairwise average cosine similarity between the description vectors. As opposed to $n$-gram agreement definitions, the score can capture common themes in descriptions even when workers use different words but with similar meaning (e.g., digit vs. number). We observe that agreement increases with longer descriptions. Qualitatively, we see that agreement is higher ($\geq 0.45$) when the images in the visualization contain fewer objects and the objects are salient. Sample descriptions from workers along with agreement scores can be found in the following examples in Appendix Section \ref{subsec:appendix_mturk_easy_examples} and \ref{subsec:appendix_mturk_hard_examples}.

\subsection{Easy examples (Table \ref{table:mturk_easy})}\label{subsec:appendix_mturk_easy_examples}

\begin{table*}[h]
\centering
\begin{tabular}{| p{4cm} | p{1cm} | p{1cm} | p{2cm} | p{2cm} | p{2cm} | p{2cm} |  }
\hline
\multirow{3}{2cm}{\textbf{Class name}} & \multirow{3}{1cm}{\textbf{Class index}} & \multirow{3}{1cm}{\textbf{Feature index}} & \multirow{3}{1cm}{\textbf{Grouping}} & \multirow{3}{2cm}{\textbf{Feature \newline visualization}} & 
\multicolumn{2}{c|}{\textbf{Cosine similarity}} \\
\cline{6-7}
& & & & & \multirow{2}{2cm}{Short description} & \multirow{2}{2cm}{Long description} \\
& & & & & &  \\
\hline
malinois & 225 & 813 & prediction & Figure \ref{fig:appendix_mturk_easy1} & 0.3368 & 0.4551 \\
\hline
greenhouse, nursery, glasshouse & \multirow{1}{1cm}{580} & \multirow{1}{1cm}{1933} & \multirow{1}{1cm}{prediction} & \multirow{1}{2cm}{Figure \ref{fig:appendix_mturk_easy2}} & \multirow{1}{1cm}{0.4094} & \multirow{1}{1cm}{0.5354}\\
\hline
black and gold garden spider, Argiope aurantia & \multirow{1}{1cm}{72} & \multirow{1}{1cm}{652} & \multirow{1}{1cm}{prediction} & \multirow{1}{2cm}{Figure \ref{fig:appendix_mturk_easy3}} & \multirow{1}{1cm}{0.1795} & \multirow{1}{1cm}{0.3162} \\
\hline
scuba diver & 983 & 1588 & prediction & Figure \ref{fig:appendix_mturk_easy4} & 0.0 & 0.3753 \\
\hline
sea cucumber, holothurian & \multirow{1}{1cm}{329} & \multirow{1}{1cm}{28} & \multirow{1}{1cm}{prediction} & \multirow{1}{2cm}{Figure \ref{fig:appendix_mturk_easy5}}  & \multirow{1}{1cm}{0.2680} & \multirow{1}{1cm}{0.4107} \\
\hline
\end{tabular}
\caption{Examples from the Amazon Mechanical Turk study that workers found as easy to describe. Short description is the answer to Q1 and Long description is the answer to Q2 in the crowd study (Figure \ref{fig:aws_questions}).}
\label{table:mturk_easy}
\end{table*}

\begin{figure*}
\captionsetup{width=.7\linewidth}
\centering
\begin{subfigure}{0.8\linewidth}
\centering
\includegraphics[trim=0cm 0cm 0cm 0.9cm, clip, width=\linewidth]{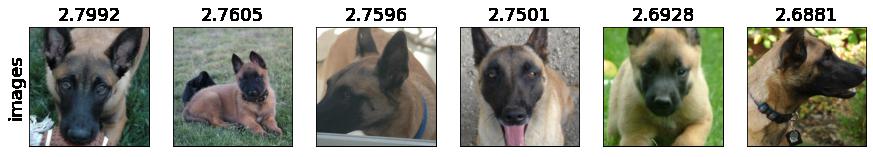}
\end{subfigure}\\
\begin{subfigure}{0.8\linewidth}
\centering
\includegraphics[trim=0cm 0cm 0cm 0.9cm, clip, width=\linewidth]{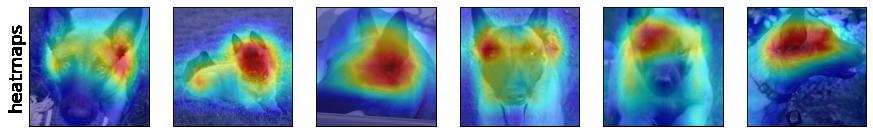}
\end{subfigure}\\
\begin{subfigure}{0.8\linewidth}
\centering
\includegraphics[trim=0cm 0cm 0cm 0.9cm, clip, width=\linewidth]{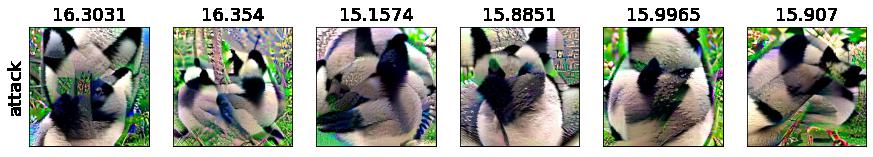}
\end{subfigure}
\centering
\caption{Visualization of feature[813], class[225] (malinois) and prediction grouping. \\Example descriptions: black fur, Canid eyes, facial fur, black and white, head region}
\label{fig:appendix_mturk_easy1}
\end{figure*}

\begin{figure*}
\centering
\begin{subfigure}{0.8\linewidth}
\centering
\includegraphics[trim=0cm 0cm 0cm 0.9cm, clip, width=\linewidth]{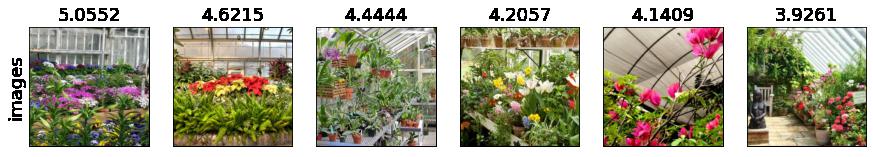}
\end{subfigure}\\
\begin{subfigure}{0.8\linewidth}
\centering
\includegraphics[trim=0cm 0cm 0cm 0.9cm, clip, width=\linewidth]{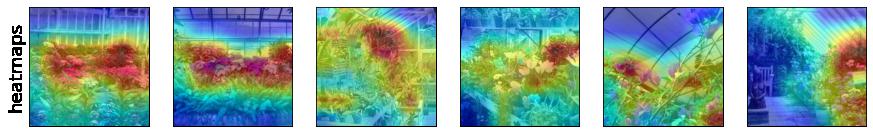}
\end{subfigure}\\
\begin{subfigure}{0.8\linewidth}
\centering
\includegraphics[trim=0cm 0cm 0cm 0.9cm, clip, width=\linewidth]{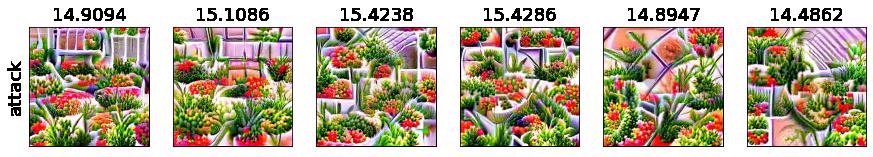}
\end{subfigure}
\captionsetup{width=.8\linewidth}
\caption{Visualization of feature[1933], class[580] (greenhouse) and prediction grouping. \\Example descriptions: plant, colorful flowers, leafy greens, bunch of plants, plant}
\label{fig:appendix_mturk_easy2}
\end{figure*}

\begin{figure*}
\centering
\begin{subfigure}{0.8\linewidth}
\centering
\includegraphics[trim=0cm 0cm 0cm 0.9cm, clip, width=\linewidth]{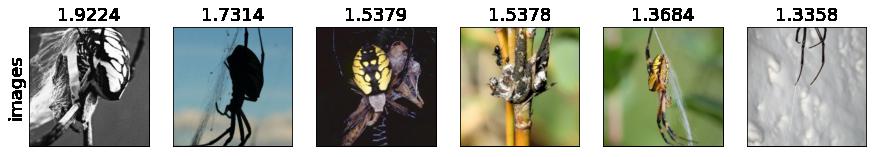}
\end{subfigure}\\
\begin{subfigure}{0.8\linewidth}
\centering
\includegraphics[trim=0cm 0cm 0cm 0.9cm, clip, width=\linewidth]{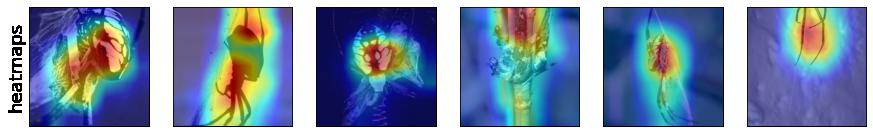}
\end{subfigure}\\
\begin{subfigure}{0.8\linewidth}
\centering
\includegraphics[trim=0cm 0cm 0cm 0.9cm, clip, width=\linewidth]{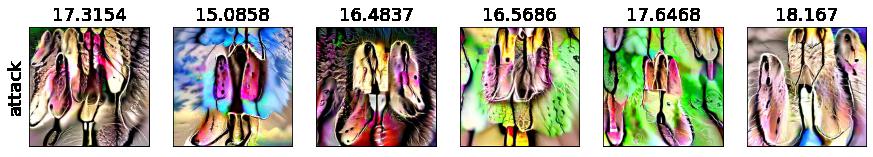}
\end{subfigure}
\captionsetup{width=.8\linewidth}
\caption{Visualization of feature[652], class[72] (argiope aurantia) and prediction grouping. \\Example descriptions: branching forms, shoes, body of creature, exotic arachnid, black color}
\label{fig:appendix_mturk_easy3}
\end{figure*}

\begin{figure*}
\centering
\begin{subfigure}{0.8\linewidth}
\centering
\includegraphics[trim=0cm 0cm 0cm 0.9cm, clip, width=\linewidth]{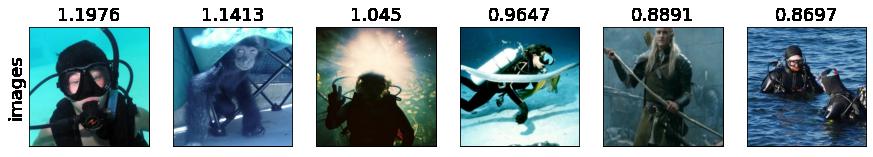}
\end{subfigure}\\
\begin{subfigure}{0.8\linewidth}
\centering
\includegraphics[trim=0cm 0cm 0cm 0.9cm, clip, width=\linewidth]{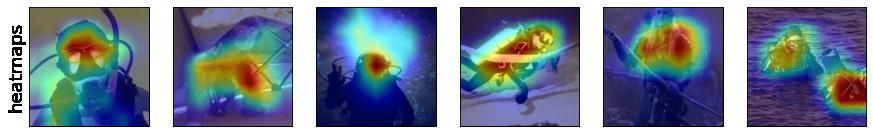}
\end{subfigure}\\
\begin{subfigure}{0.8\linewidth}
\centering
\includegraphics[trim=0cm 0cm 0cm 0.9cm, clip, width=\linewidth]{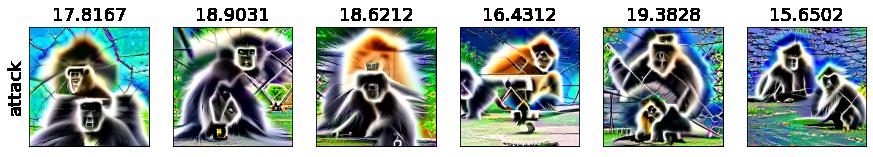}
\end{subfigure}
\captionsetup{width=.8\linewidth}
\caption{Visualization of feature[1588], class[983] (scuba diver) and prediction grouping. \\Example descriptions: tube or human, glowing faces, black, monkey-like, square face}
\label{fig:appendix_mturk_easy4}
\end{figure*}

\begin{figure*}
\centering
\begin{subfigure}{0.8\linewidth}
\centering
\includegraphics[trim=0cm 0cm 0cm 0.9cm, clip, width=\linewidth]{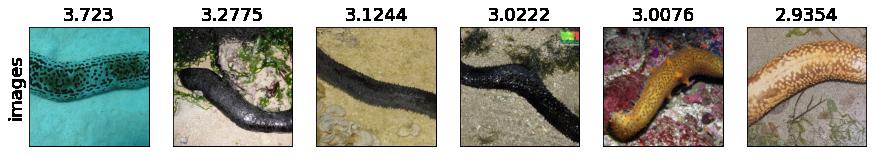}
\end{subfigure}\\
\begin{subfigure}{0.8\linewidth}
\centering
\includegraphics[trim=0cm 0cm 0cm 0.9cm, clip, width=\linewidth]{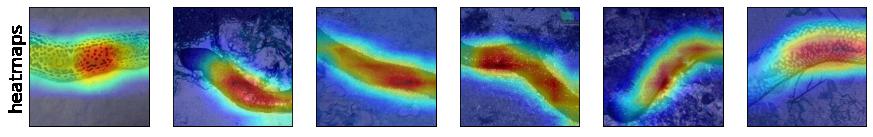}
\end{subfigure}\\
\begin{subfigure}{0.8\linewidth}
\centering
\includegraphics[trim=0cm 0cm 0cm 0.9cm, clip, width=\linewidth]{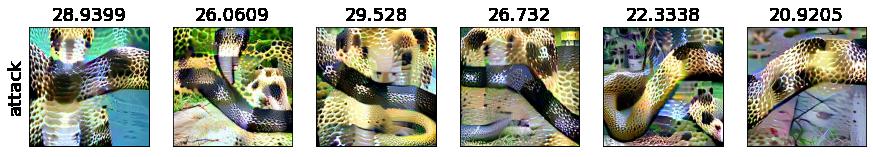}
\end{subfigure}
\captionsetup{width=.8\linewidth}
\caption{Visualization of feature[28], class[329] (sea cucumber) and prediction grouping. \\Example descriptions: spots, rainbow, tubular sea creature, Tube, Tubular organism belly}
\label{fig:appendix_mturk_easy5}
\end{figure*}

\clearpage
\subsection{Difficult examples (Table \ref{table:mturk_hard})}\label{subsec:appendix_mturk_hard_examples}

\begin{table*}[h!]
\centering
\begin{tabular}{| p{4.5cm} | p{1cm} | p{1cm} | p{2cm} | p{2cm} | p{2cm} | p{2cm} |  }
\hline
\multirow{3}{2cm}{\textbf{Class name}} & \multirow{3}{1cm}{\textbf{Class index}} & \multirow{3}{1cm}{\textbf{Feature index}} & \multirow{3}{1cm}{\textbf{Grouping}} & \multirow{3}{2cm}{\textbf{Feature \newline visualization}} & 
\multicolumn{2}{c|}{\textbf{Cosine similarity}} \\
\cline{6-7}
& & & & & \multirow{2}{2cm}{Short description} & \multirow{2}{2cm}{Long description} \\
& & & & & &  \\
\hline
great white shark, white shark, man-eater, man-eating shark, Carcharodon carcharias & \multirow{3}{1cm}{2} & \multirow{3}{1cm}{691} & \multirow{3}{1cm}{prediction} & \multirow{3}{2cm}{Figure \ref{fig:appendix_mturk_hard1}} & \multirow{3}{1cm}{0.1564} & \multirow{3}{1cm}{0.3373} \\
\hline
hermit crab & 125 & 1211 & label & Figure \ref{fig:appendix_mturk_hard2} & 0.1732 & 0.3826\\
\hline
goldfinch, Carduelis carduelis & \multirow{1}{1cm}{11} & \multirow{1}{1cm}{788} & \multirow{1}{1cm}{label} & \multirow{1}{2cm}{Figure \ref{fig:appendix_mturk_hard3}} & \multirow{1}{2cm}{0.2593} & \multirow{1}{2cm}{0.4046} \\
\hline
rock beauty, Holocanthus tricolor & \multirow{3}{1cm}{392} & \multirow{3}{1cm}{1348} & \multirow{3}{1cm}{label} & \multirow{3}{2cm}{Figure \ref{fig:appendix_mturk_hard4}} & \multirow{3}{1cm}{0.2157} & \multirow{3}{1cm}{0.4946}\\
\hline
pole & 733 & 1107 & label & Figure \ref{fig:appendix_mturk_hard5} & 0.2648 & 0.4703 \\
\hline
\end{tabular}
\caption{Examples from the Amazon Mechanical Turk study that workers found as difficult to describe. Short description is the answer to Q1 and Long description is the answer to Q2 in the crowd study (Figure \ref{fig:aws_questions}).}
\label{table:mturk_hard}
\end{table*}

\begin{figure*}
\centering
\begin{subfigure}{0.8\linewidth}
\centering
\includegraphics[trim=0cm 0cm 0cm 0.9cm, clip, width=\linewidth]{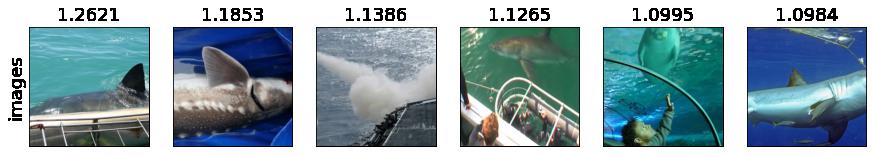}
\end{subfigure}\\
\begin{subfigure}{0.8\linewidth}
\centering
\includegraphics[trim=0cm 0cm 0cm 0.9cm, clip, width=\linewidth]{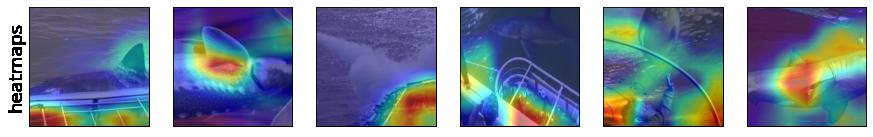}
\end{subfigure}\\
\begin{subfigure}{0.8\linewidth}
\centering
\includegraphics[trim=0cm 0cm 0cm 0.9cm, clip, width=\linewidth]{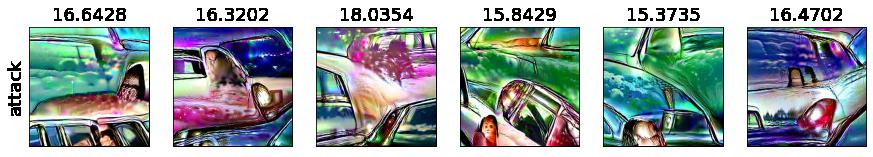}
\end{subfigure}
\captionsetup{width=.8\linewidth}
\caption{Visualization of feature[691], class[2] (white shark) and prediction grouping. \\Example descriptions: structure, high contrast lines, Psychedelic colors, triangle}
\label{fig:appendix_mturk_hard1}
\end{figure*}

\begin{figure*}
\centering
\begin{subfigure}{0.8\linewidth}
\centering
\includegraphics[trim=0cm 0cm 0cm 0.9cm, clip, width=\linewidth]{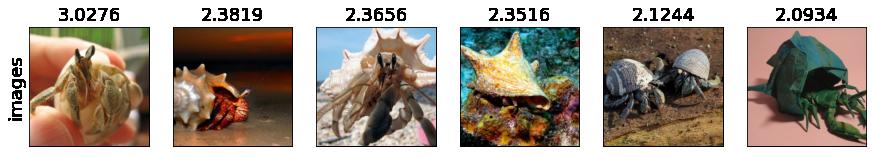}
\end{subfigure}\\
\begin{subfigure}{0.8\linewidth}
\centering
\includegraphics[trim=0cm 0cm 0cm 0.9cm, clip, width=\linewidth]{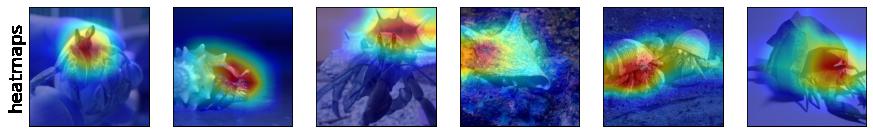}
\end{subfigure}\\
\begin{subfigure}{0.8\linewidth}
\centering
\includegraphics[trim=0cm 0cm 0cm 0.9cm, clip, width=\linewidth]{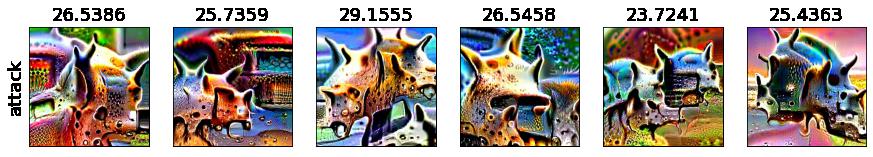}
\end{subfigure}
\captionsetup{width=.8\linewidth}
\caption{Visualization of feature[1211], class[125] (hermit crab) and label grouping. \\Example descriptions: creature body, Shells, protruded or snug-fitting, video game, hard shell}
\label{fig:appendix_mturk_hard2}
\end{figure*}

\begin{figure*}
\centering
\begin{subfigure}{0.8\linewidth}
\centering
\includegraphics[trim=0cm 0cm 0cm 0.9cm, clip, width=\linewidth]{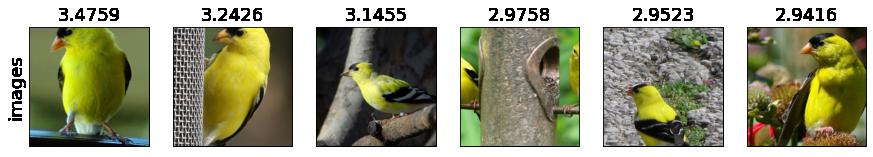}
\end{subfigure}\\
\begin{subfigure}{0.8\linewidth}
\centering
\includegraphics[trim=0cm 0cm 0cm 0.9cm, clip, width=\linewidth]{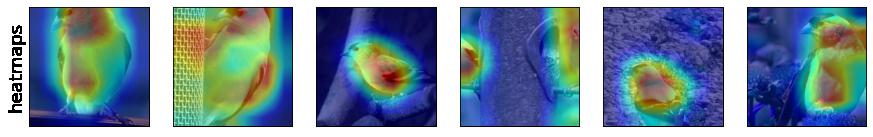}
\end{subfigure}\\
\begin{subfigure}{0.8\linewidth}
\centering
\includegraphics[trim=0cm 0cm 0cm 0.9cm, clip, width=\linewidth]{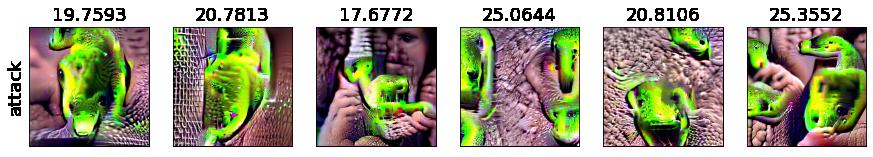}
\end{subfigure}
\captionsetup{width=.8\linewidth}
\caption{Visualization of feature[788], class[11] (goldfinch) and label grouping.\\Example descriptions: flying yellow being, rock, yellow spot, circular feathered body}
\label{fig:appendix_mturk_hard3}
\end{figure*}

\begin{figure*}
\centering
\begin{subfigure}{0.8\linewidth}
\centering
\includegraphics[trim=0cm 0cm 0cm 0.9cm, clip, width=\linewidth]{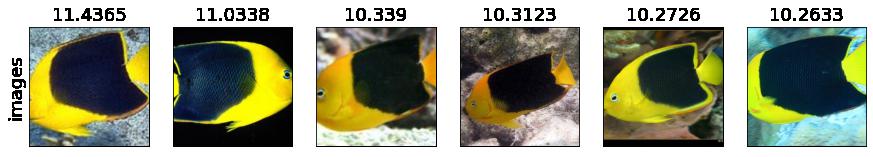}
\end{subfigure}\\
\begin{subfigure}{0.8\linewidth}
\centering
\includegraphics[trim=0cm 0cm 0cm 0.9cm, clip, width=\linewidth]{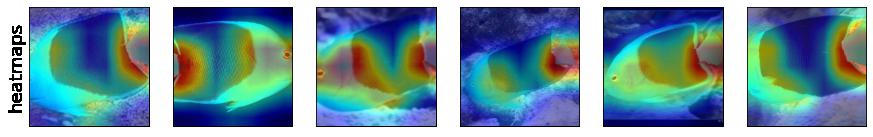}
\end{subfigure}\\
\begin{subfigure}{0.8\linewidth}
\centering
\includegraphics[trim=0cm 0cm 0cm 0.9cm, clip, width=\linewidth]{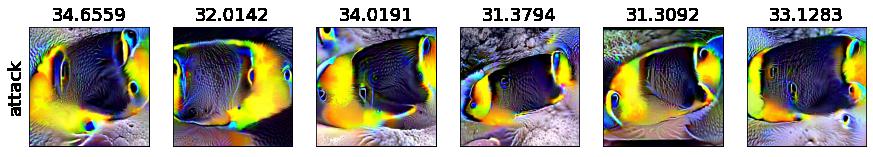}
\end{subfigure}
\captionsetup{width=.8\linewidth}
\caption{Visualization of feature[1348], class[392] (rock beauty) and label grouping.\\Example descriptions: edge, cave, nan, arrow shaped, rectangle}
\label{fig:appendix_mturk_hard4}
\end{figure*}

\begin{figure*}
\centering
\begin{subfigure}{0.8\linewidth}
\centering
\includegraphics[trim=0cm 0cm 0cm 0.9cm, clip, width=\linewidth]{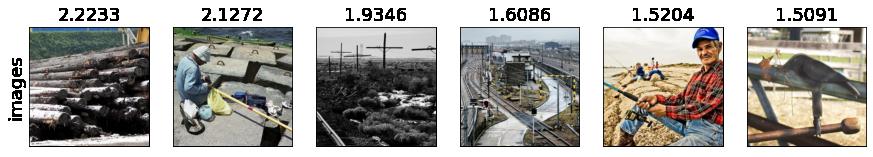}
\end{subfigure}\\
\begin{subfigure}{0.8\linewidth}
\centering
\includegraphics[trim=0cm 0cm 0cm 0.9cm, clip, width=\linewidth]{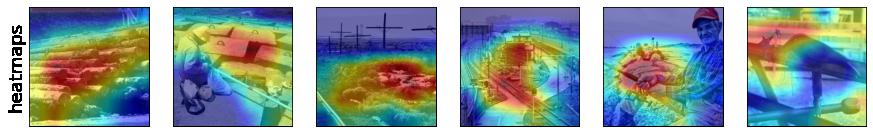}
\end{subfigure}\\
\begin{subfigure}{0.8\linewidth}
\centering
\includegraphics[trim=0cm 0cm 0cm 0.9cm, clip, width=\linewidth]{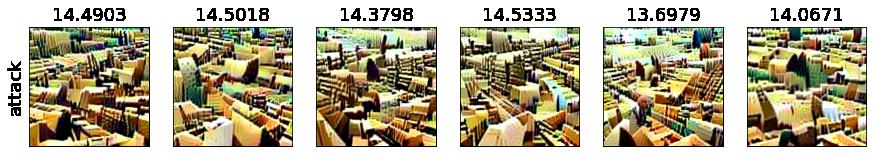}
\end{subfigure}
\captionsetup{width=.8\linewidth}
\caption{Visualization of feature[1107], class[733] (pole) and label grouping.\\Descriptions: long wooden beam, cube shapes, cells, rainbow hued circle, long pillars}
\label{fig:appendix_mturk_hard5}
\end{figure*}

\clearpage

\subsection{Examples with most votes for Section C, Feature Attacks (Question 5 in Figure \ref{fig:aws_questions}) }\label{sec:examples_feature_attack_important}

\begin{figure*}[ht!]
\centering
\begin{subfigure}{0.8\linewidth}
\centering
\includegraphics[trim=0cm 0cm 0cm 0.9cm, clip, width=\linewidth]{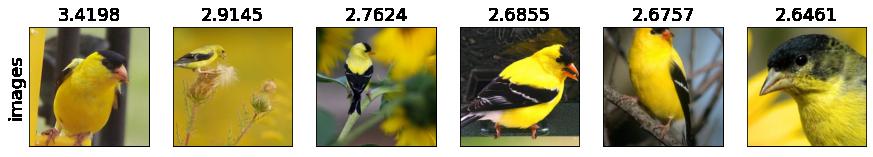}
\end{subfigure}\\
\begin{subfigure}{0.8\linewidth}
\centering
\includegraphics[trim=0cm 0cm 0cm 0.9cm, clip, width=\linewidth]{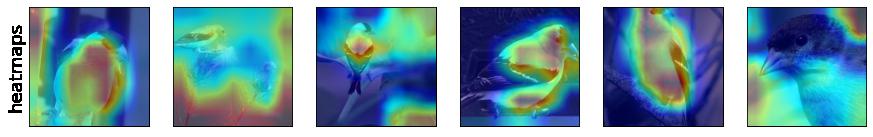}
\end{subfigure}\\
\begin{subfigure}{0.8\linewidth}
\centering
\includegraphics[trim=0cm 0cm 0cm 0.9cm, clip, width=\linewidth]{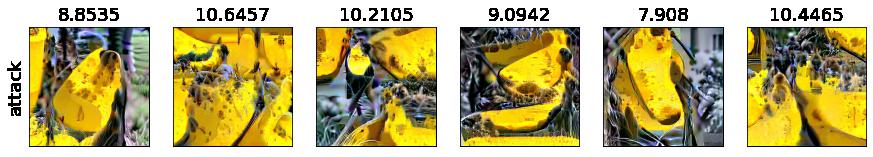}
\end{subfigure}
\captionsetup{width=.8\linewidth}
\caption{Visualization of feature[1979], class[11] (goldfinch) and label grouping.}
\label{fig:appendix_sectionc_goldfinch}
\end{figure*}

\begin{figure*}[ht!]
\centering
\begin{subfigure}{0.8\linewidth}
\centering
\includegraphics[trim=0cm 0cm 0cm 0.9cm, clip, width=\linewidth]{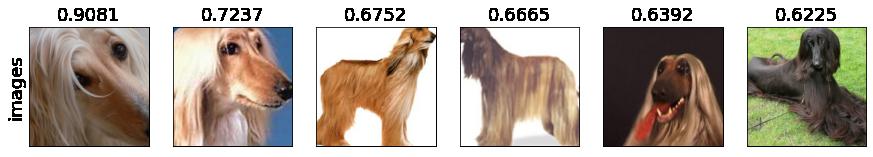}
\end{subfigure}\\
\begin{subfigure}{0.8\linewidth}
\centering
\includegraphics[trim=0cm 0cm 0cm 0.9cm, clip, width=\linewidth]{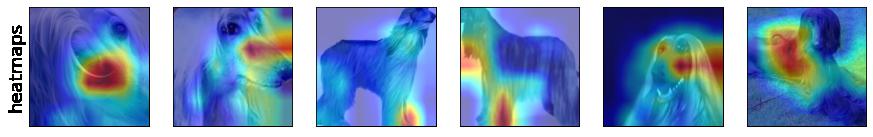}
\end{subfigure}\\
\begin{subfigure}{0.8\linewidth}
\centering
\includegraphics[trim=0cm 0cm 0cm 0.9cm, clip, width=\linewidth]{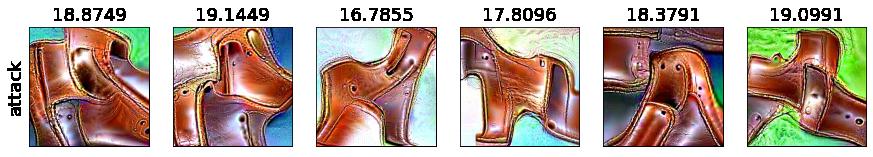}
\end{subfigure}
\captionsetup{width=.8\linewidth}
\caption{Visualization of feature[1185], class[110] (afghan hound) and prediction grouping.}
\label{fig:appendix_sectionc_afghanhound}
\end{figure*}

\begin{figure*}[ht!]
\centering
\begin{subfigure}{0.8\linewidth}
\centering
\includegraphics[trim=0cm 0cm 0cm 0.9cm, clip, width=\linewidth]{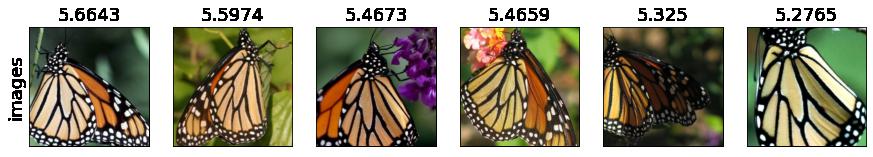}
\end{subfigure}\\
\begin{subfigure}{0.8\linewidth}
\centering
\includegraphics[trim=0cm 0cm 0cm 0.9cm, clip, width=\linewidth]{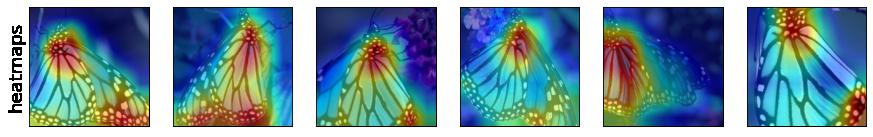}
\end{subfigure}\\
\begin{subfigure}{0.8\linewidth}
\centering
\includegraphics[trim=0cm 0cm 0cm 0.9cm, clip, width=\linewidth]{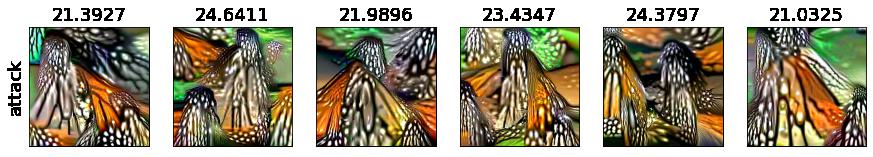}
\end{subfigure}
\captionsetup{width=.8\linewidth}
\caption{Visualization of feature[594], class[323] (monarch butterfly) and label grouping.}
\label{fig:appendix_sectionc_butterfly}
\end{figure*}

\begin{figure*}[ht!]
\centering
\begin{subfigure}{0.8\linewidth}
\centering
\includegraphics[trim=0cm 0cm 0cm 0.9cm, clip, width=\linewidth]{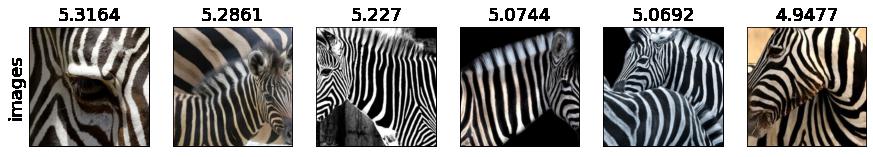}
\end{subfigure}\\
\begin{subfigure}{0.8\linewidth}
\centering
\includegraphics[trim=0cm 0cm 0cm 0.9cm, clip, width=\linewidth]{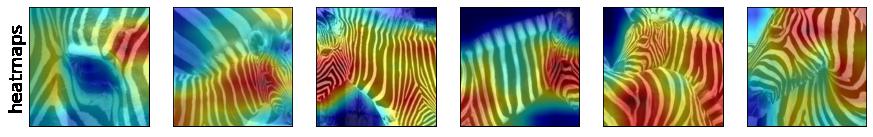}
\end{subfigure}\\
\begin{subfigure}{0.8\linewidth}
\centering
\includegraphics[trim=0cm 0cm 0cm 0.9cm, clip, width=\linewidth]{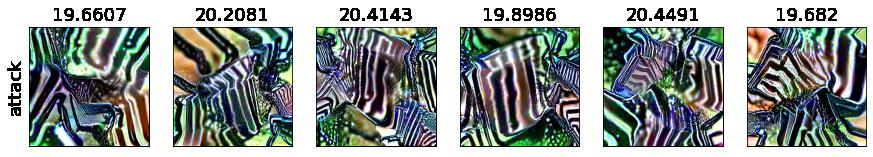}
\end{subfigure}
\captionsetup{width=.8\linewidth}
\caption{Visualization of feature[1604], class[340] (zebra) and label grouping.}
\label{fig:appendix_sectionc_zebra}
\end{figure*}
\clearpage

\subsection{Examples with most votes for Both (Question 5 in Figure \ref{fig:aws_questions}) }\label{sec:examples_both_important}

\begin{figure*}[ht!]
\centering
\begin{subfigure}{0.8\linewidth}
\centering
\includegraphics[trim=0cm 0cm 0cm 0.9cm, clip, width=\linewidth]{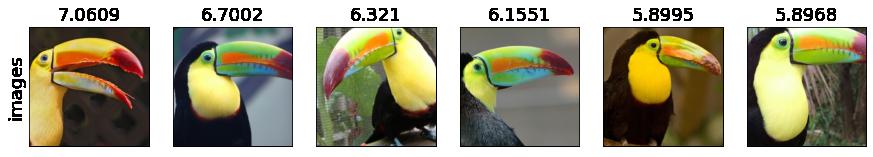}
\end{subfigure}\\
\begin{subfigure}{0.8\linewidth}
\centering
\includegraphics[trim=0cm 0cm 0cm 0.9cm, clip, width=\linewidth]{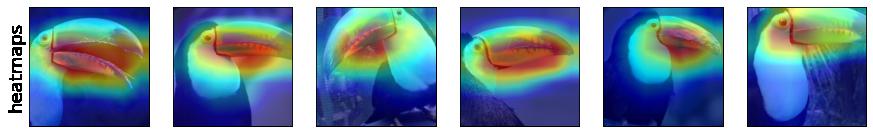}
\end{subfigure}\\
\begin{subfigure}{0.8\linewidth}
\centering
\includegraphics[trim=0cm 0cm 0cm 0.9cm, clip, width=\linewidth]{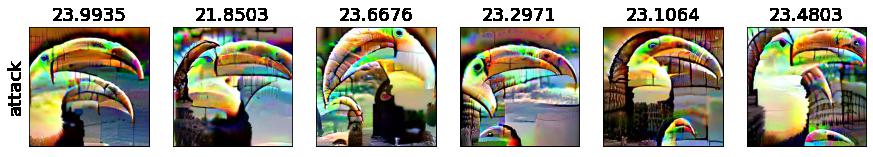}
\end{subfigure}
\captionsetup{width=.8\linewidth}
\caption{Visualization of feature[1486], class[96] (toucan) and prediction grouping.}
\label{fig:appendix_both_toucan}
\end{figure*}

\begin{figure*}[ht!]
\centering
\begin{subfigure}{0.8\linewidth}
\centering
\includegraphics[trim=0cm 0cm 0cm 0.9cm, clip, width=\linewidth]{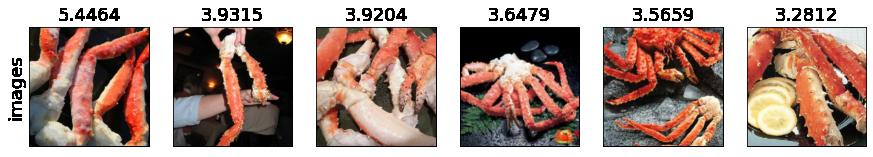}
\end{subfigure}\\
\begin{subfigure}{0.8\linewidth}
\centering
\includegraphics[trim=0cm 0cm 0cm 0.9cm, clip, width=\linewidth]{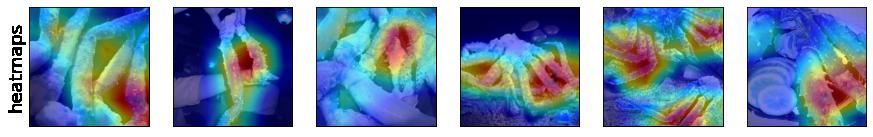}
\end{subfigure}\\
\begin{subfigure}{0.8\linewidth}
\centering
\includegraphics[trim=0cm 0cm 0cm 0.9cm, clip, width=\linewidth]{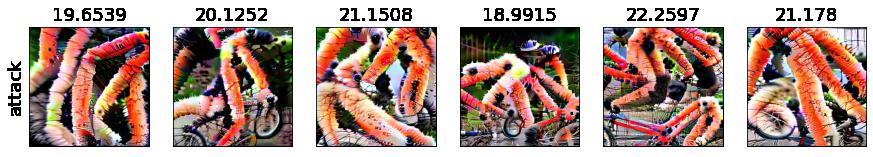}
\end{subfigure}
\captionsetup{width=.8\linewidth}
\caption{Visualization of feature[191], class[121] (crab) and prediction grouping.}
\label{fig:appendix_both_crab}
\end{figure*}

\begin{figure*}[ht!]
\centering
\begin{subfigure}{0.8\linewidth}
\centering
\includegraphics[trim=0cm 0cm 0cm 0.9cm, clip, width=\linewidth]{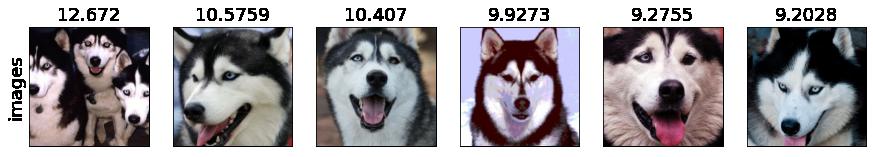}
\end{subfigure}\\
\begin{subfigure}{0.8\linewidth}
\centering
\includegraphics[trim=0cm 0cm 0cm 0.9cm, clip, width=\linewidth]{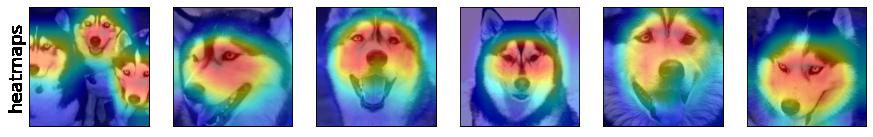}
\end{subfigure}\\
\begin{subfigure}{0.8\linewidth}
\centering
\includegraphics[trim=0cm 0cm 0cm 0.9cm, clip, width=\linewidth]{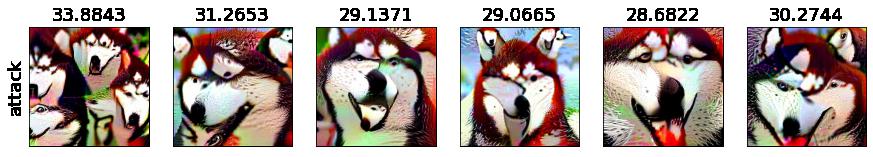}
\end{subfigure}
\captionsetup{width=.8\linewidth}
\caption{Visualization of feature[287], class[248] (husky) and label grouping.}
\label{fig:appendix_both_eskimodog}
\end{figure*}

\begin{figure*}[ht!]
\centering
\begin{subfigure}{0.8\linewidth}
\centering
\includegraphics[trim=0cm 0cm 0cm 0.9cm, clip, width=\linewidth]{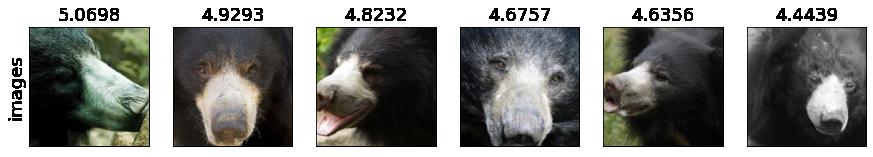}
\end{subfigure}\\
\begin{subfigure}{0.8\linewidth}
\centering
\includegraphics[trim=0cm 0cm 0cm 0.9cm, clip, width=\linewidth]{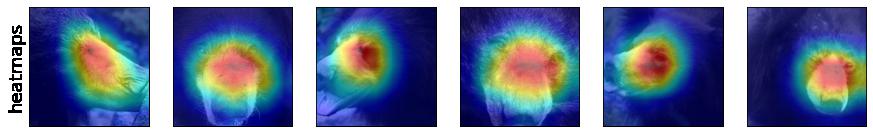}
\end{subfigure}\\
\begin{subfigure}{0.8\linewidth}
\centering
\includegraphics[trim=0cm 0cm 0cm 0.9cm, clip, width=\linewidth]{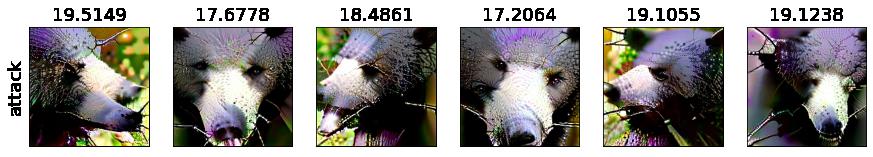}
\end{subfigure}
\captionsetup{width=.8\linewidth}
\caption{Visualization of feature[120], class[297] (sloth bear) and prediction grouping.}
\label{fig:appendix_both_slothbear}
\end{figure*}

\begin{figure*}[ht!]
\centering
\begin{subfigure}{0.8\linewidth}
\centering
\includegraphics[trim=0cm 0cm 0cm 0.9cm, clip, width=\linewidth]{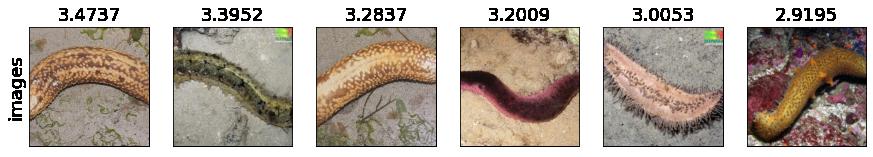}
\end{subfigure}\\
\begin{subfigure}{0.8\linewidth}
\centering
\includegraphics[trim=0cm 0cm 0cm 0.9cm, clip, width=\linewidth]{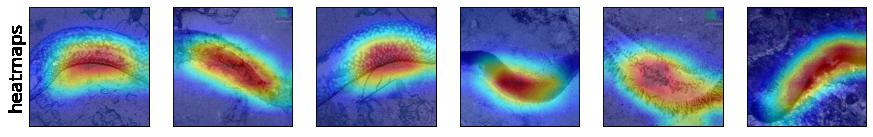}
\end{subfigure}\\
\begin{subfigure}{0.8\linewidth}
\centering
\includegraphics[trim=0cm 0cm 0cm 0.9cm, clip, width=\linewidth]{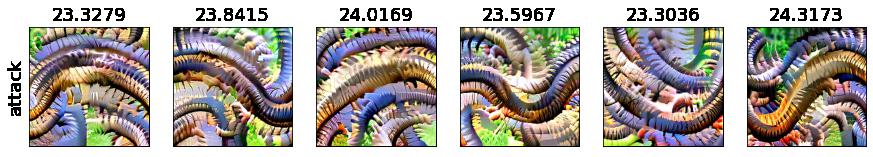}
\end{subfigure}
\captionsetup{width=.8\linewidth}
\caption{Visualization of feature[1465], class[329] (sea cucumber) and prediction grouping.}
\label{fig:appendix_both_seacucumber}
\end{figure*}

\begin{figure*}[ht!]
\centering
\begin{subfigure}{0.8\linewidth}
\centering
\includegraphics[trim=0cm 0cm 0cm 0.9cm, clip, width=\linewidth]{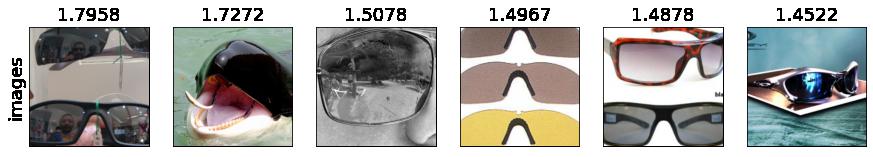}
\end{subfigure}\\
\begin{subfigure}{0.8\linewidth}
\centering
\includegraphics[trim=0cm 0cm 0cm 0.9cm, clip, width=\linewidth]{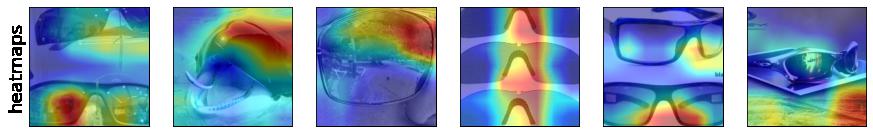}
\end{subfigure}\\
\begin{subfigure}{0.8\linewidth}
\centering
\includegraphics[trim=0cm 0cm 0cm 0.9cm, clip, width=\linewidth]{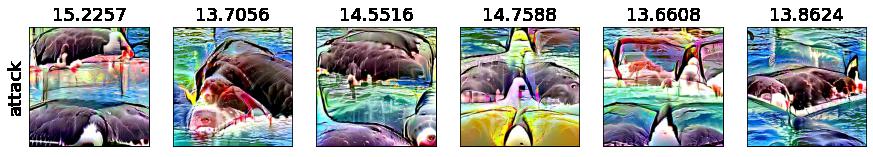}
\end{subfigure}
\captionsetup{width=.8\linewidth}
\caption{Visualization of feature[2012], class[836] (sunglass) and prediction grouping.}
\label{fig:appendix_both_sunglass}
\end{figure*}

\begin{figure*}[ht!]
\centering
\begin{subfigure}{0.8\linewidth}
\centering
\includegraphics[trim=0cm 0cm 0cm 0.9cm, clip, width=\linewidth]{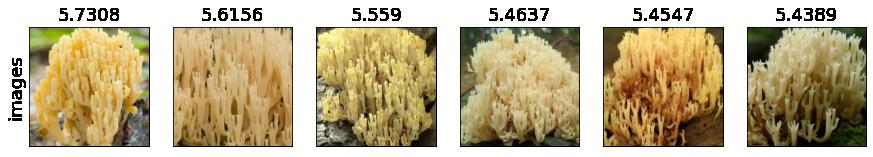}
\end{subfigure}\\
\begin{subfigure}{0.8\linewidth}
\centering
\includegraphics[trim=0cm 0cm 0cm 0.9cm, clip, width=\linewidth]{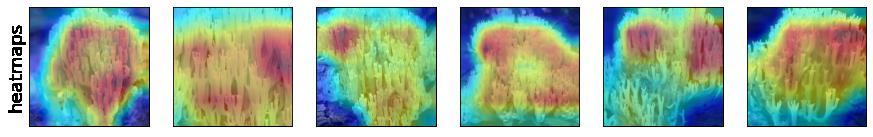}
\end{subfigure}\\
\begin{subfigure}{0.8\linewidth}
\centering
\includegraphics[trim=0cm 0cm 0cm 0.9cm, clip, width=\linewidth]{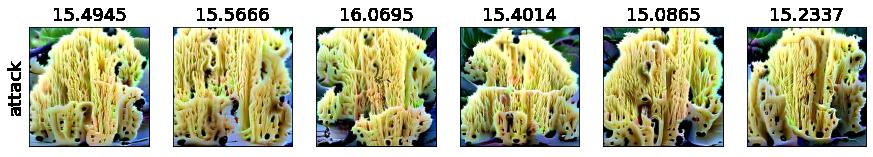}
\end{subfigure}
\captionsetup{width=.8\linewidth}
\caption{Visualization of feature[1917], class[991] (coral fungus) and prediction grouping.}
\label{fig:appendix_both_coralfungus}
\end{figure*}

\clearpage

\section{User study with ML practitioners}\label{sec:app_user_study}
\vspace{1cm}
\begin{table*}[h!]
      \centering
            \begin{tabular}{ll} \toprule
           \textbf{Role}  &  \textbf{Participants} \\
           \midrule
           ML Engineer  &  5 [P2, P4, P5, P11, P18]  \\
           Applied Scientist  & 2  [P9, P12] \\
           Researcher & 4  [P1, P7, P16, P17] \\
           Data Scientist & 3  [P10, P20, P21] \\
           \bottomrule
            \textbf{Experience in ML}  &  \textbf{Participants} \\
            \midrule
           1 - 2 years  & 1 [P2]  \\
           2 - 5 years &  4 [P5, P10, P11, P20] \\
           5 - 10 years &  5 [P4, P7, P16, P17, P18] \\
           $>$ 10 years &  4 [P1, P9, P12, P21] \\
          \bottomrule
            \end{tabular}
    \caption{Distribution of roles and years of experience in Machine Learning among ML practitioners in the study.}
    \label{tab:user_study}
\end{table*}
\vspace{1cm}
\begin{table*}[h!]
    \centering \begin{tabular}{lllrl} \toprule
           \textbf{Class id}  & \textbf{Class name}  & \textbf{Grouping}   & \textbf{Robust Resnet-50 Top-1 Error} &  \textbf{Participants} \\
           \midrule
           424 & Barbershop         & prediction    & $68.32\%$     & 3 [P10, P18, P20]     \\
           703 & Park Bench         & label         & $33.31\%$     & 3  [P9, P11, P17]     \\
           785 & Seat Belt          &  label        & $33.23\%$     & 4  [P2, P4, P12, P21] \\
           820 & Steam Locomotive   & label         & $6.69\%$      & 1  [P1]               \\
           282 & Tiger Cat          & label         & $77.15\%$     & 3  [P5, P7, P16]      \\
           \bottomrule
            \end{tabular}
    \caption{Distribution of the first class groupings among machine-learning practitioners. The five examples contained features that were considered as ``easy to describe" by Mturk workers to facilitate onboarding. The second class grouping was instead assigned randomly from the set of 120 class groupings that were part of the MTurk study.}
    \label{tab:user_study_conditions_1}
\end{table*}

\section{Comparison between the interpretations of a robust and non-robust model}\label{sec:comparison_robust_nonrobust}
To compare the interpretations of a robust model with a non-robust model, we analyzed the failures of top-$5$ classes with highest number of failures in the non-robust model (using grouping by label). The feature visualizations for the $5$ classes and the respective most important feature for failure explanation are given below. We observe that using a robust model for feature extraction and visualization leads to significantly more interpretable visualizations qualitatively. While we did not conduct quantitative comparisons with humans studies (robust vs. non-robust features) we encourage future research in this space that may exclusively focus in describing such differences at a larger extent.

\clearpage
\subsection{Class name: water jug}

\begin{figure*}[ht!]
\centering
\begin{subfigure}{0.8\linewidth}
\centering
\includegraphics[trim=0cm 0cm 0cm 0.9cm, clip, width=\linewidth]{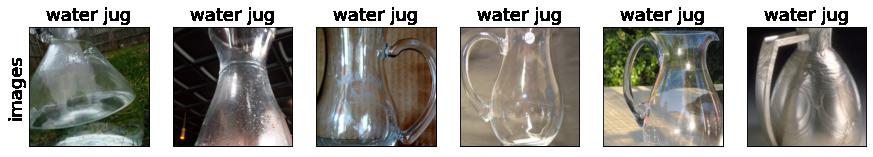}
\end{subfigure}\\
\begin{subfigure}{0.8\linewidth}
\centering
\includegraphics[trim=0cm 0cm 0cm 0.9cm, clip, width=\linewidth]{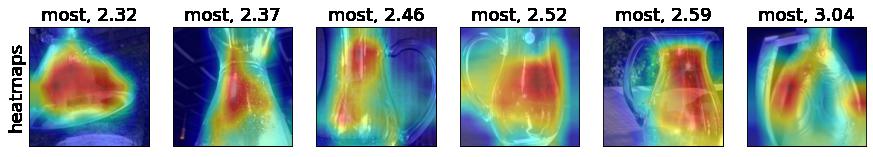}
\end{subfigure}\\
\begin{subfigure}{0.8\linewidth}
\centering
\includegraphics[trim=0cm 0cm 0cm 0.9cm, clip, width=\linewidth]{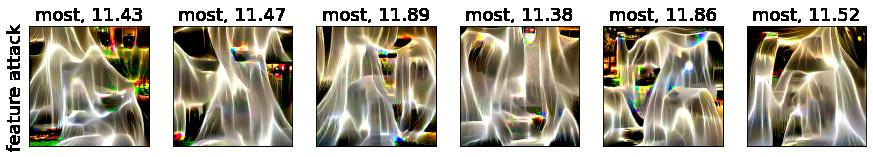}
\end{subfigure}
\captionsetup{width=.8\linewidth}
\caption{Feature visualization \textbf{using a robust model}. Visualization of feature[1725], class[899] (water jug) and label grouping.}
\label{fig:comparison_robust_waterjug}
\end{figure*}

\begin{figure*}[ht!]
\centering
\begin{subfigure}{0.8\linewidth}
\centering
\includegraphics[trim=0cm 0cm 0cm 0.9cm, clip, width=\linewidth]{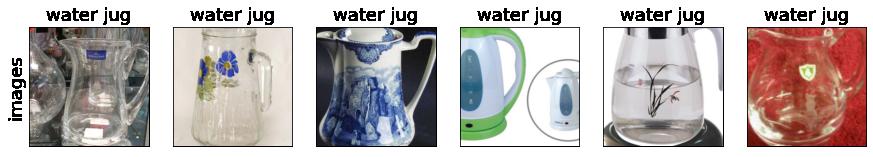}
\end{subfigure}\\
\begin{subfigure}{0.8\linewidth}
\centering
\includegraphics[trim=0cm 0cm 0cm 0.9cm, clip, width=\linewidth]{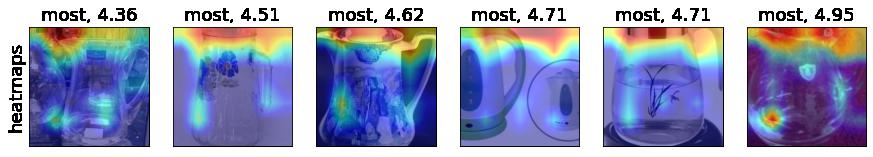}
\end{subfigure}\\
\begin{subfigure}{0.8\linewidth}
\centering
\includegraphics[trim=0cm 0cm 0cm 0.9cm, clip, width=\linewidth]{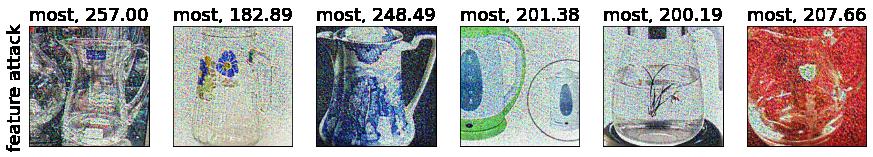}
\end{subfigure}
\captionsetup{width=.8\linewidth}
\caption{Feature visualization \textbf{using a non-robust model}. Visualization of feature[1357], class[899] (water jug) and label grouping.}
\label{fig:comparison_nonrobust_waterjug}
\end{figure*}

\clearpage

\subsection{Class name: horned viper}

\begin{figure*}[ht!]
\centering
\begin{subfigure}{0.8\linewidth}
\centering
\includegraphics[trim=0cm 0cm 0cm 0.9cm, clip, width=\linewidth]{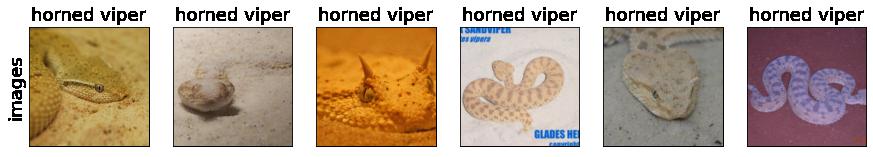}
\end{subfigure}\\
\begin{subfigure}{0.8\linewidth}
\centering
\includegraphics[trim=0cm 0cm 0cm 0.9cm, clip, width=\linewidth]{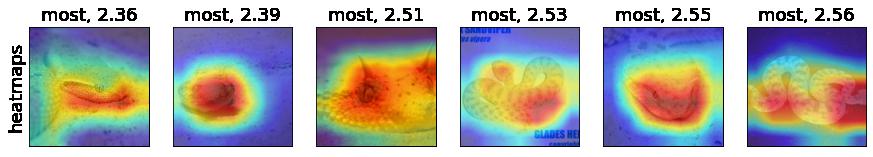}
\end{subfigure}\\
\begin{subfigure}{0.8\linewidth}
\centering
\includegraphics[trim=0cm 0cm 0cm 0.9cm, clip, width=\linewidth]{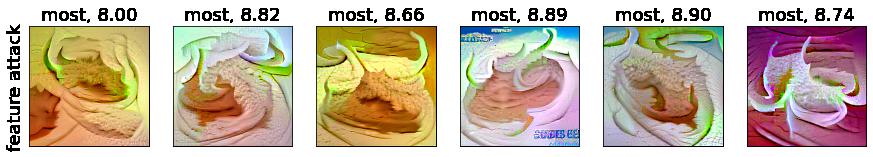}
\end{subfigure}
\captionsetup{width=.8\linewidth}
\caption{Feature visualization \textbf{using a robust model}. Visualization of feature[54], class[66] (horned viper) and label grouping.}
\label{fig:comparison_robust_hornedviper}
\end{figure*}

\begin{figure*}[ht!]
\centering
\begin{subfigure}{0.8\linewidth}
\centering
\includegraphics[trim=0cm 0cm 0cm 0.9cm, clip, width=\linewidth]{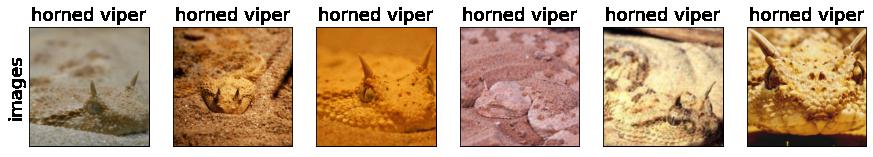}
\end{subfigure}\\
\begin{subfigure}{0.8\linewidth}
\centering
\includegraphics[trim=0cm 0cm 0cm 0.9cm, clip, width=\linewidth]{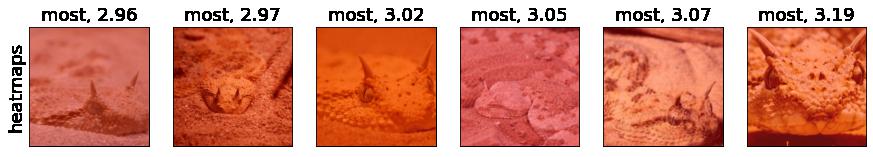}
\end{subfigure}\\
\begin{subfigure}{0.8\linewidth}
\centering
\includegraphics[trim=0cm 0cm 0cm 0.9cm, clip, width=\linewidth]{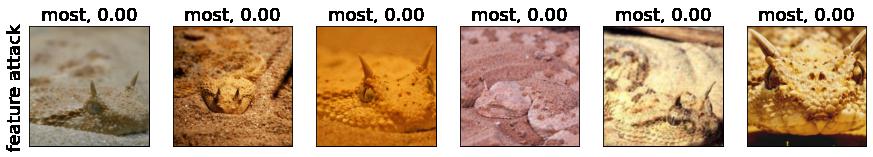}
\end{subfigure}
\captionsetup{width=.8\linewidth}
\caption{Feature visualization \textbf{using a non-robust model}. Visualization of feature[378], class[66] (horned viper) and label grouping.}
\label{fig:comparison_nonrobust_hornedviper}
\end{figure*}

\clearpage

\subsection{Class name: tiger cat}

\begin{figure*}[ht!]
\centering
\begin{subfigure}{0.8\linewidth}
\centering
\includegraphics[trim=0cm 0cm 0cm 0.9cm, clip, width=\linewidth]{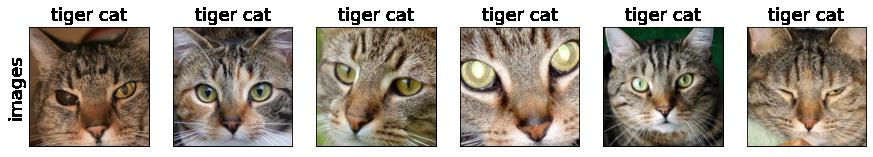}
\end{subfigure}\\
\begin{subfigure}{0.8\linewidth}
\centering
\includegraphics[trim=0cm 0cm 0cm 0.9cm, clip, width=\linewidth]{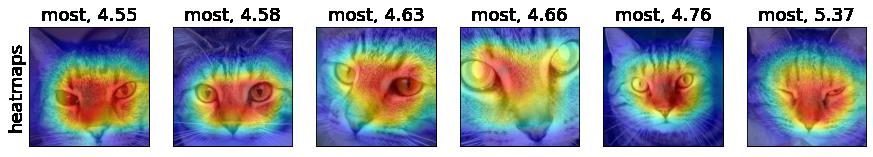}
\end{subfigure}\\
\begin{subfigure}{0.8\linewidth}
\centering
\includegraphics[trim=0cm 0cm 0cm 0.9cm, clip, width=\linewidth]{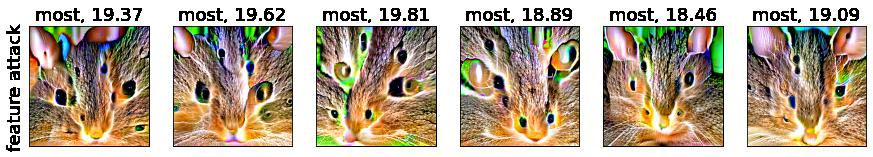}
\end{subfigure}
\captionsetup{width=.8\linewidth}
\caption{Feature visualization \textbf{using a robust model}. Visualization of feature[544], class[282] (tiger cat) and label grouping.}
\label{fig:comparison_robust_tigercat}
\end{figure*}

\begin{figure*}[ht!]
\centering
\begin{subfigure}{0.8\linewidth}
\centering
\includegraphics[trim=0cm 0cm 0cm 0.9cm, clip, width=\linewidth]{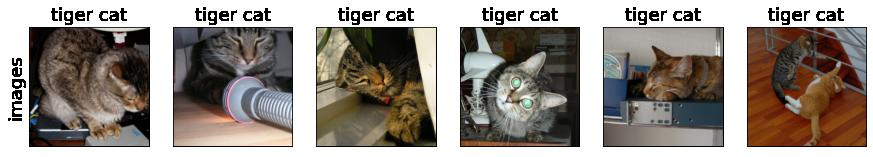}
\end{subfigure}\\
\begin{subfigure}{0.8\linewidth}
\centering
\includegraphics[trim=0cm 0cm 0cm 0.9cm, clip, width=\linewidth]{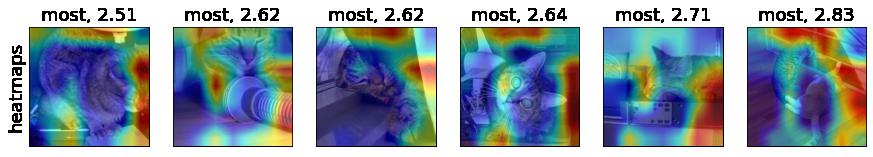}
\end{subfigure}\\
\begin{subfigure}{0.8\linewidth}
\centering
\includegraphics[trim=0cm 0cm 0cm 0.9cm, clip, width=\linewidth]{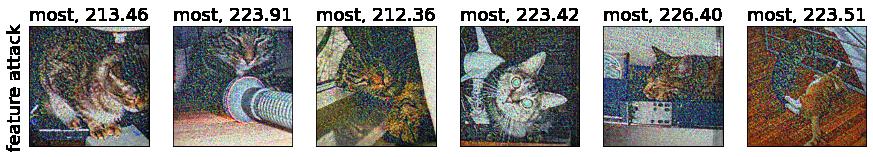}
\end{subfigure}
\captionsetup{width=.8\linewidth}
\caption{Feature visualization \textbf{using a non-robust model}. Visualization of feature[1075], class[282] (tiger cat) and label grouping.}
\label{fig:comparison_nonrobust_tigercat}
\end{figure*}

\clearpage

\subsection{Class name: tape player}

\begin{figure*}[ht!]
\centering
\begin{subfigure}{0.8\linewidth}
\centering
\includegraphics[trim=0cm 0cm 0cm 0.9cm, clip, width=\linewidth]{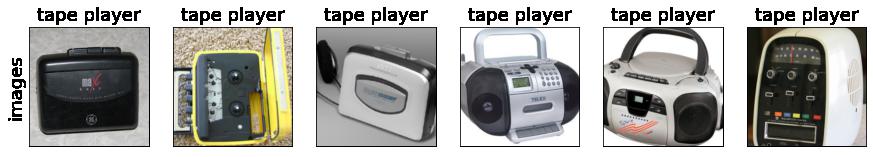}
\end{subfigure}\\
\begin{subfigure}{0.8\linewidth}
\centering
\includegraphics[trim=0cm 0cm 0cm 0.9cm, clip, width=\linewidth]{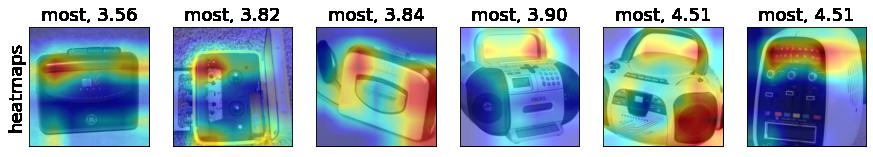}
\end{subfigure}\\
\begin{subfigure}{0.8\linewidth}
\centering
\includegraphics[trim=0cm 0cm 0cm 0.9cm, clip, width=\linewidth]{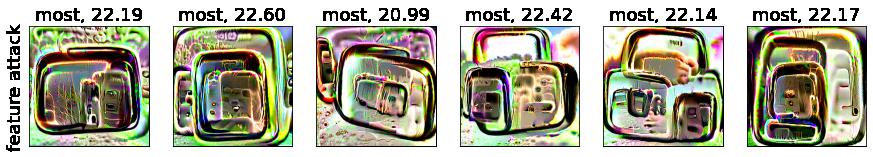}
\end{subfigure}
\captionsetup{width=.8\linewidth}
\caption{Feature visualization \textbf{using a robust model}. Visualization of feature[1751], class[848] (tape player) and label grouping.}
\label{fig:comparison_robust_tapeplayer}
\end{figure*}

\begin{figure*}[ht!]
\centering
\begin{subfigure}{0.8\linewidth}
\centering
\includegraphics[trim=0cm 0cm 0cm 0.9cm, clip, width=\linewidth]{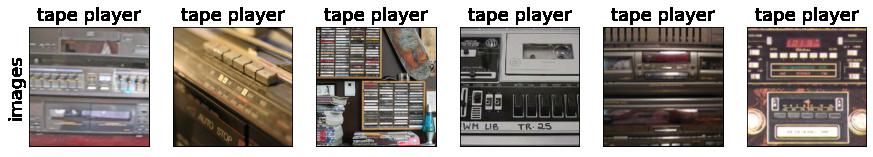}
\end{subfigure}\\
\begin{subfigure}{0.8\linewidth}
\centering
\includegraphics[trim=0cm 0cm 0cm 0.9cm, clip, width=\linewidth]{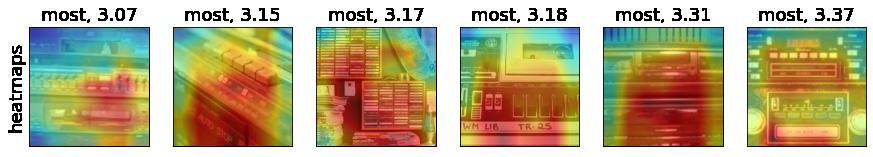}
\end{subfigure}\\
\begin{subfigure}{0.8\linewidth}
\centering
\includegraphics[trim=0cm 0cm 0cm 0.9cm, clip, width=\linewidth]{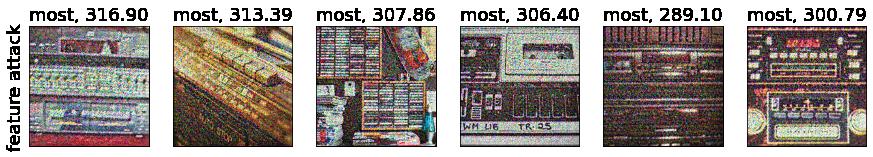}
\end{subfigure}
\captionsetup{width=.8\linewidth}
\caption{Feature visualization \textbf{using a non-robust model}. Visualization of feature[935], class[848] (tape player) and label grouping.}
\label{fig:comparison_nonrobust_tapeplayer}
\end{figure*}

\clearpage
\subsection{Class name: overskirt}

\begin{figure*}[ht!]
\centering
\begin{subfigure}{0.8\linewidth}
\centering
\includegraphics[trim=0cm 0cm 0cm 0.9cm, clip, width=\linewidth]{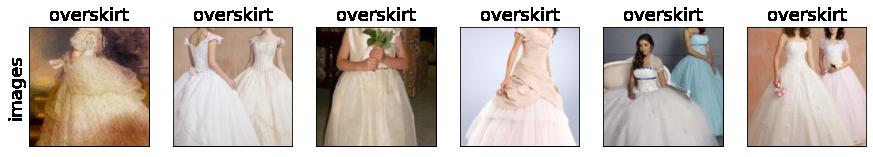}
\end{subfigure}\\
\begin{subfigure}{0.8\linewidth}
\centering
\includegraphics[trim=0cm 0cm 0cm 0.9cm, clip, width=\linewidth]{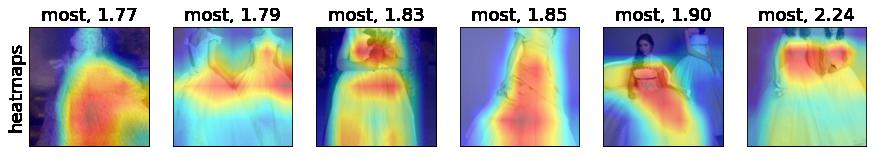}
\end{subfigure}\\
\begin{subfigure}{0.8\linewidth}
\centering
\includegraphics[trim=0cm 0cm 0cm 0.9cm, clip, width=\linewidth]{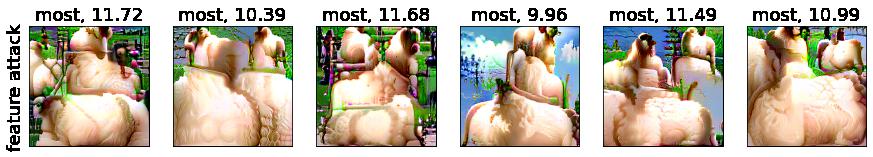}
\end{subfigure}
\captionsetup{width=.8\linewidth}
\caption{Feature visualization \textbf{using a robust model}. Visualization of feature[343], class[689] (overskirt) and label grouping.}
\label{fig:comparison_robust_maillot}
\end{figure*}

\begin{figure*}[ht!]
\centering
\begin{subfigure}{0.8\linewidth}
\centering
\includegraphics[trim=0cm 0cm 0cm 0.9cm, clip, width=\linewidth]{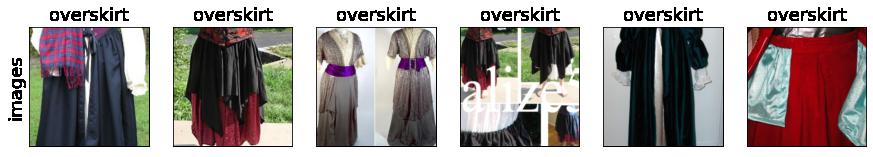}
\end{subfigure}\\
\begin{subfigure}{0.8\linewidth}
\centering
\includegraphics[trim=0cm 0cm 0cm 0.9cm, clip, width=\linewidth]{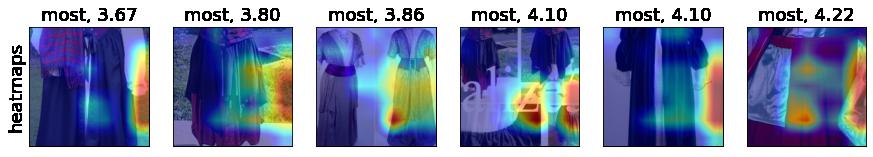}
\end{subfigure}\\
\begin{subfigure}{0.8\linewidth}
\centering
\includegraphics[trim=0cm 0cm 0cm 0.9cm, clip, width=\linewidth]{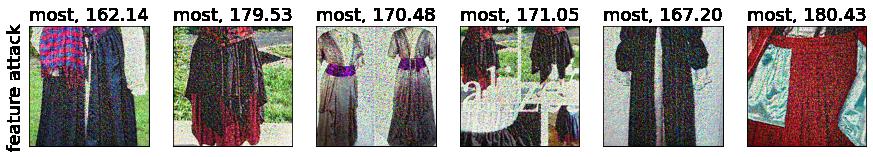}
\end{subfigure}
\captionsetup{width=.8\linewidth}
\caption{Feature visualization \textbf{using a non-robust model}. Visualization of feature[1405], class[689] (overskirt) and label grouping.}
\label{fig:comparison_nonrobust_overskirt}
\end{figure*}


\end{document}